\pdfoutput=1

\documentclass[nohyperref]{article}

\usepackage{microtype}
\usepackage{graphicx}
\usepackage{subfigure}
\usepackage{booktabs} 

\usepackage{hyperref}

\usepackage[accepted]{icml2022}

\usepackage{amsmath}
\usepackage{amssymb}
\usepackage{mathtools}
\usepackage{amsthm}

\usepackage[normalem]{ulem}
\usepackage{arydshln}
\usepackage{graphicx}
\usepackage{wrapfig}
\usepackage[font=small]{caption}
\usepackage{lipsum}
\usepackage{pifont}
\newcommand{\cmark}{\ding{51}}%
\newcommand{\xmark}{\ding{55}}%
\definecolor{navyblue}{rgb}{0.0, 0.0, 0.5}
\definecolor{darkblue}{rgb}{0.0, 0.0, 0.55}
\hypersetup{
    colorlinks = true,
    citecolor=darkblue,
    linkcolor=red,
    urlcolor  = blue,
    anchorcolor = blue}
\usepackage{enumitem}
\usepackage{multirow}

\definecolor{Blue}{rgb}{0,0,0.8}
\definecolor{Green}{rgb}{0,0.4,0.7}
\definecolor{airforceblue}{rgb}{0.36, 0.54, 0.66}
\definecolor{ao(english)}{rgb}{0.0, 0.5, 0.0}
\definecolor{azure(colorwheel)}{rgb}{0.0, 0.5, 1.0}
\definecolor{crimson}{rgb}{0.86, 0.08, 0.24}
\definecolor{darkcerulean}{rgb}{0.03, 0.27, 0.49}
\definecolor{cobalt}{rgb}{0.0, 0.28, 0.67}
\definecolor{rosegold}{rgb}{0.72, 0.43, 0.47}
\definecolor{orange-red}{rgb}{1.0, 0.27, 0.0}
\definecolor{mountainmeadow}{rgb}{0.19, 0.73, 0.56}
\definecolor{malachite}{rgb}{0.04, 0.85, 0.32}
\definecolor{darkblue}{rgb}{0.0, 0.0, 0.55}

\definecolor{customblue}{rgb}{0.2, 0.35, 0.8}

\usepackage{hyperref}
\hypersetup{colorlinks=true}
\hypersetup{linktoc=all}
\hypersetup{citecolor=darkblue}
\hypersetup{linkcolor=crimson}
\hypersetup{urlcolor=darkblue}
\usepackage[all]{hypcap}
\usepackage[percent]{overpic}
\usepackage[usestackEOL]{stackengine}

\usepackage[nameinlink]{cleveref}
\creflabelformat{equation}{#2\textup{#1}#3}  
\crefname{assumption}{assumption}{assumptions}

\definecolor{gg}{gray}{0.9}

\creflabelformat{equation}{#2\textup{#1}#3}  

\theoremstyle{plain}

\theoremstyle{definition}

\theoremstyle{remark}

\usepackage[textsize=tiny]{todonotes}

\icmltitlerunning{Factorized-FL: Agnostic Personalized Federated Learning with Kernel Factorization \& Similarity Matching}

\begin{document}

\twocolumn[
\icmltitle{Factorized-FL: Agnostic Personalized Federated Learning \\ with Kernel Factorization \& Similarity Matching}



\icmlsetsymbol{equal}{*}




\begin{icmlauthorlist}
\icmlauthor{Wonyong Jeong}{to,goo}
\icmlauthor{Sung Ju Hwang}{to,goo}
\end{icmlauthorlist}

\icmlaffiliation{to}{Graduate School of Artificial Intelligence, KAIST, Seoul, South Korea}
\icmlaffiliation{goo}{AITRICS, Seoul, South Korea}

\icmlcorrespondingauthor{Wonyong Jeong}{wyjeong@kaist.ac.kr}
\icmlcorrespondingauthor{Sung Ju Hwang}{sjhwang82@kaist.ac.kr}

\icmlkeywords{Machine Learning, ICML}

\vskip 0.3in
]



\printAffiliationsAndNotice{}  

\begin{abstract}

In real-world federated learning scenarios, participants could have their own personalized labels which are incompatible with those from other clients, due to using different label permutations or tackling completely different tasks or domains. However, most existing FL approaches cannot effectively tackle such extremely heterogeneous scenarios since they often assume that (1) all participants use a synchronized set of labels, and (2) they train on the same task from the same domain. In this work, to tackle these challenges, we introduce \texttt{Factorized-FL}, which allows to effectively tackle label- and task-heterogeneous federated learning settings by factorizing the model parameters into a pair of vectors, where one captures the common knowledge across different labels and tasks and the other captures knowledge specific to the task each local model tackles. Moreover, based on the distance in the client-specific vector space, \texttt{Factorized-FL} performs selective aggregation scheme to utilize only the knowledge from the relevant participants for each client. We extensively validate our method on both label- and domain-heterogeneous settings, on which it outperforms the state-of-the-art personalized federated learning methods.

\end{abstract}

\section{Introduction}
\label{sec:intro}

\begin{figure}
\small
    \centering
    \includegraphics[width=0.49\textwidth]{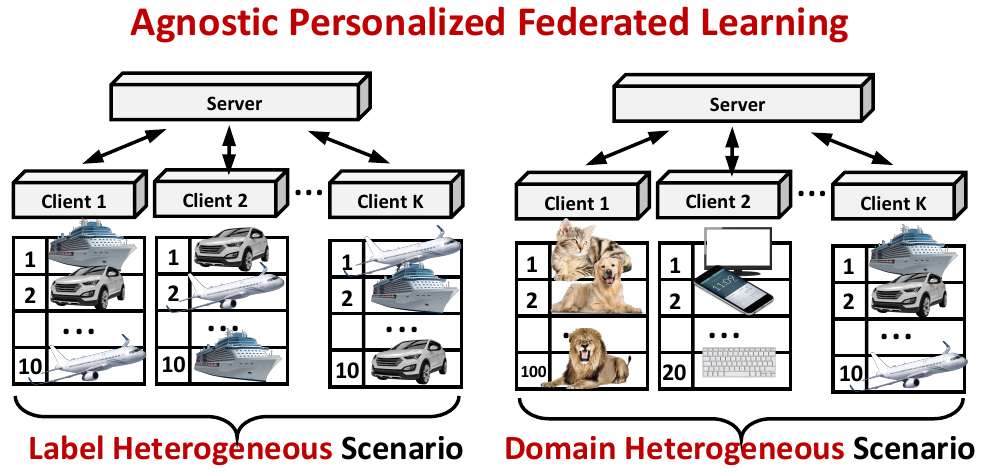} 
    \vspace{-0.15in}
    \caption{\small{\textbf{Agnostic Personalized Federated Learning Scenarios.}} \textbf{Left} labels are not synchronized across all clients. \textbf{Right} the local clients learn on different tasks and/or domains.}
    \label{fig:overview}
    \vspace{-0.25in}
\end{figure}


\begin{figure*}[t]
\small
\centering
\vspace{-0.05in}
\begin{tabular}{c c c}
    \small    
    \hspace{-0.15in}
    \includegraphics[width=0.39\textwidth]{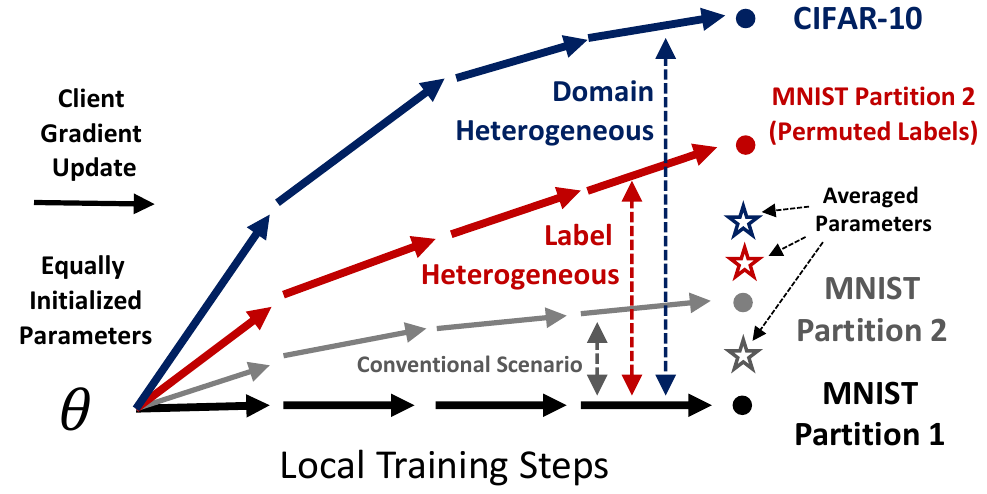} &
    \hspace{-0.2in} \includegraphics[width=0.3\textwidth]{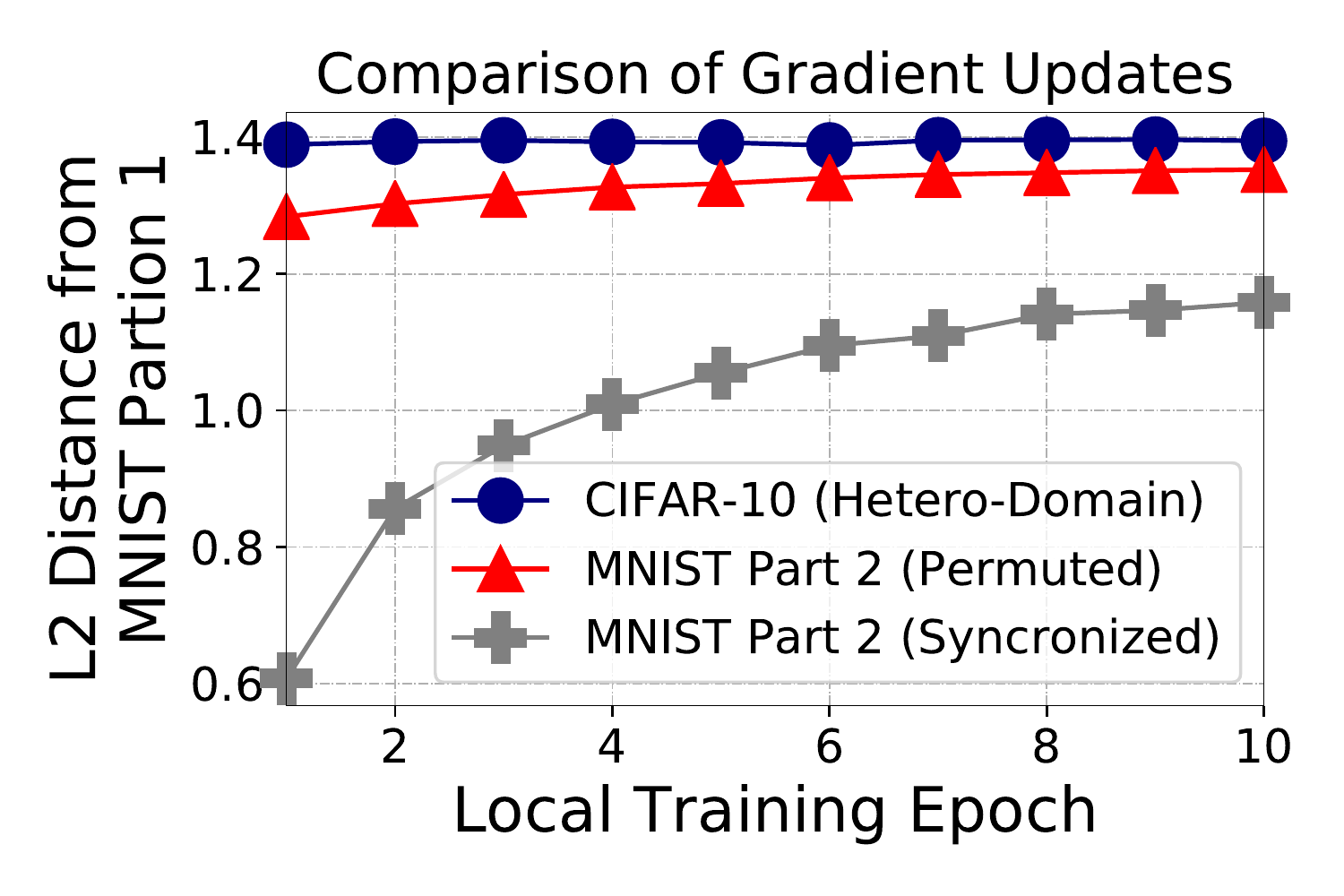} & 
    \hspace{-0.15in}
    \includegraphics[width=0.3\textwidth]{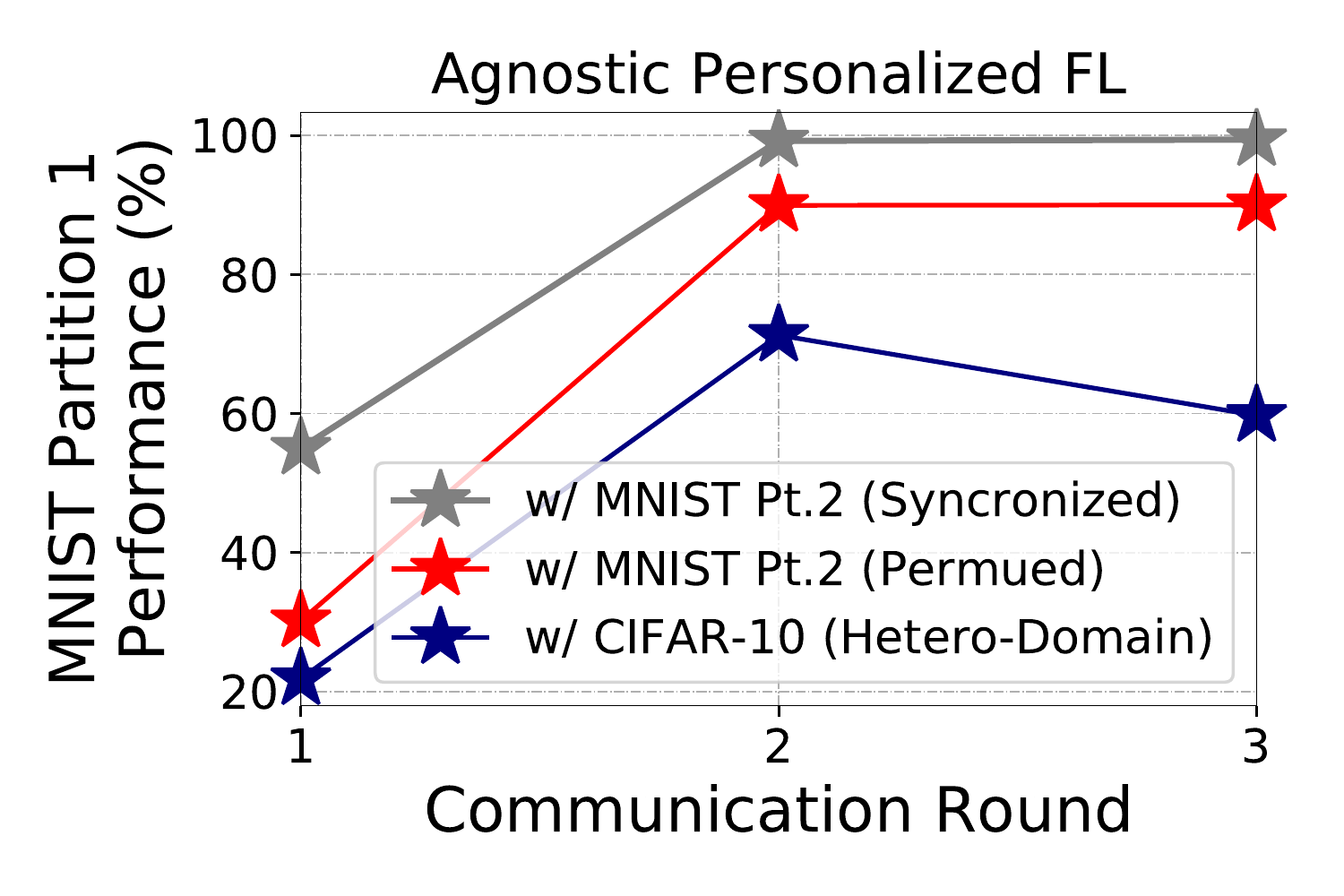}
    \\
    \hspace{-0.1in} (a) Illustration of the Parameter Space &
    \hspace{-0.1in} (b) $L_2$ Distance of Gradient Updates &
    \hspace{-0.1in} (c) Performance Degeneration

\end{tabular}
\vspace{-0.1in}
\caption{\small{\textbf{Challenges of Agnostic Personalized Federated Learning Scenarios}} (a) illustrates label and domain heterogeneity in parameter space. (b) shows the $L_2$ distance of the gradient updates from that of model trained on MNIST Partition 1. (c) shows performance degradation on MNIST partition 1 caused by the label and domain heterogeneity while performing federated learning. }
\label{fig:concept}
\vspace{-0.15in}
\end{figure*}


Personalized Federated Learning (PFL) aims to utilize the aggregated knowledge from other clients while learning a client-specific model that is specialized for its own task and data distribution, rather than learning a universal global model~\citep{arivazhagan2019federated,liang2020think,fallah2020personalized,zhang2021personalized}. While various personalized federated learning approaches have shown success in alleviating the data heterogeneity problem, yet, they are also limited as they follow the common assumptions of the standard federated learning setting, that (1) all participants use the same set of labels that are in the same order, and (2) all clients tackle the same task from the same domain.

In many real-world scenarios, the first assumption may not hold since the labels for the same task could be differently annotated depending on the user environment (Figure~\ref{fig:overview} Left). For example, when working with the same set of semantic classes, the labels across multiple clients could have a completely different ordering of the classes. In client 1, the label 1 may denote the ``Ship" class, while in client 2, the label 10 may denote the same class. Also, the same ``Car" class may be given the label ``Vehicle" or ``SUV". 

The second assumption severely limits the pool of devices that can participate in the collaborative learning process. However, clients working on different tasks and domains may have similar classes, or a common underlying knowledge, that may be helpful for the local models being trained at other clients (Figure~\ref{fig:overview} Right). Thus, it would be helpful if we can allow such domain-heterogeneous models to communicate the common knowledge across tasks and domains. 

However, federated learning under label and domain heterogeneity is a non-trivial problem, as most methods suffer from severe performance degeneration in such settings (Table~\ref{tbl:permuted}). We analyze this phenomenon in Figure~\ref{fig:concept}. Specifically, we train equally-initialized models on four different datasets and observe how different the gradient updates becomes as training goes on (we measure normalized $L_2$ distance of them). Learning on two MNIST partitions (split by an instance-wise manner) show the smallest difference (Figure~\ref{fig:concept} (a) and (b) Gray). Interestingly, simply permuting the labels of the partition 2 makes the gradient updates to largely diverge from the original gradients, which results in more severe heterogeneity compared to those of the the model trained with synchronized labels (Figure~\ref{fig:concept} (a) and (b) Red). Moreover, learning on completely different dataset (CIFAR-10) makes model gradients diverge more severely compared to learning on the dataset with permuted labels (Figure~\ref{fig:concept} (a) and (b) Blue). We conjecture that conventional loss function, i.e. cross entropy, and the corresponding back-propagation process do not actually care about task homogeneity, and thus it is not guaranteed that model parameters are identically updated when labels are permuted. These particularly lead to severe performance degeneration when performing federated learning (Figure~\ref{fig:concept} (c)). We measure performance on MNIST partition 1 while aggregating a model trained on the different dataset for every 3 epochs. We observe averaged models suffer from crucial performance degeneration in both  FL scenarios.

We name this challenging problem as the Agnostic Personalized Federated Learning (APFL) problem, where participants with personalized labels or from multiple domains can collaboratively learn while benefiting each other. An APFL problem has two critical challenges: (1) Label Heterogeneity for the discrepancy of the labels, due to the lack of a synchronized labeling scheme across the clients; and (2) Domain Heterogeneity for the discrepancy in the task and domains tackled by each participant. 

To tackle these challenges, we propose a novel method \texttt{Factorized-FL}, which factorizes model parameters into basis vectors and aggregate them in the factorized parameter space. This allows to factorize the the model aggregation to take place in a semantic parameter basis space which is more robust to the use of different labels. Also, the factorization results in the separation of the client-general and client-specific knowledge, and thus prevents the aggregation of incompatible knowledge across clients. Moreover, to further alleviate the model from collapsing into a degenerate solution, we measure the task similarity across the clients using the factorized parameters, to allow selective aggregation of the knowledge among the relevant models that work on similar tasks or domains. We extensively validate our method on both label- and domain-heterogeneous settings, and show that our method significantly outperforms the current state-of-the-art personalized federated learning methods. This work can be summarized as follows:

\vspace{-0.05in}
\begin{itemize}[leftmargin=0.2in]
    \vspace{-0.1in}    
	\item We introduce Agnostic Personalized Federated Learning (APFL) and study its two critical challenges, Label and Domain Heterogeneity.
	\vspace{-0.075in}
	\item We propose a novel FL method named Factorized-FL, which factorizes model parameters to reduce parameter dimensionality for alleviating knowledge collapse, and utilize task-level similarity for matching relevant clients.
	\vspace{-0.2in}
	\item We extensively validate our method in both label- and domain-heterogeneous scenarios and show our method outperforms the current state-of-the-art methods.
\end{itemize}

\vspace{-0.15in}
\section{Related Work}

\paragraph{Federated learning} A variety of algorithms have been proposed for federated learning since the introduction of \texttt{FedAvg}~\cite{McMahan2017CommunicationEfficientLO}, but, we specifically focus on works that aim to tackle the heterogeneity problems, e.g. Non-IID. Some studies focus on regularization methods~\citep{mohri2019agnostic,li2020federated}, correcting disparity between server and clients ~\cite{Wang2020Federated,karimireddy2021scaffold}, or contrastive learning~\citep{li2021modelcontrastive}. While we mostly consider the task-level heterogeneity problem in this paper, many existing works also tackle architecture-level heterogeneity~\citep{seo2020federated,zhu2021datafree,diao2021heterofl,shamsian2021personalized}.


\textbf{Personalized federated learning} aims to improve the individual local clients instead of learning the universal global model via the mixture methods~\citep{mansour2020approaches,deng2020adaptive,hanzely2021federated},   meta-learning approaches~\citep{fallah2020personalized}, or partial network aggregation~\citep{arivazhagan2019federated,liang2020think}. Recent approaches avoid aggregating irrelevant other clients that is not helpful. \citet{zhang2021personalized} downloads and evaluate other clients locally to aggregate only beneficial clients. \citet{sattler2019clustered,duan2021fedgroup} measure client-wise similarity by using the gradient updates. Our method also measures client similarity but in a more efficient and effective way, simply utilizing a factorized vector.

\textbf{Re-parameterization for federated learning} \citet{jeong2021federated,yoon2021federated} decompose model parameters (use an additional set of parameters) to train them with different objectives, which do not reduce the dimensionality of model parameters. Some approaches factorize high dimensional model parameters into lower dimensional space, i.e. low rank matrices.~\citet{konevcny2016federated} introduces structured update which model directly learns 
factorized parameter space. ~\citet{anonymous2022fedpara} propose to use the Hadamard product of low rank matrices to enhance communication efficiency. Unlike prior works, we utilize rank-1 vectors which separately capture task-general and the client-specific knowledge for the extremely heterogeneous FL scenarios without losing expressiveness via sparse bias matrices.

\section{Problem Definition}
\label{sec:def}
We begin with the formal definition of the conventional federated learning scenario, and then introduce our novel Agnostic Personalized Federated Learning (APFL) problem. 


\subsection{Preliminaries}
\label{subsec:fl}
Our main task is solving a given multi-class classification problem in an FL framework. Let $f_g$ be a global model (neural network) at the global server and $\mathcal{F}=\{f_k\}^{K}_{k=1}$ be a set of $K$ local neural networks, where $K$ is the number of local clients.  $\mathcal{D}=\{\textbf{x}_{i}, y_{i}\}^{N}_{i=1}$ be a given dataset, where $N$ is the number of instances, $\textbf{x}_i \in \mathbb{R}^{W \times H \times D}$ is the $i_{th}$ examples in a size of width $W$, height $H$, and depth $D$, with a corresponding target label $y_i \in \{1,\dots,C\}$ for the $C$-way multi-class classification problem. The given dataset $\mathcal{D}$ is then disjointly split into $K$ sub-partitions $\mathcal{P}_{k}=\{\textbf{x}_{k,i}, y_{k,i}\}^{N_{k}}_{i=1}$ s.t. $\mathcal{D} = \bigcup_{k=1}^{K} \mathcal{P}_{k}$, which are distributed to the corresponding local model $f_k$. Let $R$ be the total number of the communication rounds and $r$ denote the index of the $r_{th}$ communication round. At the first round $r$=$1$, the global model $f_g$ initialize the global weights $\theta^{(1)}_{f_g}$ and broadcasts $\theta^{(1)}_{f_g}$ to an arbitrary subset of local models that are available for training at round $r$, such that $\mathcal{F}^{(r)}\subset\mathcal{F}$, $|\mathcal{F}^{(r)}|=K^{(r)}$, and $K^{(r)} \leq K$, where $K^{(r)}$ is the number of available local models at round $r$. Then the active local models $f_{k}\in\mathcal{F}^{(r)}$ perform local training to minimize loss $\mathcal{L}( \theta^{(r)}_{k})$ on the corresponding sub-partition $\mathcal{P}_{k}$ and update their local weights $\theta^{(r+1)}_{k} \leftarrow \theta^{(r)}_{k}-\eta\nabla\mathcal{L}(\theta^{(r)}_{k})$, where $\theta^{(r)}_{k}$ is the set of weights for the local model $f_k$ at round $r$ and $\mathcal{L}(\cdot)$ is the loss function. When the local training is done, the global model $F$ collects and aggregates the learned weights $\theta^{(r+1)}_{f_g} \leftarrow \frac{N_{k}}{N}\sum_{i=1}^{K^{(r)}} \theta_{k}^{(r)}$ and then broadcasts newly updated weights to the local models available at the next round $r+1$. These learning procedures are repeated until the final round $R$. This is the standard setting for centralized federated learning, which aims to find a single global model that works well across all local data. On the other hand, Personalized Federated Learning aims to adapt the individual local models $f_{1:K}$ to their local data distribution $\mathcal{P}_{1:K}$, to obtain specialized solution for each task at the local client, while utilizing the knowledge from other clients. Thus merging the local knowledge for personalized FL is not necessarily done in the form of $\theta^{(r+1)}_{f_g} \leftarrow \frac{N_{k}}{N}\sum_{i=1}^{K^{(r)}} \theta_{k}^{(r)}$, and the specific ways to utilized the knowledge from others depends on the specific algorithm, i.e. $\theta^{(r+1)}_k\leftarrow \theta^{(r)}_k + \sum_{i\neq k}^{K^{(r)}} \omega_i(\theta_{k}^{(r)}-\theta_{i}^{(r)})$, wher $\omega(\cdot)$ is weighing function ~\citep{zhang2021personalized}.

\begin{figure*}[t]
\small
    \centering
    \includegraphics[width=\textwidth]{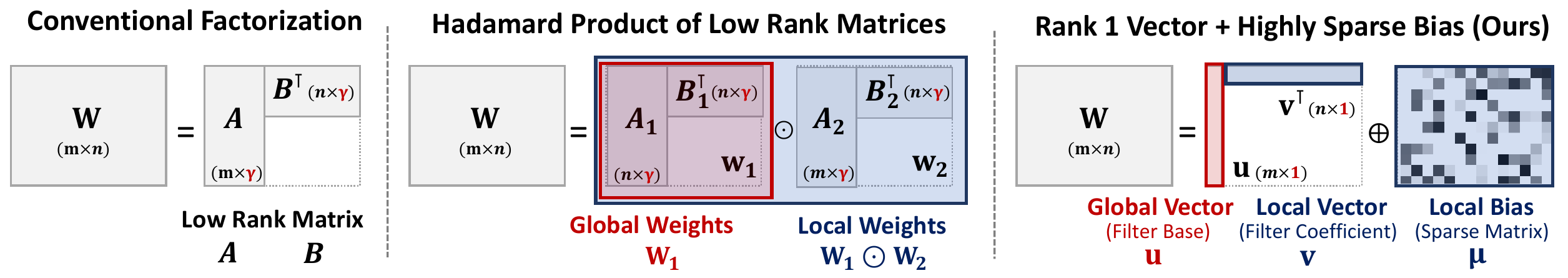} 
    \vspace{-0.25in}
    \caption{\small{\textbf{Illustration of Parameter Factorization Methods:} Left shows conventional matrix factorization with two low rank matrices with rank $\gamma$. Middle represents the method utilizing Hadamard product of low rank matrix for federated learning~\citep{anonymous2022fedpara}. Right illustrates our factorization method for agnostic personalized federated learning, which utilizes rank 1 vectors and highly sparse bias.} }
    \label{fig:factorize}
    \vspace{-0.15in}
\end{figure*}

\begin{figure}[t]
\small
\centering
\vspace{-0.05in}
\begin{tabular}{c c}
    \small    
    
    \hspace{-0.15in} \includegraphics[width=0.24\textwidth]{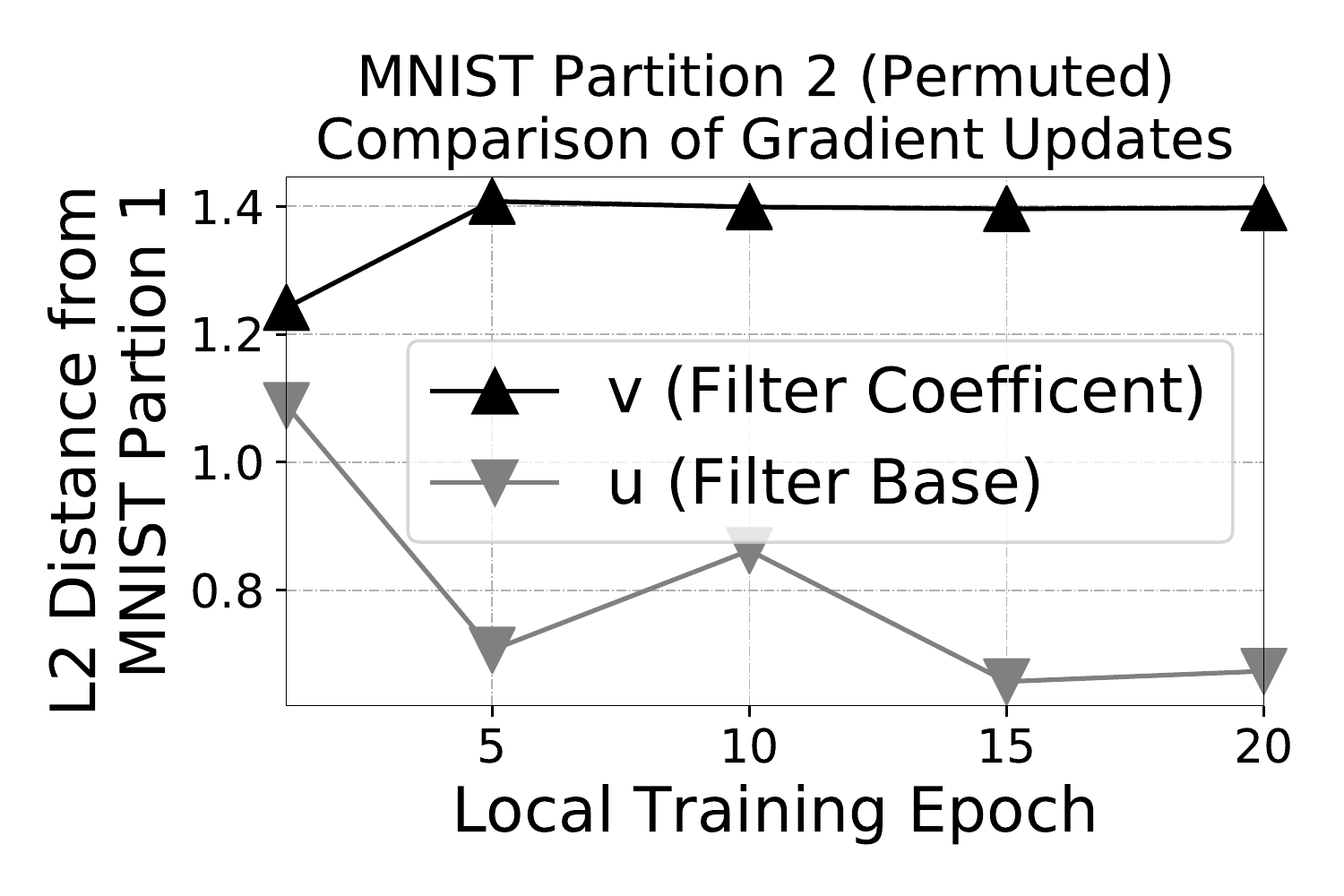} & 
    \hspace{-0.225in} \includegraphics[width=0.24\textwidth]{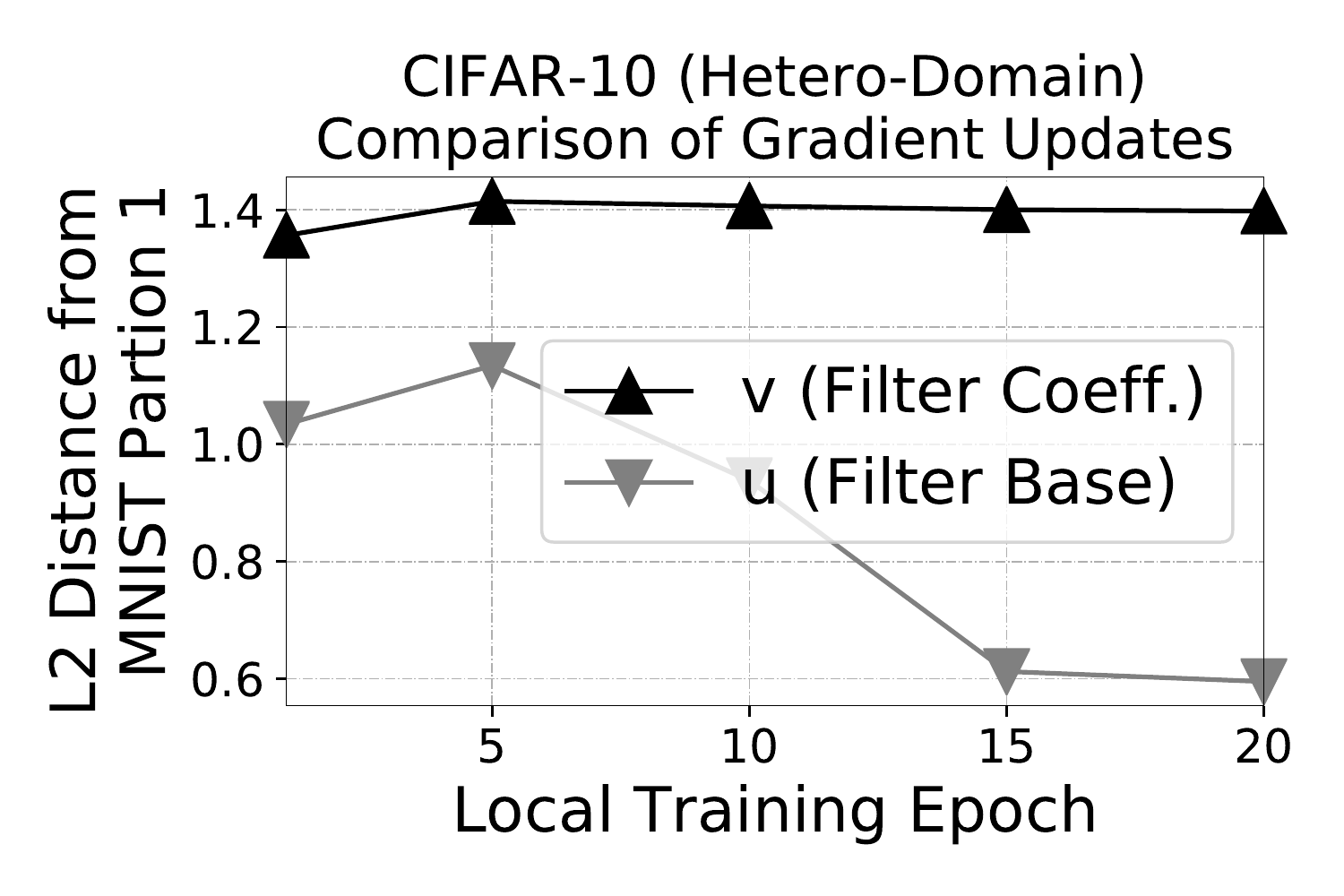}
    \\
    \hspace{-0.1in} (a) MNIST Part. 2 (Permuted) &
    \hspace{-0.125in} (b) CIFAR-10 (Hetero-Domain)
\end{tabular}
\vspace{-0.1in}
\caption{\small{\textbf{Analysis of $\textbf{u}$ and $\textbf{v}$:}} We plot normalized $L_2$ distance of the gradient updates of factorized parameters $\textbf{u}$ and $\textbf{v}$ while learning on (a) MNIST Partition 2 and (b) CIFAR-10 compared to learning on MNIST Partition 1. }
\label{fig:f_analysis}
\vspace{-0.25in}
\end{figure}

\subsection{Agnostic Personalized Federated Learning}
\label{subsec:apfl}

Agnostic Personalized Federated Learning (APFL) is a scenario where any local participants from diverse domains with their own personalized labeling schemes can collaboratively learn, benefiting each other. There exist two critical challenges that need to be tackled to achieve this objective: (1) Label Heterogeneity and (2) Domain Heterogeneity.

\vspace{-0.125in}
\paragraph{Label Heterogeneity } This scenario assumes that the labeling schemes are not perfectly synchronized across all clients, as described in Section~\ref{sec:intro} and Figure~\ref{fig:overview} Left. Most underlying setting for this scenario is the same as the conventional single-domain setting with synchronized labels that is described in Section~\ref{subsec:fl}, except that labels are arbitrarily permuted amongst clients. The local data $\mathcal{P}_{k}$ for the local model $f_k$ is now defined as $\mathcal{P}_{k}=\{\textbf{x}_{k,i}, \varphi_{k}(y_{k,i})\}^{N_{k}}_{i=1}$, where $\varphi_{k}(\cdot)$ is a mapping function for the local model $f_k$ which maps a given class $y_{k,i}$ with a randomly permuted label $p_{k,i}$=$\varphi_{k}(y_{k,i})$. Let the $j$th layer out of $L$ layers in the neural networks of local model $f_k$ be $\ell_{k}^{j}$ and the last layer $\ell_{k}^L$ be the classifier layer. Since each client has differently permuted labels, the personalized classifiers $\ell_{1:K}^{L}$ are no longer compatible to each other. While we can merge the layers below the classifier in this setting, training with heterogeneous labels could still lead to large disparity in the local gradients even in the initial communication round, as described in Figure \ref{fig:concept}.

\vspace{-0.125in}
\paragraph{Domain Heterogeneity} This scenario presumes that local clients learn on their own dataset $\mathcal{D}$, that are completely different from the datasets that are used at other clients, as described in Section~\ref{sec:intro} and Figure~\ref{fig:overview} Right. In this setting, $K$ disjoint datasets $\mathcal{D}_{1:K}$ are assigned to the $K$ local clients $f_{1:K}$, where $\mathcal{D}_k=\{\textbf{x}_{k,i}, y_{k,i}\}^{N_k}_{i=1}$ is the dataset assigned to the local model $f_k$. The number of target classes may differ across clients, such that $y_{k,i} \in \{1,\dots,C_k\}$. We assume complete disjointness across clients, such that there is no instance-wise and class-wise overlap across the datasets: $\varnothing =\bigcap_{k=1}^K \mathcal{D}_k$. Similarly to the label-heterogeneous scenario described above, the personalized classifiers $\ell_{1:K}^{L}$ are no longer compatible to each other due to the heterogeneity in the data and the labels. Hence, the aggregation is done for the layers before the classifier, but they will be also incompatible as the learned model weights will be largely different across domains.


\section{Factorized Federated Learning}
\label{sec:method}

We now provide detailed descriptions of our novel algorithm \texttt{Factorized-FL}. 

\subsection{Kernel Factorization}
\label{subsec:facto}

\citet{Wang2020Federated} discussed that the conventional knowledge aggregation, that is often performed in a coordinate-wise manner, may have severe detrimental effects on the averaged model. This is because the deep neural networks have extremely high-dimensional parameters and thus meaningful element-wise neural matching is not guaranteed when aggregating the weights across different models trained under diverse settings. 

\begin{figure}
\small
    \centering
    \includegraphics[width=0.41\textwidth]{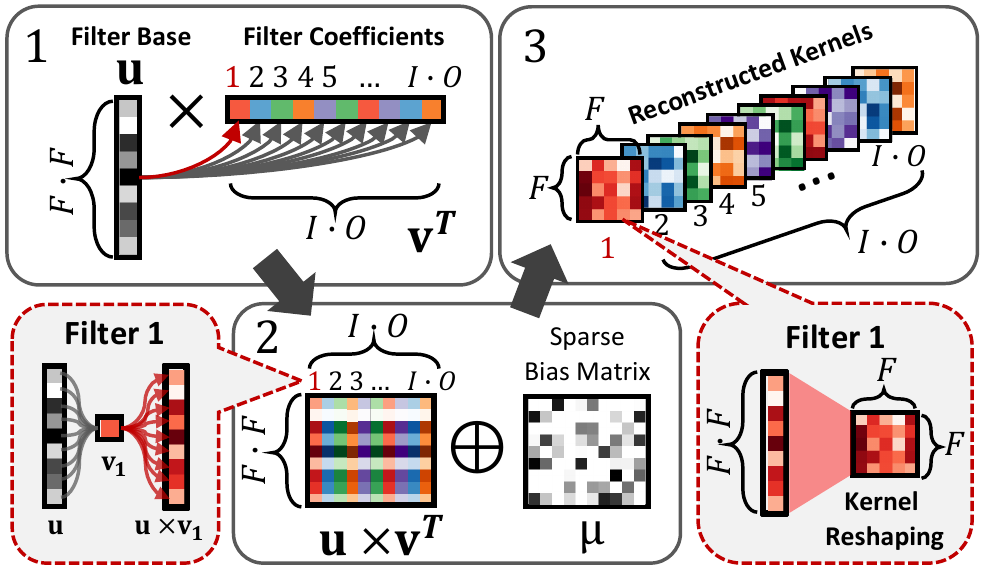} 
    \vspace{-0.05in}
    \caption{\small{\textbf{Illustration of Kernel Factorization \& Reconstruction}} (1) we multiply two factorized vectors $\textbf{u}$ and $\textbf{v}$ to obtain kernel matrix. (2) we add the sparse bias matrix $\mu$ to complement non-linearity. (3) we reshape the matrix into the original kernel shape. }
    \label{fig:factorize_detail}
    \vspace{-0.3in}
\end{figure}

One naive solution to this problem is to factorize model parameters into lower dimensional space, i.e. low rank matrices, as shown in Figure~\ref{fig:factorize} (Left). Conventional approaches, such as SVD, Tucker, or Canonical Polyadic decomposition, however, factorize model parameters after training~\citep{lebedev2014speeding,phan2020stable} is done. Thus, the dimensionality at the time of knowledge aggregation will remain the same as the unfactorized model. ~\citet{konevcny2016federated,anonymous2022fedpara} pre-decompose model parameters to low rank matrices for FL scenarios. While~\citet{konevcny2016federated} use naive low rank matrices, ~\citet{anonymous2022fedpara} uses two sets of low rank matrices to improve expressiveness and utilize them as global and local weights (Figure~\ref{fig:factorize} (Middle)). Unlike prior works, our approach utilizes rank-1 vectors to perform aggregation in the lowest subspace possible for compatibility, while effectively yet efficiently enhancing expressiveness with sparse bias matrices, as shown in Figure~\ref{fig:factorize} (Right) and Figure~\ref{fig:factorize_detail}. Another crucial difference of our method from the previous factorization methods is that, our rank-1 vectors have distinct roles. Our factorization will separate the common knowledge from the task- or domain-specific knowledge, since $\textbf{u}$ could be thought as the bases  (the common knowledge across clients) and $\textbf{v}$ could be thought as the coefficients (client-specific information). 

In Figure~\ref{fig:f_analysis} (a) and (b), which shows the experimental results with the factorized model, we observe that $\textbf{u}$ trained on two datasets becomes closer to that of another dataset (MNIST Partition 1) while $\textbf{v}$ (personalized filter coefficient) remain largely different as federated learning goes on. With this observation, we further aggregate $\textbf{u}$ while allowing $\textbf{v}$ to be different across clients, to allow personalized FL. Further, we use the client specific $\textbf{v}$ for similarity matching, to identify relevant local models from other clients. In following paragraphs, we describe our factorization method in detail, for both fully-connected and convolutional layers. \vspace{-0.15in}

\paragraph{Factorization of Fully-Connected Layers} We assume that each local model $f_k$ has a set of local weights $\theta_k$ across all layers; that is, $\theta_k=\{\textbf{W}^{i}_k\}^{L}_{i=1}$. The dimensionality of the dense weight $\textbf{W}_k^{i}$ for each fully connected layer is $\textbf{W}_k^{i} \in \mathbb{R}^{I \times O}$, where $I$ and $O$ indicate respective input and output dimensions. We can reduce the $I \times O$ complexity by factorizing the high order matrix into the outer product of two vectors as follows:
\begin{equation}
\begin{split}
\textbf{W}_k^{i} = \textbf{u}_k^{i} \times \textbf{v}_k^{i \intercal}, \text{where } \textbf{u}_k^{i} \in \mathbb{R}^{I}, \textbf{v}_k^{i} \in \mathbb{R}^{O}
\end{split}
\end{equation}
 However, such extreme factorization of the weight matrices may result in the loss of expressiveness in the parameter space. Thus, we additionally introduce a highly sparse bias matrix $\mu$ to further capture the information not captured by the outer product of the two vectors as follows:
\begin{equation}
\begin{split}
\textbf{W}_k^{i} = \textbf{u}_k^{i} \times \textbf{v}_k^{i \intercal} \oplus {\mu}_k^{i}, 
\text{where } \\ \textbf{u}_k^{i} \in \mathbb{R}^{I},  
\textbf{v}_k^{i} \in \mathbb{R}^{O}, {\mu}_k^{i} \in \mathbb{R}^{I \times O} 
\end{split}
\end{equation}
We initialize $\mu$ with zeros so that it can gradually capture the additional expressiveness that are not captured by $\textbf{u}$ and $\textbf{v}$ during training. We can control its sparsity by the hyper-parameter for the sparsity regularizer described in~\ref{subsec:algo}.

\vspace{-0.05in}
\paragraph{Factorization of Convolutional Layers} The difference between the fully-connected and convolutional layers is that the convolutional layers have multiple kernels (or filters) such that $\textbf{W}_k^{i} \in \mathbb{R}^{F \times F \times I \times O}$, where $F$ is a size of filters (we assume the filter size is equally paired for the simplicity). To induce $\textbf{u}$ to capture base filter knowledge and $\textbf{v}$ to learn filter coefficient, it is essential to design $\textbf{u}\in \mathbb{R}^{F \cdot F}$ and $\textbf{v}\in \mathbb{R}^{I \cdot O}$, but not in arbitrary ways, such as $\textbf{u}\in \mathbb{R}^{I \cdot F}$ and $\textbf{v}\in \mathbb{R}^{O \cdot F}$ or $\textbf{u}\in \mathbb{R}^{O}$ and $\textbf{v}\in \mathbb{R}^{I \cdot F \cdot F}$. We observe that performance is degenerated when the parameters are ambiguously factorized  (Figure~\ref{fig:analysis} (h)). Our proposed factorization method for convolutional layers are as follows:
\vspace{-0.1in}
\begin{equation}
\begin{split}
\textbf{W}_k^{i} = \pi(\textbf{u}_k^{i} \times \textbf{v}_k^{i \intercal} \oplus {\mu}_k^{i}), \text{where }  \textbf{u}_k^{i} \in \mathbb{R}^{F \cdot F }, \textbf{v}_k^{i} \in \mathbb{R}^{I \cdot O }, \\
{\mu}_k^{i} \in \mathbb{R}^{F \cdot F \times I \cdot O}, \pi(\cdot): \mathbb{R}^{F \cdot F \times I \cdot O} \rightarrow \mathbb{R}^{F \times F \times I \times O},
\end{split}
\end{equation}
$\pi(\cdot)$ is the weight reshaping function. Note that we reparameterize our model \textit{at initialization time}. Then we reconstruct and train full weights of each layer $\textbf{W}_k^{1:L}$, while optimizing $\textbf{u}_k^{1:L}$, $\textbf{v}^{1:L}_k$, and ${\mu}^{1:L}_k$, respectively, during training phase.

\subsection{Similarity Matching}
\label{subsec:sim}

\begin{figure}[t]
\small
    \centering
    \includegraphics[width=0.49\textwidth]{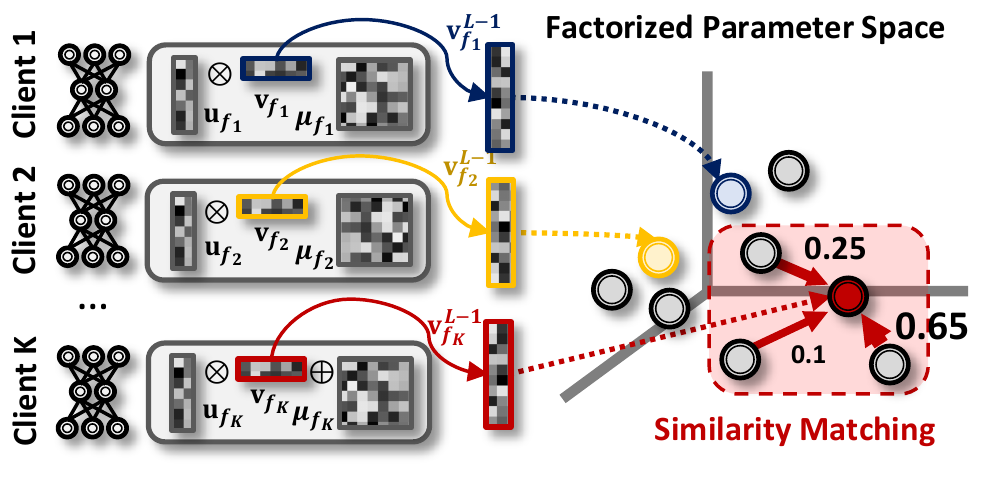} 
    \vspace{-0.3in}
    \caption{\small{\textbf{Illustration of Similarity Matching:}} We match relevant clients utilizing the factorized vector $\textbf{v}$ that captures client-specific knowledge. Then we aggregate $\textbf{u}$ based on the similarity. }
    \label{fig:framework}
    \vspace{-0.2in}
\end{figure}

Since we assume task- and domain-heterogeneous FL scenarios, aggregating the parameter bases across all clients may not be optimal, since some of them could be highly irrelevant. \citet{yoon2021federated} and \citet{zhang2021personalized} also demonstrated that avoiding aggregation of irrelevant models from other clients improves local model performance. \citet{yoon2021federated} achieve this goal by taking the weighted combination of task-specific weights from other clients, and \citet{zhang2021personalized} suggest downloading the models from other clients and evaluating their performance on a local validation set, at each client. However, since they require additional communication and computing cost at the local clients, we provide a more efficient yet effective approach to find and match models that are beneficial to each other. 

\vspace{-0.1in}
\paragraph{Efficient similarity matching} Our method utilizes factorized vector $\textbf{v}_k$ for measuring similarity across different models, at the central server. Since $\textbf{v}$ are devised to learn personalized coefficient, we assume that clients trained on similar task or domain will have similar $\textbf{v}$. Specifically, we only use $\textbf{v}^{L-1}_{k}$ of the second last layer (before classifier layer) for similarity matching. The similarity matching function $\Omega(\cdot)$, is defined as the cosine similarity between target client $f_k$ and the other clients $\{f_i\}_{i\neq k}^K$, as follows:
\begin{equation}
\label{eq:sim}
\begin{split}
\Omega(\textbf{v}_{f_k}^{L-1}, \textbf{v}_{f_{i\neq k:K}}^{L-1}) = \{\sigma_i |  
\sigma_i = \frac{\textbf{v}_{f_k} \cdot \textbf{v}_{f_i}}{\|\textbf{v}_{f_k}\|\|\textbf{v}_{f_i}\|}
, \sigma_i \geq \tau \}_{i \neq k}^{K}
\end{split}
\end{equation}
The similarity scores for those with the cosine similarity scores lower than the given threshold $\tau$, are set to zero.
Our method is significantly more efficient than similarity matching approaches which use full gradient updates for clustering clients~\citep{sattler2019clustered,duan2021fedgroup}.

\vspace{-0.1in}
\paragraph{Personalized weighted averaging} We allow each local model to perform weighted aggregation of the model weights from other clients, utilizing their similarity scores:
\begin{equation}
\begin{split}
\textbf{u}_{k}^l \leftarrow \frac{\text{exp}({\epsilon \cdot \sigma_i})}{\sum_{i=1}^K \text{exp}(\epsilon \cdot \sigma_i)} \sum_{i=1}^{K} \textbf{u}_i^l, s.t. \forall  l \in \{1,2,\dots,L\}
\end{split}
\end{equation}
where $\epsilon$ is a hyperparameter for scaling the similarity score $\sigma_i$. We always set $\sigma_k$, the similarity score for itself, as $1.0$. 

\subsection{Learning Objective}
\label{subsec:algo}

Now we describe our final learning objective. Instead of utilizing the single term $\theta_k$ for local weights of neural network $f_k$, now let $\mathcal{U}_k$, $\mathcal{V}_k$, and $\mathcal{M}_k$ be sets of $\textbf{u}_k$, $\textbf{v}_k$, and $\mu_k$ of all layers in $f_k$, s.t. $\mathcal{U}_k=\{\textbf{u}^{i}_k\}^{L}_{i=1}$, $\mathcal{V}_k=\{\textbf{v}^{i}_k\}^{L}_{i=1}$, and $\mathcal{M}_k=\{{\mu}^{i}_k\}^{L}_{i=1}$, then our local objective function is,
\begin{equation}
\begin{split}
\min_{\mathcal{U}_k,\mathcal{V}_k,\mathcal{M}_k} \sum_{\mathcal{B} \in \mathcal{D}_k} \mathcal{L}(\mathcal{B}; \mathcal{U}_k, \mathcal{V}_k, \mathcal{M}_k) + \lambda_{\text{sparsity}} ||\mathcal{M}_k||_1,
\end{split}
\label{eq:loss}
\end{equation}
where $\mathcal{L}$ is the standard cross-entropy loss performed on all minibatch $\mathcal{B} \in \mathcal{D}_k$. We add the $L_1$ sparsity inducing regularization term to make the bias parameters highly sparse, controlling its effect with a hyperparameter $\lambda_{\text{sparsity}}$. Please see our pseudo-coded algorithm Algorithm 1 in Appendix~\ref{appdx:algorithm}.

\section{Experiment}
\label{sec:exp}
We validate our method on label- and domain-heterogeneous FL scenarios, against relevant baselines.

\subsection{Experimental Setup}
\label{subsec:setup}

\paragraph{Models} We first consider well-known baseline FL methods, such as (1) \texttt{FedAvg}~\citep{McMahan2017CommunicationEfficientLO} and (2) \texttt{FedProx}~\citep{li2018federated}. We evaluate our factorization technique with (3) \texttt{pFedPara}~\citep{anonymous2022fedpara} which also uses kernel factorization technique for personalized FL scenario. Our similarity matching approach is compared to (4) \texttt{Clustered-FL}~\citep{sattler2019clustered} and (5) \texttt{FedFOMO}~\citep{zhang2021personalized}, which measuring client-wise similarity or helpfulness. (6) \texttt{Per-FedAvg}~\citep{fallah2020personalized} is also used for evaluation as it shows great performance on heterogeneous federated learning scenarios. We also show local training model, (7) \texttt{Stand-Alone}, for the lower bound performance. We introduce an additional variant of \texttt{Factorized-FL}, which aggregates not only the parameter bases across all client, but also coefficient and bias terms well. Particularly, for the standard FL scenarios, i.e. iid or non-iid, where label- and domain-heterogeneity does not exist, aggregation of separately learned knowledge can further effectively improve local performance. We name such model \texttt{Factorized-FL $\beta$}. Please see the Appendix~\ref{appdx:details} for detailed implementation and training details.

\vspace{-0.1in}
\paragraph{Datasets} (1) \textit{Label Heterogeneous Scenario}: we use CIFAR-10 and SVHN datasets and we create four different partitions for each dataset, which are conventional iid and non-iid as well as permuted iid and permuted non-iid, which labels are permuted and incompatible to each other. We split the datsets into $20$ partitions and then simply permuted the labels on the same partition for label permuted settings. (2) \textit{Domain Heterogeneous Scenario}: we use CIFAR-100 datasets and create five sub-datasets grouped by $10$ similar classes, such as Household Objects, Fruits\&Foods, Trees\&Flowers, Transport, and Animals. We assign $4$ clients for each sub-dataset, i.e. Client 1-4 to Household Objects, Client 5-8 to Fruits\&Foods, Client 9-12 to Trees\&Flowers, Client 13-16 to Transport, and Client 17-20 to Animals. We then permute the labels for all partitions to simulate further realistic scenarios. Further descriptions are elaborated in Appendix~\ref{appdx:dataset}.

\subsection{Experimental Result}
\label{subsec:result}

\paragraph{Label-heterogeneous FL} 
As shown in Table~\ref{tbl:permuted} (Top), for the standard IID and Non-IID settings, all FL methods obtain higher performance than the local training baseline (\texttt{Stand-Alone}), which confirms that the locally learned knowledge is beneficial to others, when the data and label distributions are homogeneous across clients. However, when the labels are not synchronized across all clients (Permuted IID/Non-IID), all previous FL methods achieve significantly degenerated performance, even lower than that of the local training baseline. Again, note that we do not share the classifier layers to ensure fairness across all algorithms in this permuted settings. We conjecture that this is caused by the label permutation leading the local model to evolve a permuted set of features that are not coordinate-wise compatible to others when aggregated. Contrarily, our method \texttt{Factorized-FL} shows consistent performance regardless of whether labels are permuted or not. \texttt{Factorized-FL $\beta$} even largely outperforms all baseline models with significantly superior performance. Test accuracy curves over communication round and transmission cost are visualized in Appendix~\ref{appdx:experiment}.

\begin{table*}[t]

\small
\caption{\textbf{Performance comparison of label and domain heterogeneous scenario.} \textbf{Top (label heterogeneous scenario):} we train $20$ clients on each dataset for $250$ (CIFAR-10 \& SVHN) training iterations ($E$=$5$, $R$=$50$). \textbf{Bottom (domain \& label heterogeneous scenario):} We train $20$ clients for $500$ training iterations ($E$=$5$,$R$=$100$) on $20$ sub-datasets from $5$ heterogeneous domains ($4$ partitions per domain). Labels are also permuted for all partitions.  We measure averaged performance over three trials with different seeds.}
\label{tbl:permuted}
\vspace{-0.1in}

\resizebox{\textwidth}{!}{
\begin{tabular}{clccccccccc}
\toprule
\midrule
\multirow{2}{*}{ \textbf{Dataset}} &  \multirow{2}{*}{\textbf{Method}} & 
\multicolumn{2}{c}{\textbf{Standard IID}} & \multicolumn{2}{c}{\textbf{Permuted IID}} & 
& 
\multicolumn{2}{c}{\textbf{Standard Non-IID}} & \multicolumn{2}{c}{\textbf{Permuted Non-IID}} \\

& & 
Accuracy [\%]  &  Cost [Gb]  & 
Accuracy [\%]  &  Cost [Gb]  &  
& 
Accuracy [\%]  &  Cost [Gb]  & 
Accuracy [\%]  &  Cost [Gb] \\

\midrule

\multirow{8}{*}{CIFAR-10} & \texttt{Stand-Alone} & 
{64.31} \scriptsize($\pm$ 1.08) & - & 
{63.93} \scriptsize($\pm$ 0.90) & - & 
& 
{47.79} \scriptsize($\pm$ 0.91) & - &
{46.06} \scriptsize($\pm$ 1.03) & - \\

& \texttt{FedAvg}~\citep{McMahan2017CommunicationEfficientLO} & 
{70.28} \scriptsize($\pm$ 0.82) & 20.39 & 
{65.31} \scriptsize($\pm$ 1.28) & 20.39 &  
& 
{53.08} \scriptsize($\pm$ 1.4) & 20.39  &
{48.90} \scriptsize($\pm$ 1.25) & 20.39 \\

& \texttt{FedProx}~\citep{li2018federated} & 
{70.54} \scriptsize($\pm$ 0.73) & 20.39 & 
{66.28} \scriptsize($\pm$ 0.90) & 20.39 &   
& 
{53.56} \scriptsize($\pm$ 0.55) & 20.39 &
{47.86} \scriptsize($\pm$ 0.83) & 20.39 \\

& \texttt{Clustered-FL}~\citep{sattler2019clustered} & 
{69.48} \scriptsize($\pm$ 1.02) & 20.39 & 
{65.77} \scriptsize($\pm$ 1.03) & 20.39 &
& 
{53.93} \scriptsize($\pm$ 1.57) & 20.39 &
{49.00} \scriptsize($\pm$ 0.32) & 20.39 \\

& \texttt{Per-FedAvg}~\citep{fallah2020personalized} & 
{70.84} \scriptsize($\pm$ 1.01) & 20.39 & 
{65.58} \scriptsize($\pm$ 0.74) & 20.39 & 
& 
{53.35} \scriptsize($\pm$ 2.87) & 20.39 &
{47.60} \scriptsize($\pm$ 1.01) & 20.39 \\

& \texttt{FedFOMO}~\citep{zhang2021personalized} & 
{70.19} \scriptsize($\pm$ 0.79) & 122.33 & 
{64.26} \scriptsize($\pm$ 0.92) & 122.33 & 
& 
{50.69} \scriptsize($\pm$ 1.61) & 122.33 &
{46.73} \scriptsize($\pm$ 1.04) & 122.33 \\

& \texttt{pFedPara}~\citep{anonymous2022fedpara} & 
{67.96} \scriptsize($\pm$ 1.25) &   7.4 & 
{65.12} \scriptsize($\pm$ 1.27) & 7.4  & 
& 
{55.88} \scriptsize($\pm$ 1.28) &  7.4 &
{50.22} \scriptsize($\pm$ 0.92) & 7.4 \\

\cmidrule{2-11}
& \texttt{Factorized-FL} (Ours) & 
{66.97} \scriptsize($\pm$ 1.36) &  \textbf{0.32}  & 
{67.91} \scriptsize($\pm$ 1.08) & \textbf{0.32}  & 
& 
{50.34} \scriptsize($\pm$ 1.33) & \textbf{0.32} &
{50.24} \scriptsize($\pm$ 1.03) & \textbf{0.32} \\

& \texttt{Factorized-FL $\beta$} (Ours) & 
\textbf{{76.26} \scriptsize($\pm$ 1.05)} &  18.25 & 
\textbf{{70.59} \scriptsize($\pm$ 2.07)} & 18.25 & 
& 
\textbf{{65.30} \scriptsize($\pm$ 1.38)} & 18.25 &
\textbf{{56.61} \scriptsize($\pm$ 1.10)} & 18.25 \\

\midrule
\midrule

\multirow{8}{*}{SVHN} & \texttt{Stand-Alone} &  
{84.18} \scriptsize($\pm$ 0.37) & - & 
{84.32} \scriptsize($\pm$ 0.31) & - &
& 
{62.50} \scriptsize($\pm$ 0.84) & - &
{62.11} \scriptsize($\pm$ 0.78) & - \\

& \texttt{FedAvg}~\citep{McMahan2017CommunicationEfficientLO} & 
{88.53} \scriptsize($\pm$ 0.32) & 20.39 & 
{87.83} \scriptsize($\pm$ 0.29) & 20.39 & 
& 
{76.03} \scriptsize($\pm$ 0.90) & 20.39 &
{69.73} \scriptsize($\pm$ 0.91) & 20.39 \\

& \texttt{FedProx}~\citep{li2018federated} & 
{89.04} \scriptsize($\pm$ 0.33) & 20.39 & 
{87.31} \scriptsize($\pm$ 0.21) & 20.39 & 
& 
{76.61} \scriptsize($\pm$ 0.92) & 20.39 &
{69.40} \scriptsize($\pm$ 0.73) & 20.39 \\

& \texttt{Clustered-FL}~\citep{sattler2019clustered} & 
{88.02} \scriptsize($\pm$ 0.37) & 20.39 & 
{87.33} \scriptsize($\pm$ 0.29) & 20.39 & 
& 
{74.27} \scriptsize($\pm$ 0.83) & 20.39 &
{68.84} \scriptsize($\pm$ 0.84) & 20.39 \\

& \texttt{Per-FedAvg}~\citep{fallah2020personalized} & 
{88.46} \scriptsize($\pm$ 0.53) & 20.39 & 
{87.29} \scriptsize($\pm$ 0.24) & 20.39 & 
& 
{74.90} \scriptsize($\pm$ 0.58) & 20.39 &
{68.67} \scriptsize($\pm$ 0.79) & 20.39 \\

& \texttt{FedFOMO}~\citep{zhang2021personalized} & 
{88.34} \scriptsize($\pm$ 0.26) & 122.33 & 
{84.03} \scriptsize($\pm$ 0.34) & 122.33 & 
& 
{72.12} \scriptsize($\pm$ 0.96) & 122.33 &
{61.45} \scriptsize($\pm$ 0.93) & 122.33 \\

& \texttt{pFedPara}~\citep{anonymous2022fedpara} & 
{88.70} \scriptsize($\pm$ 0.25) &  7.4  & 
{88.24} \scriptsize($\pm$ 0.22) &  7.4 &  
& 
{75.36} \scriptsize($\pm$ 0.93) &  7.4 &
{70.26} \scriptsize($\pm$ 0.85) & 7.4  \\

\cmidrule{2-11}
& \texttt{Factorized-FL} (Ours) & 
{86.56} \scriptsize($\pm$ 0.39) &  \textbf{0.32} & 
{86.31} \scriptsize($\pm$ 0.27)  & \textbf{0.32} & 
& 
{66.25} \scriptsize($\pm$ 0.71) & \textbf{0.32} &
{66.12} \scriptsize($\pm$ 0.79)& \textbf{0.32}  \\

& \texttt{Factorized-FL $\beta$} (Ours) & 
\textbf{{91.04} \scriptsize($\pm$ 0.73)} &  18.25 & 
\textbf{{89.57} \scriptsize($\pm$ 0.47)} & 18.25 & 
& 
\textbf{{81.07} \scriptsize($\pm$ 0.53)} & 18.25 &
\textbf{{74.63} \scriptsize($\pm$ 0.84)} & 18.25 \\

\midrule
\bottomrule
\end{tabular}}

\vspace{0.1in}

\resizebox{\textwidth}{!}{
    \begin{tabular}{l c c c c c c cc}
    \toprule
    \midrule
    
    \multirow{2}{*}{\textbf{Method}} & 
    \textbf{Household} & 
    \textbf{Fruit\&Food} & 
    \textbf{Tree\&Flower} &
    \textbf{Transport} & 
    \textbf{Animals} & &
    \multicolumn{2}{c}{\textbf{AVERAGE}} \\
    
    & 
    Accuracy [\%] & 
    Accuracy [\%] & 
    Accuracy [\%] & 
    Accuracy [\%] & 
    Accuracy [\%] & &
    Accuracy [\%] &  Cost [Gb] \\
    
    \midrule
    
    \texttt{Stand-Alone} & 
    {59.38} \scriptsize($\pm$ 0.70) & 
    {63.74} \scriptsize($\pm$ 1.76) & 
    {61.20} \scriptsize($\pm$ 0.64) & 
    {63.22} \scriptsize($\pm$ 2.12) & 
    {58.40} \scriptsize($\pm$ 1.20) & &
    {61.35} \scriptsize($\pm$ 1.90) & - \\
    
    \texttt{FedAvg}~\citep{McMahan2017CommunicationEfficientLO} & 
    {55.08} \scriptsize($\pm$ 2.49) & 
    {63.18} \scriptsize($\pm$ 2.45) & 
    {57.76} \scriptsize($\pm$ 1.77) & 
    {57.96} \scriptsize($\pm$ 2.97) & 
    {53.61} \scriptsize($\pm$ 1.21) & &
    {56.42} \scriptsize($\pm$ 1.65) & {40.78}  \\
    
    \texttt{FedProx}~\citep{li2018federated} & 
    {56.77} \scriptsize($\pm$ 2.59) & 
    {61.33} \scriptsize($\pm$ 1.28) & 
    {58.14} \scriptsize($\pm$ 0.51) & 
    {55.79} \scriptsize($\pm$ 0.82) & 
    {51.43} \scriptsize($\pm$ 2.17) & &
    {56.71} \scriptsize($\pm$ 1.52) & {40.78} \\
    
    \texttt{Clustered-FL}~\citep{sattler2019clustered} & 
    {59.44} \scriptsize($\pm$ 2.31) & 
    {66.93} \scriptsize($\pm$ 0.88) & 
    {60.03} \scriptsize($\pm$ 1.13) & 
    {62.17} \scriptsize($\pm$ 2.55) & 
    {55.01} \scriptsize($\pm$ 2.14) & &
    {59.20} \scriptsize($\pm$ 2.16) & {40.78}\\
    
    \texttt{Per-FedAvg}~\citep{fallah2020personalized} & 
    {64.01} \scriptsize($\pm$ 1.56) & 
    {67.68} \scriptsize($\pm$ 1.07) & 
    {61.62} \scriptsize($\pm$ 1.86) & 
    {64.36} \scriptsize($\pm$ 1.27) & 
    {60.25} \scriptsize($\pm$ 0.88) & &
    {62.92} \scriptsize($\pm$ 1.60) & {40.78}\\
    
    \texttt{FedFOMO}~\citep{zhang2021personalized} & 
    {59.70} \scriptsize($\pm$ 1.78) & 
    {64.32} \scriptsize($\pm$ 1.48) & 
    {63.87} \scriptsize($\pm$ 2.19) & 
    {62.57} \scriptsize($\pm$ 0.97) & 
    {57.75} \scriptsize($\pm$ 2.28) & &
    {62.07} \scriptsize($\pm$ 1.80) & {244.66} \\
    
    \texttt{pFedPara}~\citep{anonymous2022fedpara} &
    {60.35} \scriptsize($\pm$ 3.30) & 
    {65.56} \scriptsize($\pm$ 0.60) & 
    {61.98} \scriptsize($\pm$ 2.02) & 
    {60.16} \scriptsize($\pm$ 6.66) & 
    {56.12} \scriptsize($\pm$ 2.86) & &
    {61.11} \scriptsize($\pm$ 2.61) & {15.98} \\
    
    \cmidrule{1-9}
    \texttt{Factorized-FL} (Ours) & 
    \textbf{{64.06} \scriptsize($\pm$ 0.16)} & 
    \textbf{{68.55} \scriptsize($\pm$ 0.16)} & 
    \textbf{{64.39} \scriptsize($\pm$ 2.23)} & 
    \textbf{{66.93} \scriptsize($\pm$ 1.03)} & 
    \textbf{{61.33} \scriptsize($\pm$ 3.56)} & &
    \textbf{{64.49} \scriptsize($\pm$ 1.57)} & \textbf{{0.64}} \\
    
    \midrule
    \bottomrule
    \end{tabular}}
    
\end{table*}

\vspace{-0.1in}

\begin{figure}[t]
\small
\centering
\vspace{-0.05in}
\includegraphics[width=0.48\textwidth]{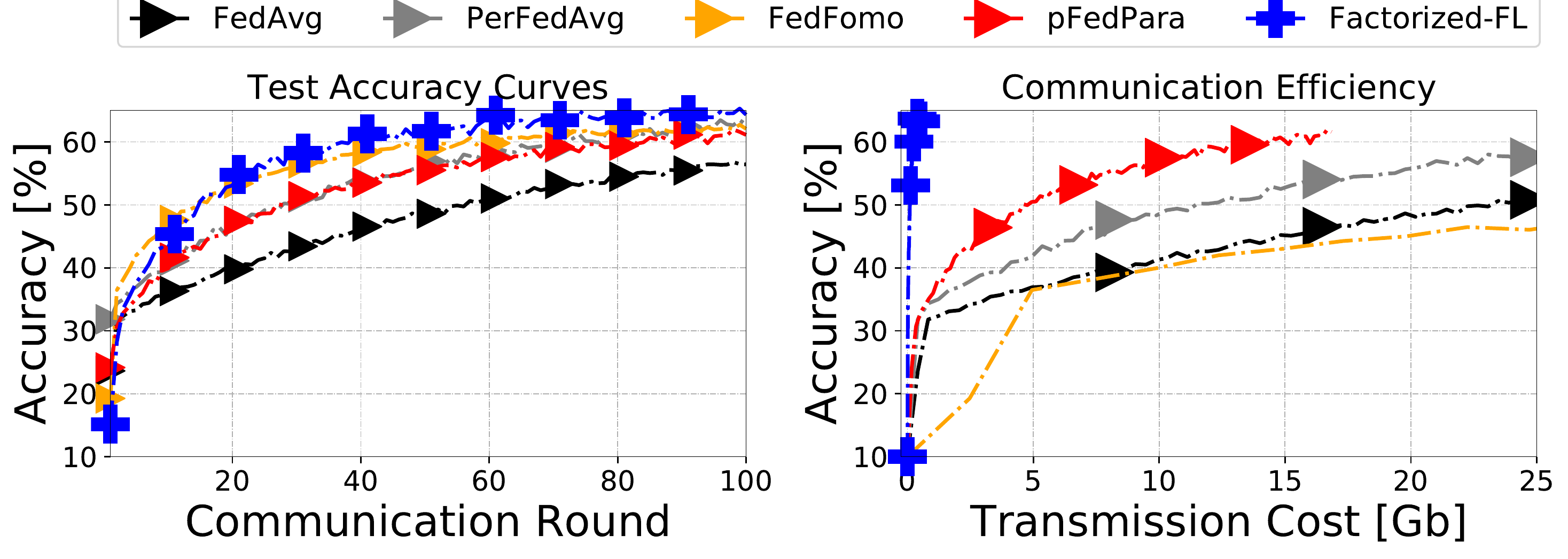} 
\caption{\small{\textbf{Test Accuracy Curves \& Communication Costs (GBytes)}} We plot the averaged test accuracy curves over communication rounds and transmission costs (GBytes) for domain-heterogeneous setting (corresponding to the Table~\ref{tbl:permuted} Bottom). }
\label{fig:acc_plot}
\vspace{-0.25in}
\end{figure}

\paragraph{Domain-heterogeneous FL}
Table~\ref{tbl:permuted} (Bottom) shows the experimental results for the domain and label heterogeneous scenarios. We observe that the conventional FL baselines, i.e. \texttt{FedAvg}, \texttt{FedProx}, fail to obtain better performance over purely local training baseline (\texttt{Stand-Alone}) due to the naive aggregation of extremely heterogeneous knowledge, which causes detrimental knowledge collapse. \texttt{FedFOMO} and \texttt{Clustered-FL} shows slightly higher performance ($1-2\%p$) over \texttt{Stand-Alone} model, as they can avoid irrelevant clients when aggregating local knowledge. The other personalized FL methods, i.e. \texttt{Per-FedAvg} and \texttt{pFedPara}, also show $1-2\%p$ higher performance over \texttt{Stand-Alone} model as they are specialized for personalized FL scenarios. 
However, on average, our method largely outperforms all baseline models even with the smallest communication costs, as shown in Figure~\ref{fig:acc_plot}. In the figure we plot the convergence rate of our Factorized-FL framework over communication round and transmission cost, compared to baseline models. Our method consistently obtain superior performance in the extremely heterogeneous scenarios, with significantly faster convergence and superior accuracy per transmission cost. Specifically, unlike \texttt{pFedPara} which uses low rank matrices for knowledge sharing, as we only communicate with factorized vectors, such as  $\textbf{u}$ for base knowledge sharing and $\textbf{v}$ for similarity matching, our method can largely reduce the communication costs while achieving superior performance over it, as shown in Figure~\ref{fig:acc_plot} (Right).

\begin{figure*}[t]
\small
\centering
\begin{minipage}{0.71\textwidth}
    \resizebox{\linewidth}{!}{
        \begin{tabular}{ccccc}
            \includegraphics[width=0.24\linewidth]{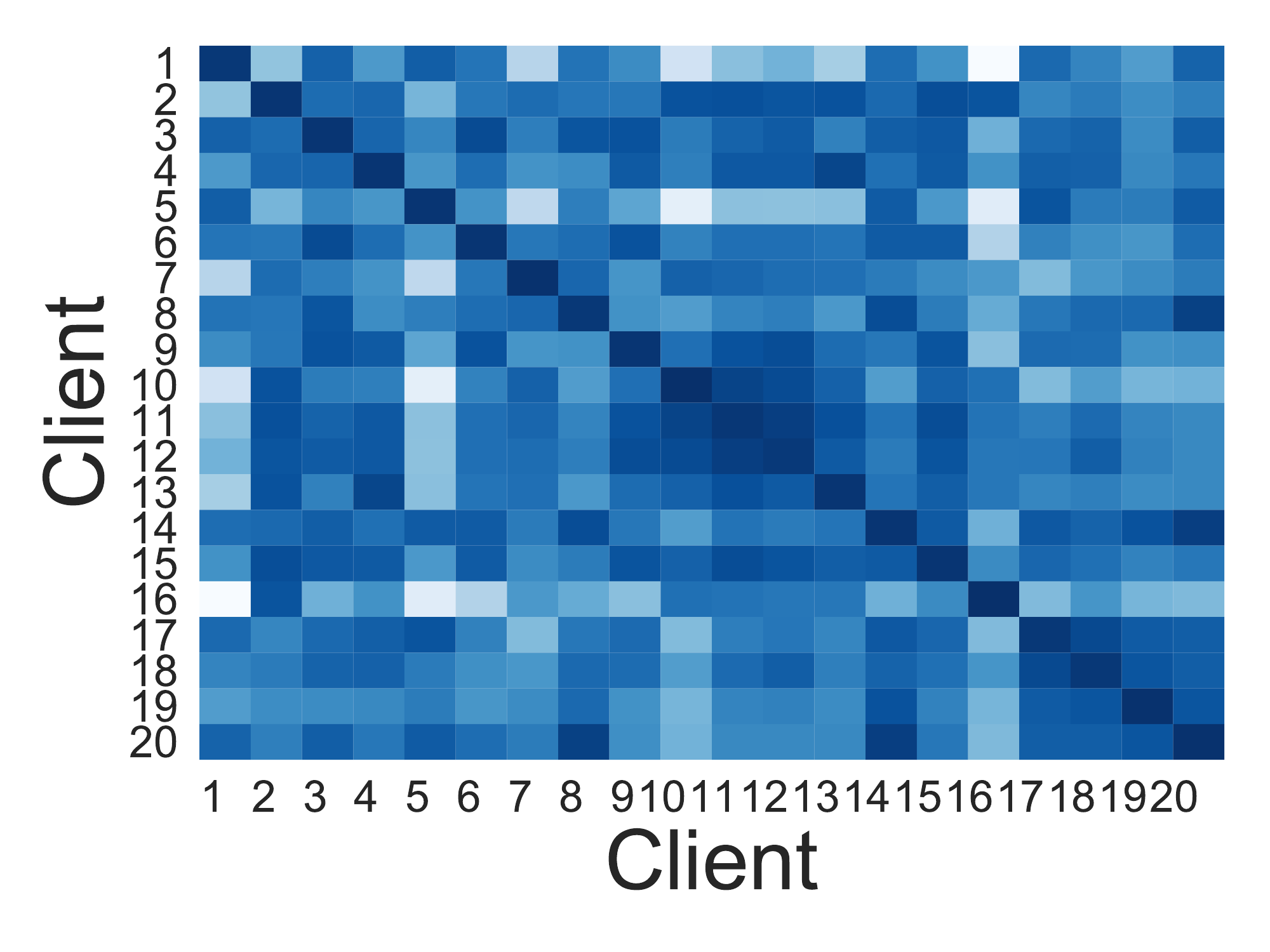} &
            \includegraphics[width=0.24\linewidth]{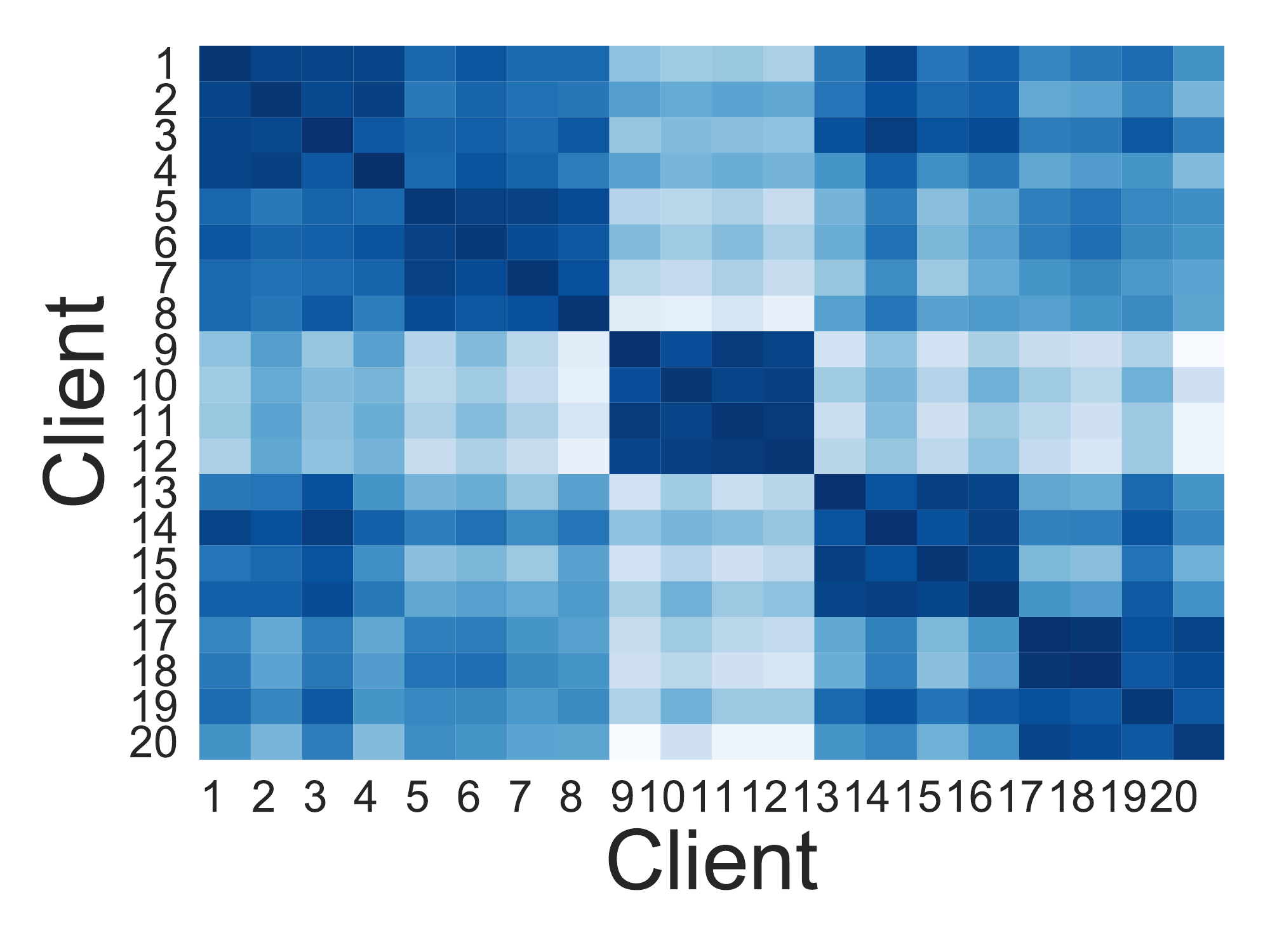} &
            \includegraphics[width=0.24\linewidth]{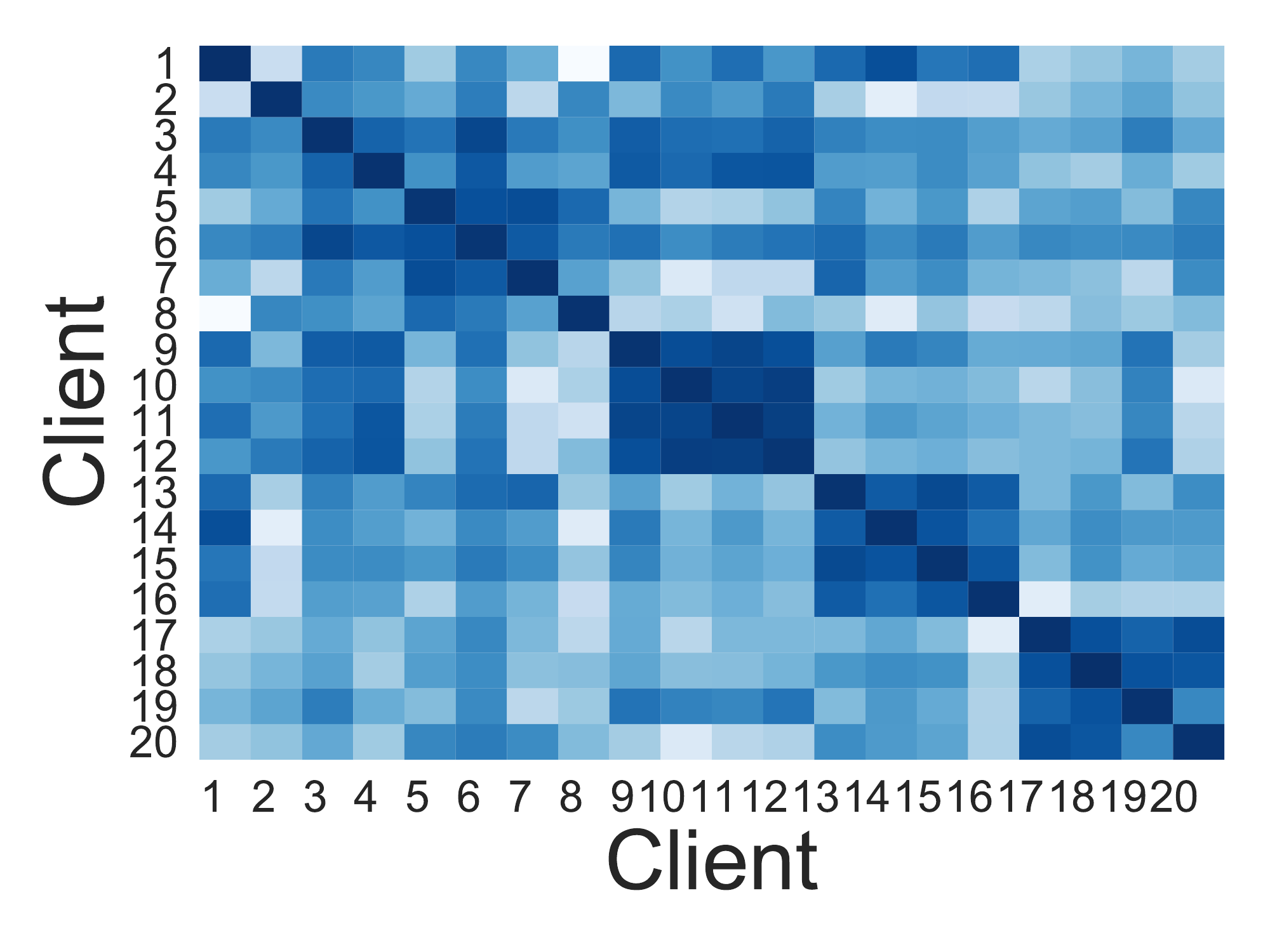} &
            \includegraphics[width=0.24\linewidth]{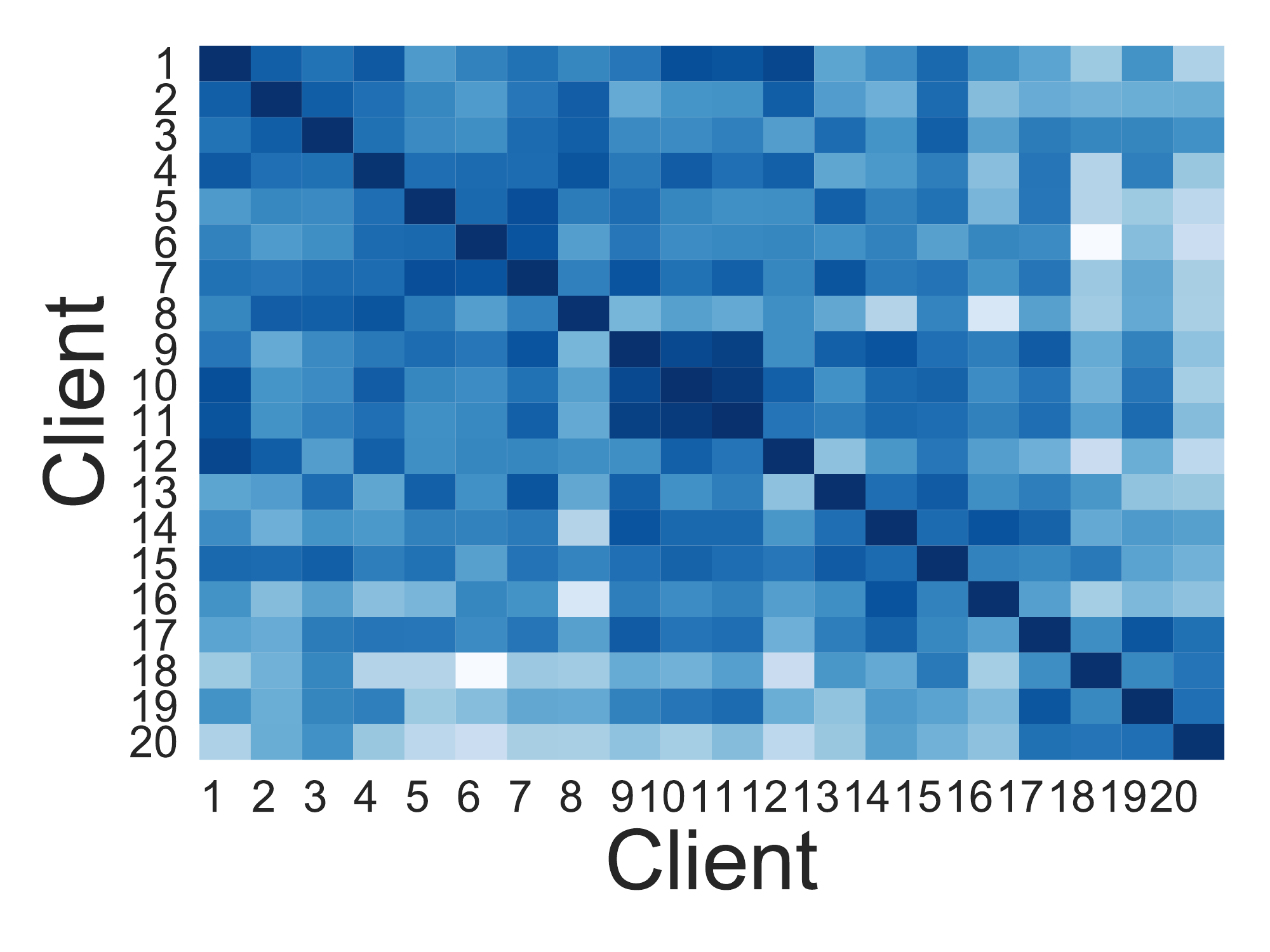} &
            \includegraphics[width=0.24\linewidth]{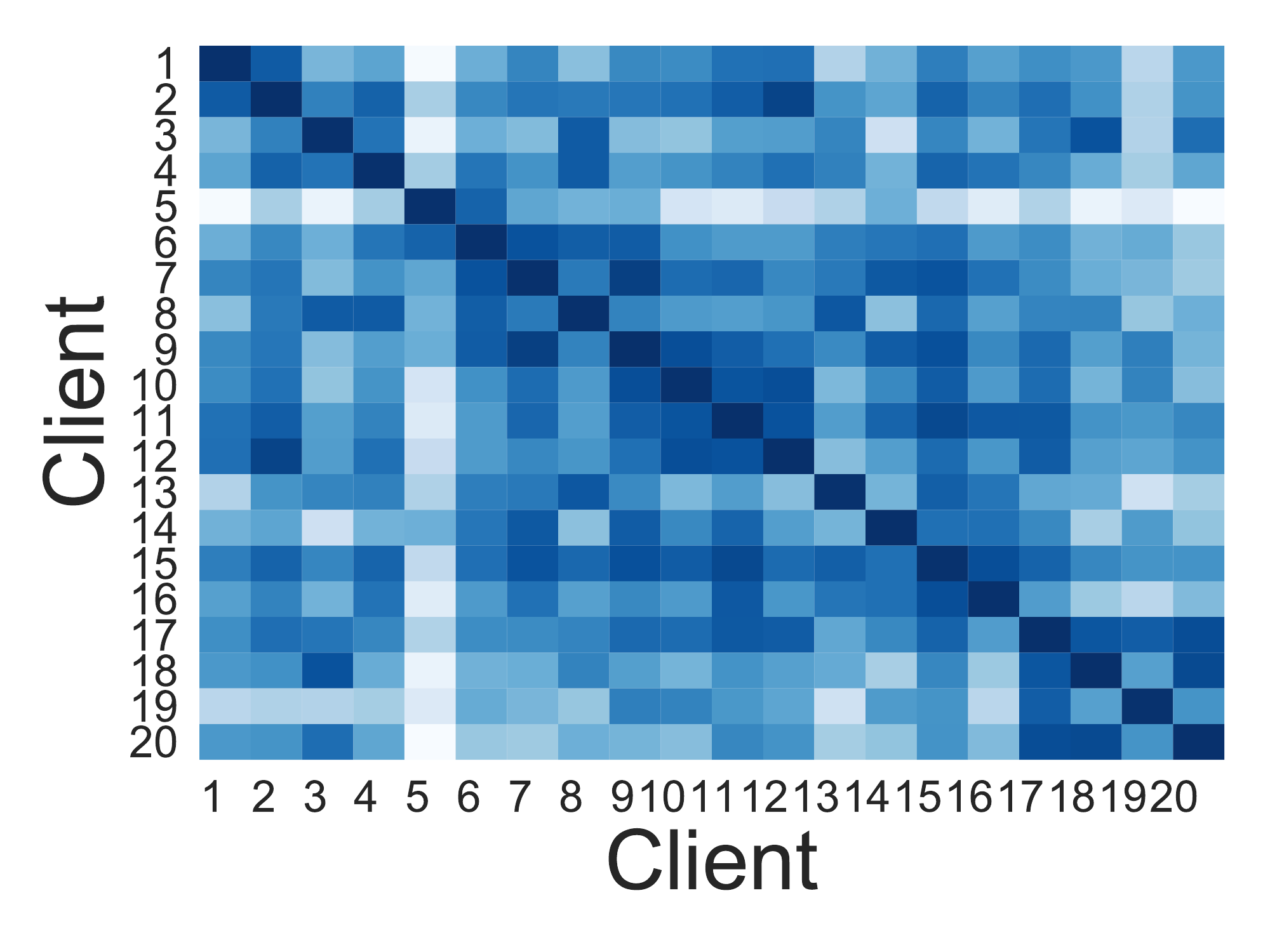} 
            \\
            \includegraphics[width=0.24\textwidth]{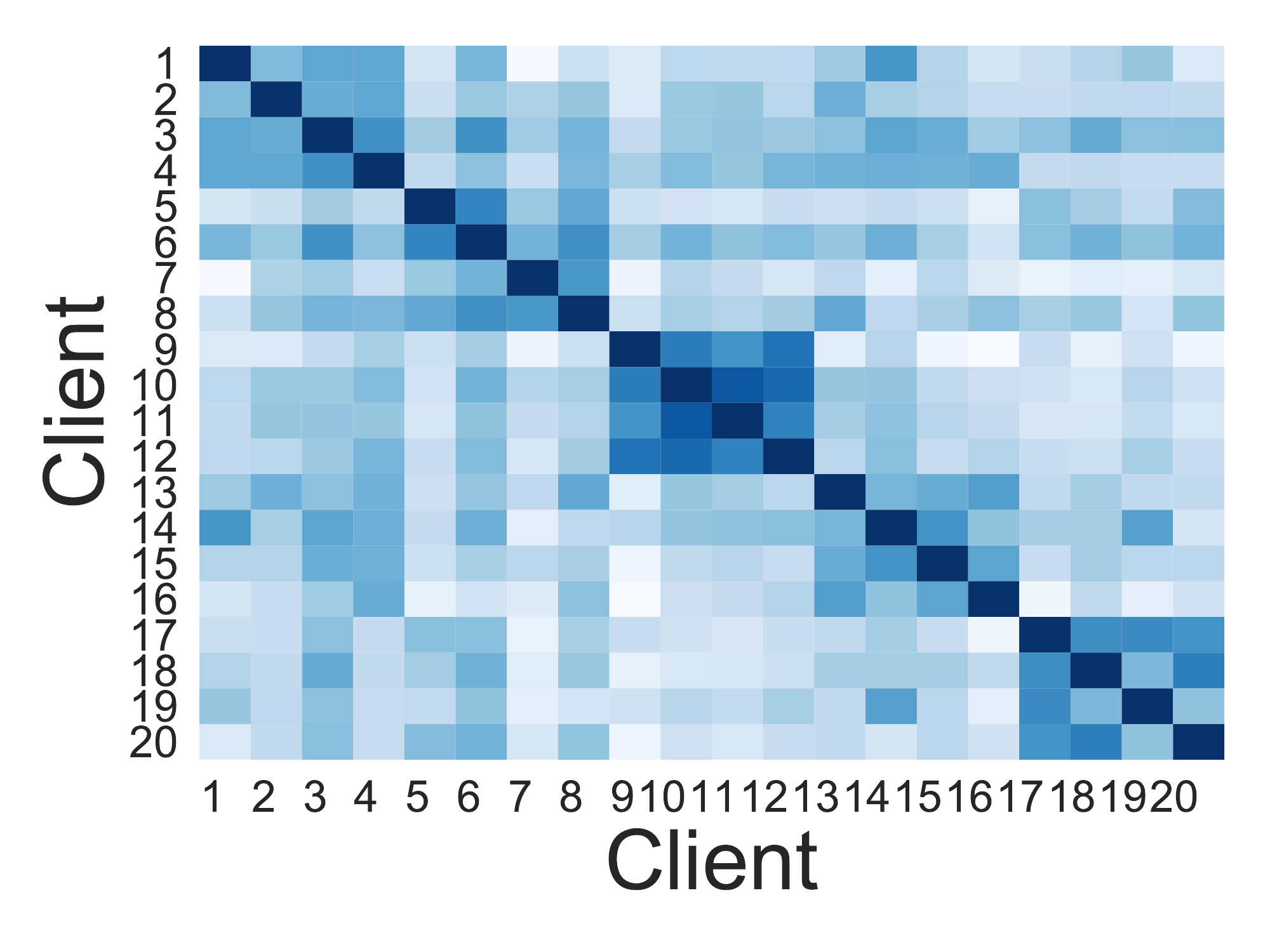} &
            \includegraphics[width=0.24\textwidth]{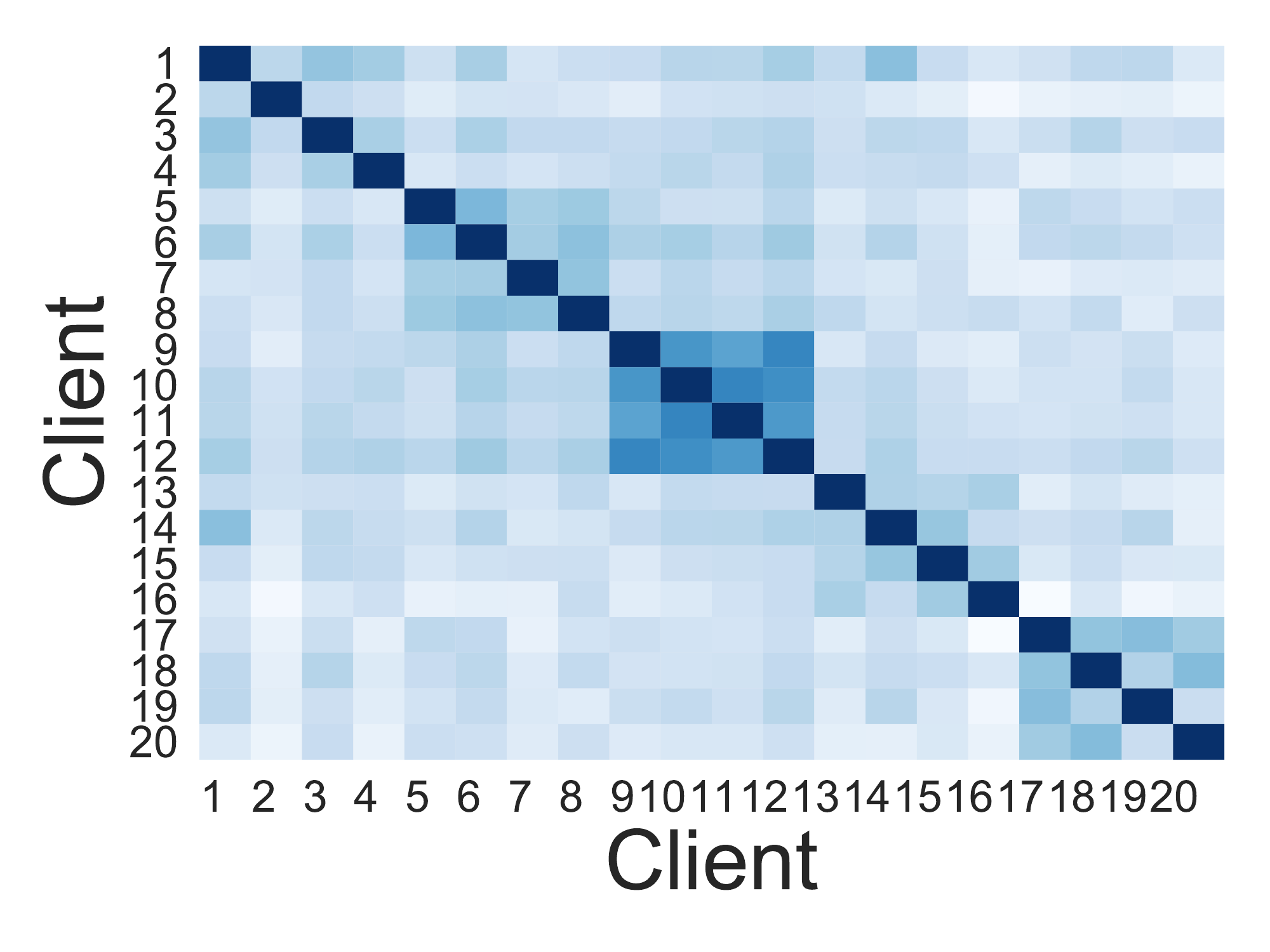} &
            \includegraphics[width=0.24\textwidth]{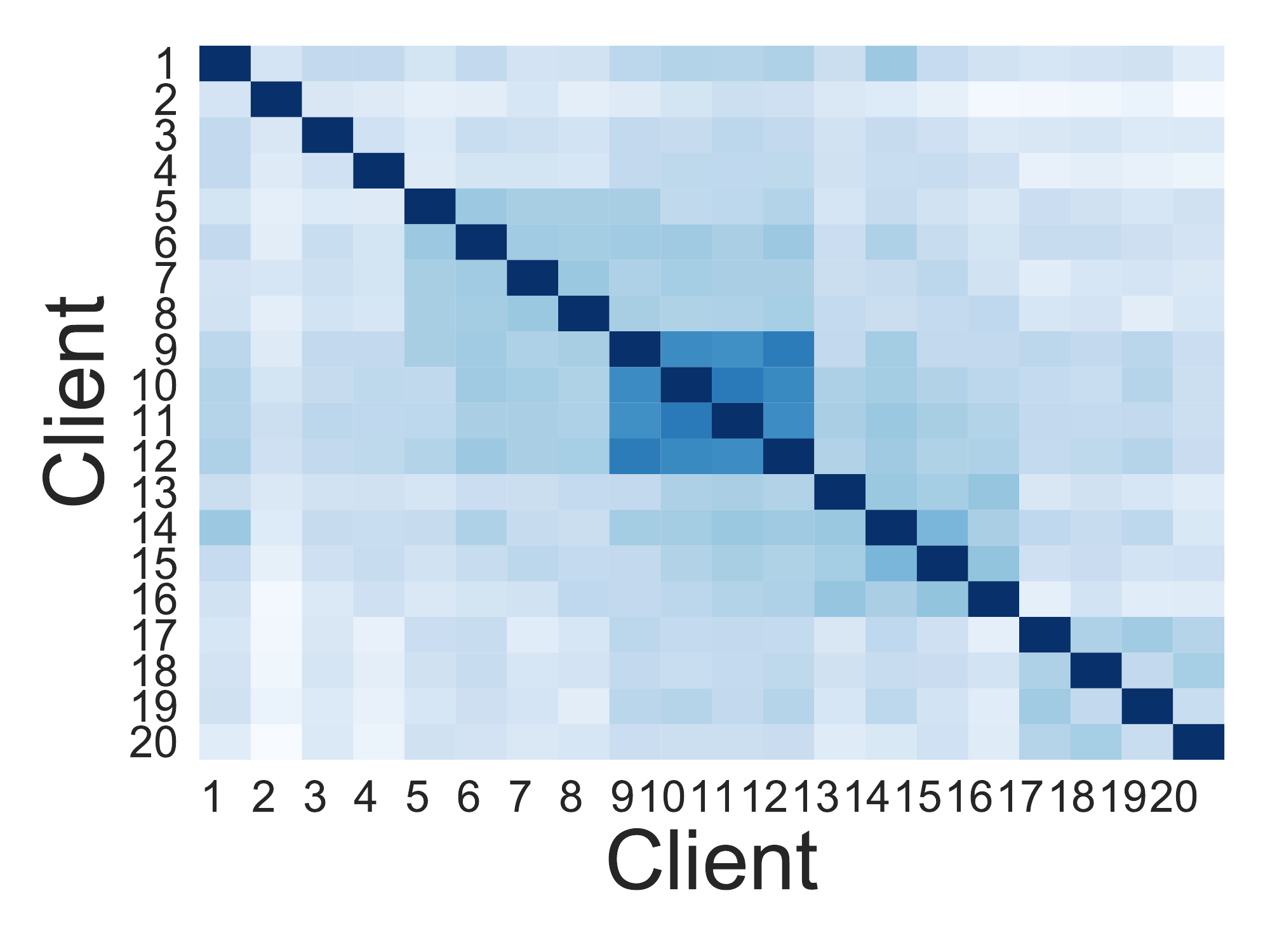} &
            \includegraphics[width=0.24\textwidth]{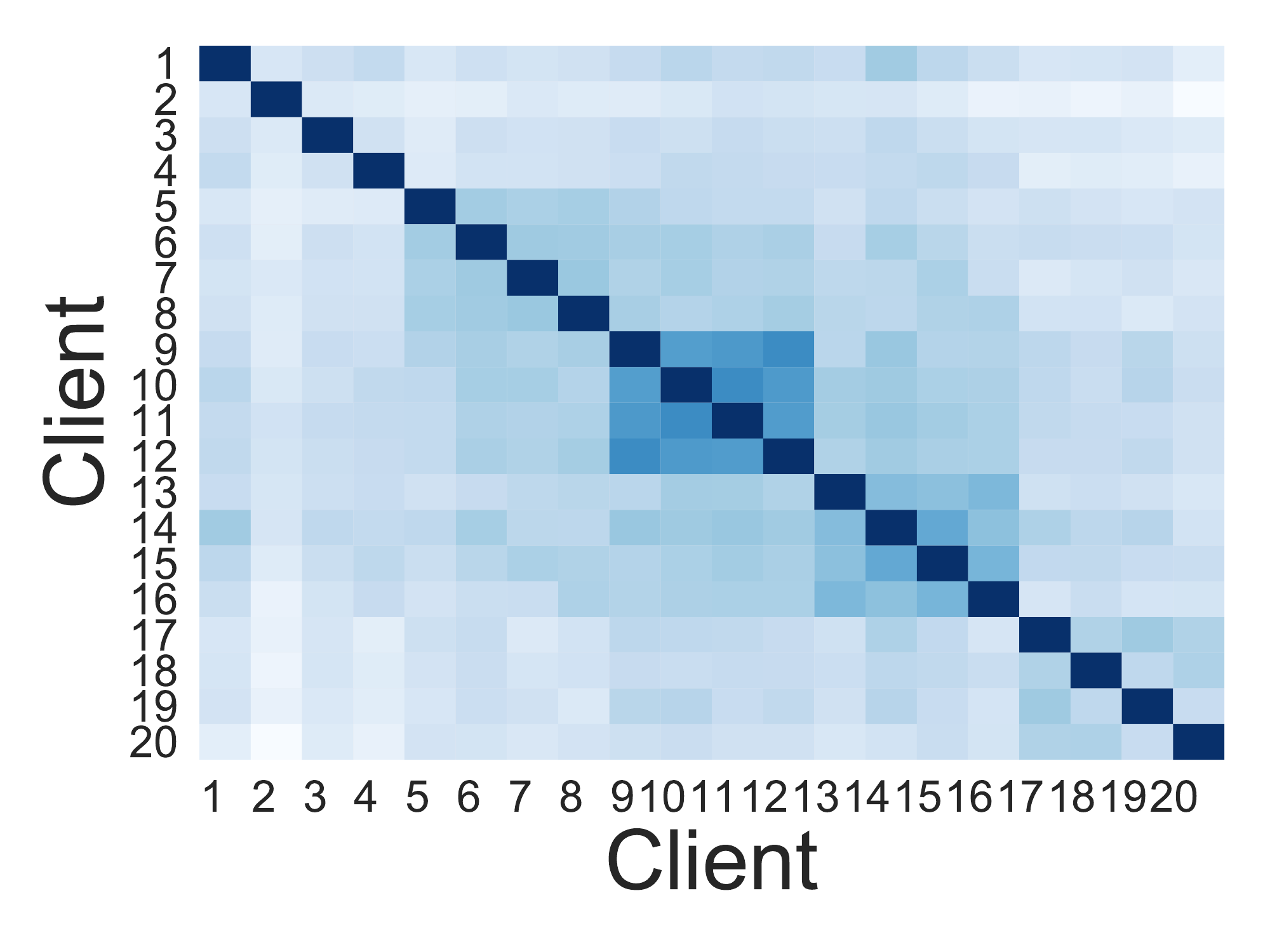} &
            \includegraphics[width=0.24\textwidth]{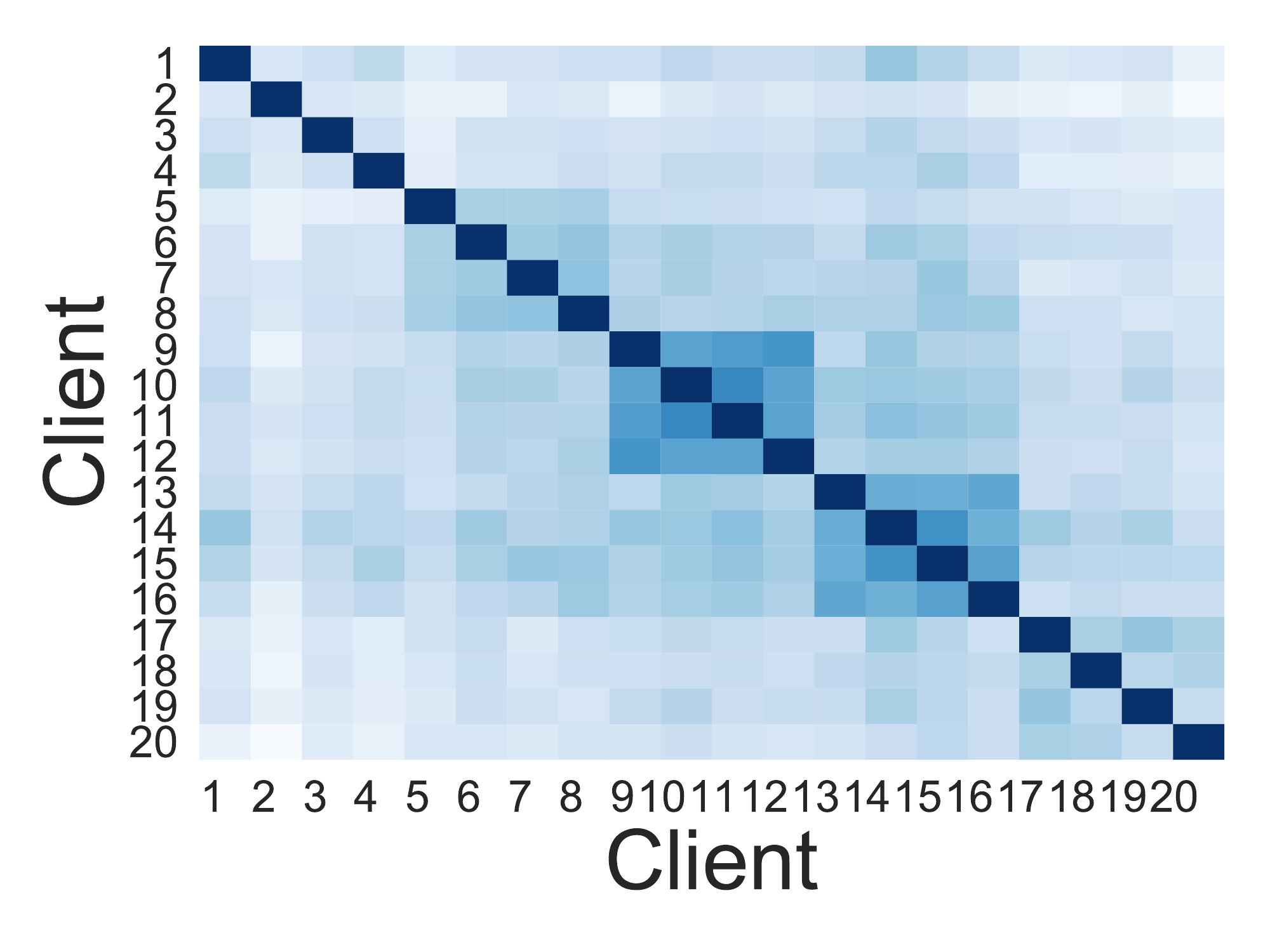} 
        \end{tabular}
    }
\end{minipage}
\begin{minipage}{0.28\textwidth}
    \includegraphics[width=0.95\textwidth]{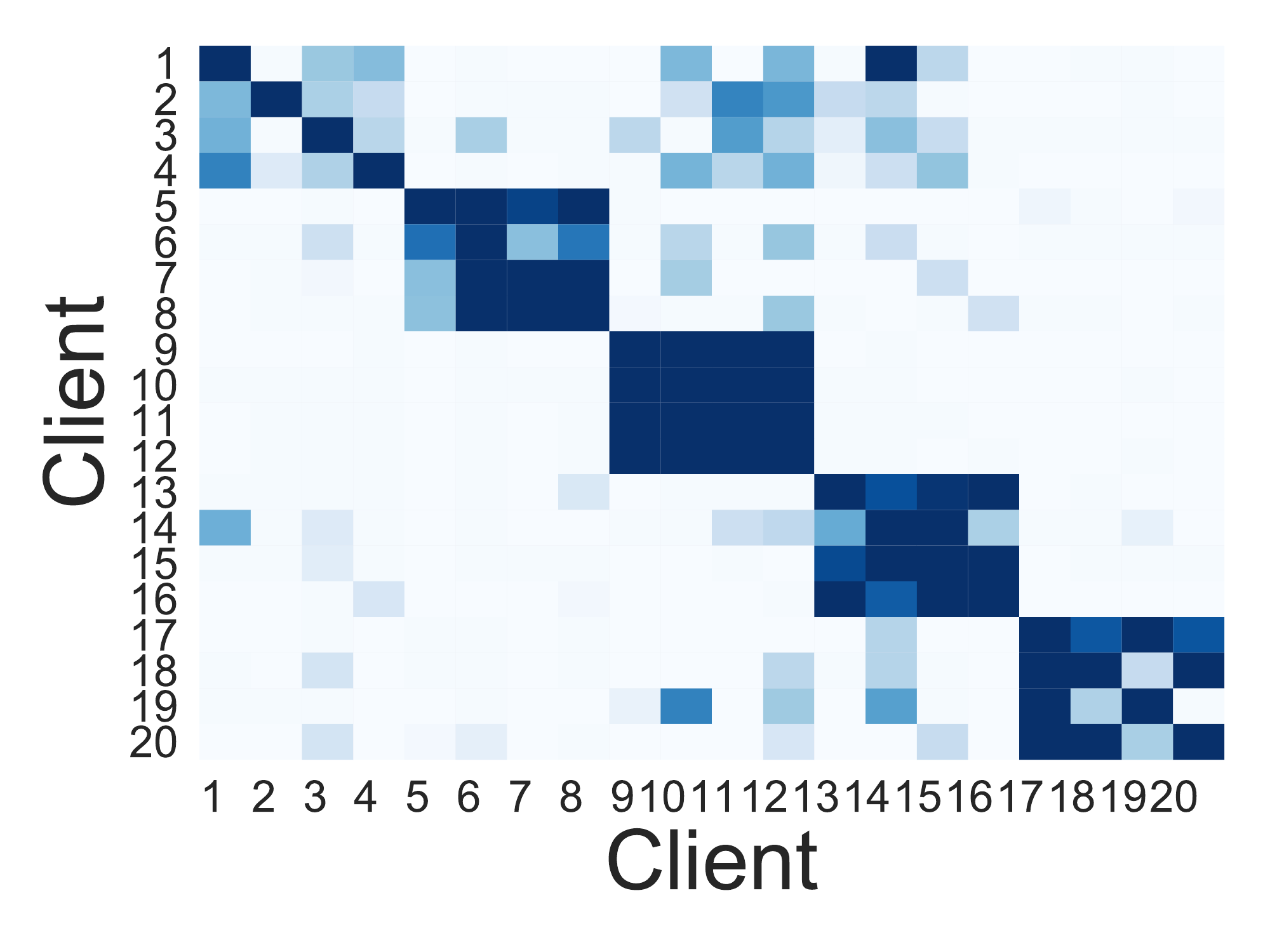} 
\end{minipage}
\\
(a) $r$=$1$ 
\hfil \hspace{0.05in} (b) $r$=$25$ 
\hfil \hspace{0.05in} (c) $r$=$50$ 
\hfil \hspace{0.05in} (d) $r$=$75$ 
\hfil \hspace{0.05in} (e) $r$=$100$
\hfil \hspace{0.05in} (f) Freq. of Client Matching
\\

\hspace{-0.15in}
\begin{minipage}{0.24\textwidth}
\includegraphics[width=0.98\textwidth]{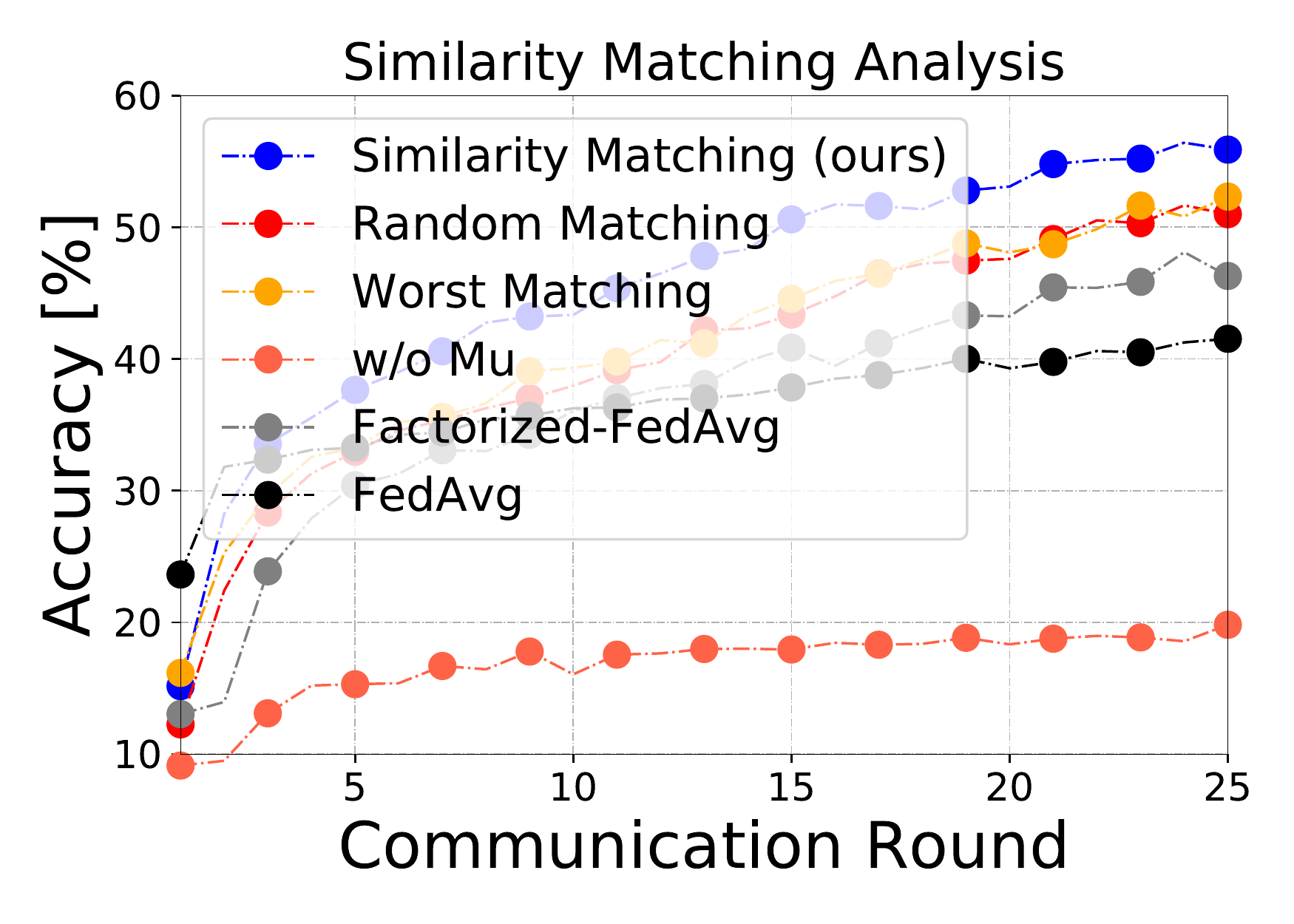} 
\end{minipage}
\begin{minipage}{0.45\textwidth}
    \resizebox{0.99\textwidth}{!}{
        \begin{tabular}{cccccc}
            \toprule 
    	    \midrule
    	    \centering
    	    \multirow{2}{*}{Backbone} & Num. & Head & Regular & $I \cdot F \times O \cdot F$  & $F \cdot F \times I \cdot O$  \\
    	     & FC & BN & Kernel & Factorization & Factorization (Ours) \\
    	    \midrule
    	    \multirow{3}{*}{ResNet-9} & 1 & - & $61.29\%$ & $60.96\%$ & $\textbf{62.34\%}$ \\
    	    & 2 & \cmark & $56.08\%$ & $60.81\%$ & $\textbf{62.65\%}$ \\
     	    & 2 & \xmark & $50.01\%$ & $63.42\%$ & $\textbf{64.39\%}$ \\
     	    \midrule
            \multirow{3}{*}{ResNet-18} & 1 & - & $63.06\%$ & $55.72\%$ & $\textbf{66.23\%}$ \\
    	     & 2 & \cmark & $58.36\%$ & $61.39\%$ & $\textbf{68.78\%}$ \\
    	     & 2 & \xmark  & $51.26\%$ & $63.32\%$ & $\textbf{67.93\%}$ \\
    	    \midrule
    	    \bottomrule 
        \end{tabular}
        \label{tbl:analysis_factorization_3}
    }
\end{minipage}
\begin{minipage}{0.29\textwidth}
    \resizebox{0.99\textwidth}{!}{
        \begin{tabular}{ccccc}
            \toprule 
    	    \midrule
    	    \centering
    	    \multirow{2}{*}{Backbone} & \multirow{2}{*}{$\lambda_\text{sparse}$} & Num. & Acc. \\
    	    & & Params. & (\%) \\
    	    \midrule
    	     ResNet-9 & - & 2.574 M (100\%) & $61.95\%$ \\
    	     \midrule
    	     \multirow{5}{*}{\shortstack{Factorized \\ ResNet-9}} &  1e-4 & 2.731 M (106\%) & $63.31\%$ \\
    	      &  3e-4 & 2.515 M (97\%) & $63.74\%$ \\
    	      &  5e-4 & 2.283 M (88\%) & $63.12\%$ \\
    	      &  7e-4 & 2.068 M (80\%) & $62.69\%$ \\
    	      &  1e-3 & 1.784 M (69\%) & $62.67\%$ \\
    	    \midrule
    	    \bottomrule
        \end{tabular}
        \label{tbl:analysis_factorization_4}
    }
\end{minipage}\hfill
\\
\hspace{-0.4in} (g) Ablation Study 
\hfil \hspace{-0.35in} (h) Factorization of Different Architectures
\hfil \hspace{0.05in} (i) Sparsity Analysis

\vspace{-0.05in}
\caption{\small{\textbf{In-depth analysis on \texttt{Factorized-FL} algorithms.} \textbf{Top:} While $\textbf{u}^{L-1}$ learns task-general knowledge (upper row), $\textbf{v}^{L-1}$ captures task-specific knowledge and can be utilized for clustering relevant clients (bottom row). We show cosine similarity from round (a) $r$=$1$ to (e) $r$=$100$. (f) shows the frequency of client matching after $100$ rounds. Darker colors indicate higher scores. (g) ablation study, (h) applicability of our factorization methods, and (i) sparsity analysis of the hyperparameter $\lambda_{\textbf{sparsity}}$. }}
\label{fig:analysis}
\vspace{-0.2in}
\end{figure*}

\paragraph{Effect of kernel factorization} 
In Figure~\ref{fig:analysis} (g), we perform an ablation study of our factorization method in the domain-heterogeneous scenario. To clearly see the effectiveness of our factorization methods, we compare \texttt{Factorized-FedAvg}, a variant of Factorized-FL $\beta$ without similarity matching, against \texttt{FedAvg}. As shown, \texttt{Factorized-FedAvg} achieves higher performance over the original \texttt{FedAvg} model. As the only difference between the two is whether kernel is factorized or not, this clearly demonstrates that our factorization method alone improves the model performance by alleviating knowledge collapse.

We further analyze the effect of the sparse bias matrix $\mathcal{M}$. When we remove $\mathcal{M}$ from the \texttt{Factorized-FedAvg} model, w/o Mu in the figure, we observe large performance drop. This shows that that the bias term is essential in compensating for the loss of expressiveness from rather extreme factorization of the weight matrices into rank-1 vectors. With only $\mathcal{U}$ and $\mathcal{V}$, we use $90\%$ less model parameters ($0.27 M$) compared to the regular kernel model ($2.574 M$). 

We can control the sparsity of bias $\mathcal{M}$ by varying the hyper-parameter $\lambda_{\text{sparsity}}$ which gives intensity for the $L_1$ regularizer described in Eq.~\ref{eq:loss}. In Figure~\ref{fig:analysis} (i), we train Client 1 on CIFAR-10 IID Partition 1 for 20 epochs using a single factorized model and a regular model, respectively, and report their model size and performance. As shown, our factorized model still outperforms regular models $(2.574 M)$ even with $30\%$ less parameters $(1.784 M)$ (we further analyze on the effect of the sparsity in Appendix~\ref{appdx:sparisty}). Under the same experimental setup, we also experiment with ResNet-18 architecture to verify the scalability of our method (Figure~\ref{fig:analysis} (h)). Regardless of the model size, our factorized kernel models (the right most) consistently shows better performance. Interestingly, when we add more (factorized) fully-connected layer with or without batch normalization, our method obtains performance improvements, while regular kernel CNN models suffer from performance degeneration. These results demonstrate that our factorization method is scalable, consistent, and reliable in terms of model architectures consisting of convolutional, fully-connected layers, batch normalization, and skip connections. Additionally, we observe that model factorized by $\textbf{u} \in \mathbb{R}^{F \cdot F}$ and $\textbf{v} \in \mathbb{R}^{I \cdot O} $ perform better than the model factorized by $\textbf{u} \in \mathbb{R}^{I \cdot F}$ and $\textbf{v} \in \mathbb{R}^{O \cdot F} $. This is because in the former case, the factorization will separate the base filter knowledge from  the task-specific configurations of the filters, as described in Figure~\ref{fig:factorize_detail}, but the factorization does not have such a natural interpretation in the latter case.

\vspace{-0.1in}
\paragraph{Effect of similarity matching} 
To verify the efficacy of our similarity matching method using the personalized factorized vector $\textbf{v}$, we visualize the inter-client similarity of $\textbf{u}^{L-1}_{f_k}$ and $\textbf{v}^{L-1}_{f_k}$ form the second last layer of $20$ clients  on domain heterogeneous setting. As shown in Figure~\ref{fig:analysis} from (a) round 1 to (e) round 100, we observe $\textbf{u}^{L-1}_{f_k}$ (upper row) are indeed highly correlated with other clients as the similarity scores are high (the darker color indicate higher values) while $\textbf{v}^{L-1}_{f_k}$ (bottom row) are relatively uncorrelated to each other, as expected as our assumption that $\mathcal{U}_{f_k}$ capture base knowledge across all clients and $\mathcal{V}_{f_k}$ capture personalized knowledge. Further, we also observe $\textbf{v}^{L-1}_{f_k}$ obtains higher similarity to that of parameters trained on the same domains (but with permuted labels), i.e. Client 1-4, Client 5-8, Client 9-12, Client 13-16, and Client 17-20, showing that they are effective in task- and domain-level similarities across models. We further visualize the frequency of the client matching across clients in Figure (f), after $100$ communication rounds. With only a single vector parameter $\textbf{v}^{L-1}_{f_k}$, we both efficiently and effectively find which clients will be helpful to certain other clients. 

For further analysis on our similarity matching, we compare it against Random and Worst Matching baselines under the multi domain scenario. The random matching baseline randomly selects three arbitrary models to be aggregated at each round, while the worst matching baseline selects three most dissimilar models. As shown in Table~\ref{fig:analysis} (g), both random and worst matching methods significantly suffer from the performance degeneration compared to our best matching strategy. This shows that our similarity matching algorithm is indeed effective, selecting beneficial knowledge from other clients.

\section{Conclusion}
We introduced a realistic federated learning scenario where the labeling schemes are not synchronized across all participants (label heterogeneity) and the tasks and the domains tackled by each local model is different from those of others (domain heterogeneity). We then proposed a novel federated learning framework to tackle this problem, whose local model weights are factorized into the product of two vectors plus a sparse bias term. We then aggregate only the first vectors, for them to capture the common knowledge across clients, while allowing the other vectors and the sparse bias term to be client-specific, accounting for label and domain heterogeneity. Further, we use the client-specific vectors to measure the similarity scores across local models, which are then used for weighted averaging, for personalized federated learning of each local model. Our method not only avoids knowledge collapse from aggregating incompatible parameters across heterogeneous models, but also significantly reduces the communication costs. We validate our method on both label and domain heterogeneous settings, on which it largely outperforms relevant baselines.

\bibliography{citation}
\bibliographystyle{icml2022}
\newpage
\clearpage
\appendix
\onecolumn

\paragraph{Organization} We provide in-depth descriptions for our algorithms, experimental setups, i.e. dataset configurations, implementation \& training details, and additional experimental results \& analysis that are not covered in the main document, as organized as follows:

\begin{itemize}
    \item \textbf{Section~\ref{appdx:algorithm}}: We provide our pseudo-code algorithms for \texttt{Factorized-FL} and \texttt{Factorized-FL $\beta$}. 
    \item \textbf{Section~\ref{appdx:dataset}}: We describe dataset configurations for label- and domain-heterogenous scenario.
    \item \textbf{Section~\ref{appdx:details}} - We elaborate on detailed implementation and training details for our methods and baselines.
    \item \textbf{Section~\ref{appdx:experiment}} - We provide additional experimental results and analysis. 
\end{itemize}

\section{\texttt{Factorized-FL} Algorithms}
\label{appdx:algorithm}

In this section, we describe our pseudo-code algorithms for \texttt{Factorized-FL} and \texttt{Factorized-FL $\beta$} in Algorithm~\ref{algo:factorized_fl} and~\ref{algo:factorized_fl_beta}. Our \texttt{Factorized-FL} has strength for not only reducing the dimensionality of model parameters by factorizing them into rank 1 vector spaces and the additional highly-sparse matrices, but also effectively learning client-general and task-specific knowledge. Particularly, \texttt{Factorized-FL} transmits a small portion of the models which are a set of $\textbf{u}$ ($\mathcal{U}$) and a single vector $\textbf{v}^{L-1}$ form the second last layer of neural networks, which significantly reduces communication costs while showing strong performance in label- and domain-heterogeneous scenarios, as shown in Section~\ref{sec:exp} in the main document.

\begin{figure*}[h!]
    \begin{minipage}[t]{.5\linewidth}
        \begin{algorithm}[H]
            \small
            \caption{\textbf{\texttt{Factorized-FL} Algorithm}}
        	\begin{algorithmic}[1]
        	    \label{algo:factorized_fl}
        	    \STATE  $R$: number of communication rounds, $E$: number of epochs, $K$: number of clients, $\mathcal{F}$: a set of clients, $f_k$: a $k$th client, $\Omega(\cdot)$: our similarity matching function, $\epsilon$: hyper-parameter for scaling similarity score $\sigma$, $L$: number of layers in neural networks. $\mathcal{U}_{k}$, $\mathcal{V}_{k}$, and $\mathcal{M}_{k}$: factorized parameters of our  \texttt{Factorized-FL}.\\
                \STATE   \textbf{Function} RunServer()
                    \STATE initialize $\mathcal{F} $
                    \FOR{each round $r=1,2,\dots, R$}
                        \STATE $\mathcal{F}^{(r)} \leftarrow$ select $K^{(r)}$ clients from $\mathcal{F}$
                        \FOR{each client $f^{(r)}_k \in \mathcal{F}^{(r)}~\textbf{\mbox{in parallel}}$}
                            \IF{$r>1$} 
                                \STATE $\{\sigma_i \}_{i \neq k}^{K^{(r)}} \leftarrow \Omega(\textbf{v}_{f_k}^{L-1}, \textbf{v}_{f_{i\neq k:K^{(r)}}}^{L-1})$
                                \STATE $\mathcal{U}_{k}^{(r)} \leftarrow \frac{\text{exp}({\epsilon \cdot \sigma_i})}{\sum_{i=1}^{K^{(r)}} \text{exp}(\epsilon \cdot \sigma_i)} \sum_{i=1}^{K^{(r)}} \mathcal{U}_i^{(r)}$
                            \ENDIF 
                            \STATE $\mathcal{U}_k^{(r+1)}, \textbf{v}_{f_k}^{L-1}\leftarrow \text{RunClient} (\mathcal{U}_{k}^{(r)})$
                    	\ENDFOR
                    \ENDFOR 
                \STATE   \textbf{Function} RunClient($\mathcal{U}^{(r)}_k$)
                    \STATE $\mathcal{U}_{k}^{(r+1)} \leftarrow \mathcal{U}^{(r)}_k$  
                    \FOR{each local epoch $e$ from $1$ to $E$}
                        \FOR{minibatch $\mathcal{B} \in \mathcal{D}_{k}$}
                            \STATE $\theta_{\mathcal{U} \times \mathcal{V} \oplus \mathcal{M}} \leftarrow \theta_{\mathcal{U} \times \mathcal{V} \oplus \mathcal{M}}-\eta\nabla\mathcal{L}(\mathcal{B}; \theta_{\mathcal{U} \times \mathcal{V} \oplus \mathcal{M}})$
                        \ENDFOR
                    \ENDFOR
                    \STATE \textbf{return} $\mathcal{U}_{k}^{(r+1)}, \textbf{v}_{f_k}^{L-1}$
        	\end{algorithmic}
    	\end{algorithm}
    \end{minipage}
    \begin{minipage}[t]{.5\linewidth}
        \begin{algorithm}[H]
        \small
        \label{algo:factorized_fl_beta}
    	\caption{\textbf{\texttt{Factorized-FL $\beta$} Algorithm}}
    	    \begin{algorithmic}[1]
    	        \label{algo:factorized_fl_beta}
                \STATE   \textbf{Function} RunServer()
                    \STATE initialize $\mathcal{F} $
                    \FOR{each round $r=1,2,\dots, R$}
                        \STATE $\mathcal{F}^{(r)} \leftarrow$ select $K^{(r)}$ clients from $\mathcal{F}$
                        \FOR{each client $f^{(r)}_k \in \mathcal{F}^{(r)}~\textbf{\mbox{in parallel}}$}
                            \IF{$r>1$} 
                                \STATE $\{\sigma_i \}_{i \neq k}^{K^{(r)}} \leftarrow \Omega(\textbf{v}_{f_k}^{L-1}, \textbf{v}_{f_{i\neq k:K^{(r)}}}^{L-1})$
                                \STATE $\mathcal{U}_{k}^{(r)} \leftarrow \frac{\text{exp}({\epsilon \cdot \sigma_i})}{\sum_{i=1}^{K^{(r)}} \text{exp}(\epsilon \cdot \sigma_i)} \sum_{i=1}^{K^{(r)}} \mathcal{U}_i^{(r)}$
                                \STATE $\mathcal{V}_{k}^{(r)} \leftarrow \frac{\text{exp}({\epsilon \cdot \sigma_i})}{\sum_{i=1}^{K^{(r)}} \text{exp}(\epsilon \cdot \sigma_i)} \sum_{i=1}^{K^{(r)}} \mathcal{V}_i^{(r)}$
                                \STATE $\mathcal{M}_{k}^{(r)} \leftarrow \frac{\text{exp}({\epsilon \cdot \sigma_i})}{\sum_{i=1}^{K^{(r)}} \text{exp}(\epsilon \cdot \sigma_i)} \sum_{i=1}^{K^{(r)}} \mathcal{M}_i^{(r)}$
                            \ENDIF 
                            \STATE $\mathcal{U}_k^{(r+1)},  \mathcal{V}_k^{(r+1)}, \mathcal{M}_k^{(r+1)}$
                            \STATE \hspace{7em} $\leftarrow \text{RunClient} (\mathcal{U}_{k}^{(r)}, \mathcal{V}_{k}^{(r)}, \mathcal{M}_{k}^{(r)})$
                    	\ENDFOR
                    \ENDFOR 
                \STATE   \textbf{Function} RunClient($\mathcal{U}^{(r)}_k, \mathcal{V}_{k}^{(r)}, \mathcal{M}_{k}^{(r)})$)
                    \STATE $\mathcal{U}_{k}^{(r+1)} \leftarrow \mathcal{U}^{(r)}_k$, $\mathcal{V}_{k}^{(r+1)} \leftarrow \mathcal{V}^{(r)}_k$, $\mathcal{M}_{k}^{(r+1)} \leftarrow \mathcal{M}^{(r)}_k$  
                    \FOR{each local epoch $e$ from $1$ to $E$}
                        \FOR{minibatch $\mathcal{B} \in \mathcal{D}_{k}$}
                            \STATE $\theta_{\mathcal{U} \times \mathcal{V} \oplus \mathcal{M}} \leftarrow \theta_{\mathcal{U} \times \mathcal{V} \oplus \mathcal{M}}-\eta\nabla\mathcal{L}(\mathcal{B}; \theta_{\mathcal{U} \times \mathcal{V} \oplus \mathcal{M}})$
                        \ENDFOR
                    \ENDFOR
                    \STATE \textbf{return} $\mathcal{U}_{k}^{(r+1)},\mathcal{V}_{k}^{(r+1)},\mathcal{M}_{k}^{(r+1)}$
                    
        	\end{algorithmic}
    	\end{algorithm}
    \end{minipage}
\end{figure*}

\section{Dataset Configurations}
\label{appdx:dataset}
In this section, we describe detailed configurations for datasets that we used in label- and domain-heterogeneous scenarios.

\subsection{Label Heterogeneous Scenario}

We use CIFAR-10 and SVHN for the label-heterogeneous scenario. We first split each dataset into train, validation, and test sets for CIFAR-10 ($48,000$/$6,000$/$6,000$) and SVHN ($79,431$/$9,929$/$9,929$). We then split the train set into $K$ local partitions $\mathcal{P}_{1:20}$ ($K$=$20$) for iid partitions (all instances in each class are evenly distributed to all clients) or for the non-iid partitions (instances in each class are sampled from Dirichlet distribution with $\alpha$=$0.5$). We further permute the labels for each class per local partition $\mathcal{P}_k$ for permuted iid and permuted non-iid scenarios. We use different random seed per client, i.e. $\textit{fixed global seed} + \textit{client id}$, for example, $1234 + 0$ for Client 1 and $1234 + 19$ for Client 20. We provide permutations of labels that we used for each dataset in Table~\ref{tbl:dataset_1}.

\begin{table*}[h]
    \caption{\textbf{Label permutations for label-heterogeneous scenario} We provide permutations of labels for each dataset. These permutations are randomly generated based on different seeds, calculated by $\textit{fixed global seed} + \textit{client id}$. }
    \vspace{-0.1in}
    \resizebox{\linewidth}{!}{
    \begin{tabular}{c c c c cccccccccccccccccccc}
        \toprule 
        \midrule
        \multirow{2}{*}{Dataset} & \multirow{2}{*}{Class} & Original & \multicolumn{20}{c}{ Client No. } \\
        & & Labels  & 1 & 2 & 3 & 4 & 5 & 6 & 7 & 8 & 9 & 10 & 11 & 12 & 13 & 14 & 15 & 16 & 17 & 18 & 19 & 20 \\
        \midrule
        \midrule
        \multirow{10}{*}{CIFAR-10} & Airplane & 0 & 2 & 5 & 3 & 0 & 8 & 2 & 4 & 4 & 2 & 2 & 0 & 6 & 1 & 8 & 4 & 0 & 0 & 6 & 9 & 7 \\
        & Automobile & 1 & 8 & 4 & 1 & 5 & 1 & 8 & 1 & 0 & 9 & 7 & 7 & 3 & 6 & 2 & 0 & 4 & 1 & 8 & 1 & 4 \\
        & Bird & 2 & 3 & 0 & 5 & 3 & 4 & 9 & 9 & 5 & 4 & 1 & 5 & 5 & 3 & 0 & 1 & 6 & 3 & 7 & 6 & 5 \\
        & Cat & 3 &  5 & 9 & 0 & 7 & 9 & 3 & 8 & 8 & 7 & 6 & 2 & 7 & 8 & 7 & 6 & 7 & 4 & 4 & 3 & 8 \\
        & Deer & 4 & 6 & 2 & 6 & 9 & 6 & 5 & 2 & 1 & 0 & 9 & 6 & 1 & 7 & 3 & 7 & 5 & 5 & 3 & 8 & 9 \\
        & Dog & 5 & 4 & 1 & 4 & 8 & 5 & 6 & 6 & 6 & 3 & 4 & 9 & 4 & 4 & 4 & 5 & 9 & 8 & 1 & 2 & 0 \\
        & Frog & 6 & 9 & 3 & 2 & 1 & 2 & 0 & 3 & 2 & 6 & 3 & 3 & 0 & 5 & 6 & 2 & 1 & 7 & 2 & 5 & 1 \\
        & Horse & 7 & 0 & 7 & 9 & 4 & 3 & 7 & 0 & 3 & 5 & 0 & 1 & 2 & 9 & 5 & 3 & 2 & 9 & 5 & 0 & 6 \\
        & Ship & 8 & 1 & 8 & 7 & 6 & 7 & 4 & 5 & 7 & 8 & 5 & 8 & 9 & 0 & 9 & 8 & 3 & 6 & 0 & 7 & 2 \\
        & Truck & 9 & 7 & 6 & 8 & 2 & 0 & 1 & 7 & 9 & 1 & 8 & 4 & 8 & 2 & 1 & 9 & 8 & 2 & 9 & 4 & 3 \\
        \midrule
        \multirow{10}{*}{SVHN} & Digit 0 & 10 & 2 & 5 & 3 & 0 & 8 & 2 & 4 & 4 & 2 & 2 & 0 & 6 & 1 & 8 & 4 & 0 & 0 & 6 & 9 & 7 \\
        & Digit 1 & 1 & 8 & 4 & 1 & 5 & 1 & 8 & 1 & 0 & 9 & 7 & 7 & 3 & 6 & 2 & 0 & 4 & 1 & 8 & 1 & 4 \\
        & Digit 2 & 2 & 3 & 0 & 5 & 3 & 4 & 9 & 9 & 5 & 4 & 1 & 5 & 5 & 3 & 0 & 1 & 6 & 3 & 7 & 6 & 5 \\
        & Digit 3 & 3 &  5 & 9 & 0 & 7 & 9 & 3 & 8 & 8 & 7 & 6 & 2 & 7 & 8 & 7 & 6 & 7 & 4 & 4 & 3 & 8 \\
        & Digit 4 & 4 & 6 & 2 & 6 & 9 & 6 & 5 & 2 & 1 & 0 & 9 & 6 & 1 & 7 & 3 & 7 & 5 & 5 & 3 & 8 & 9 \\
        & Digit 5 & 5 & 4 & 1 & 4 & 8 & 5 & 6 & 6 & 6 & 3 & 4 & 9 & 4 & 4 & 4 & 5 & 9 & 8 & 1 & 2 & 0 \\
        & Digit 6 & 6 & 9 & 3 & 2 & 1 & 2 & 0 & 3 & 2 & 6 & 3 & 3 & 0 & 5 & 6 & 2 & 1 & 7 & 2 & 5 & 1 \\
        & Digit 7 & 7 & 0 & 7 & 9 & 4 & 3 & 7 & 0 & 3 & 5 & 0 & 1 & 2 & 9 & 5 & 3 & 2 & 9 & 5 & 0 & 6 \\
        & Digit 8 & 8 & 1 & 8 & 7 & 6 & 7 & 4 & 5 & 7 & 8 & 5 & 8 & 9 & 0 & 9 & 8 & 3 & 6 & 0 & 7 & 2 \\
        & Digit 9 & 9 & 7 & 6 & 8 & 2 & 0 & 1 & 7 & 9 & 1 & 8 & 4 & 8 & 2 & 1 & 9 & 8 & 2 & 9 & 4 & 3 \\
        \midrule
        \bottomrule 
    \label{tbl:dataset_1}
    \end{tabular}}
\end{table*}

\vspace{-0.3in}
\subsection{Domain Heterogeneous Scenario}

We use CIFAR-100 datasets ($60,000$) and create five sub-datasets grouped by $10$ similar classes, such as Fruits\&Foods ($6,000$), Transport ($6,000$), Household Objects ($6,000$), Animals ($6,000$), Trees\&Flowers ($6,000$). We then split train ($4,800$), test ($600$), validation ($600$) sets for each sub-datset. To have $20$ clients in total, we assign four clients per subdataset, and split each train set into $4$ partitions, making a single partition contains $1,200$ instances. Additionally, we further permute the labels for those $20$ partitions to simulate more realistic scenarios where labeling schemes are not synchronized across all clients even in the same domain (sub-dataset). We provide class division and label permutation information in Table~\ref{tbl:dataset_2}.

\section{Implementation \& Training Details} 
\label{appdx:details}
In this section, we provide detailed implementation and training details that are not described in the main document.

\subsection{ResNet-9 Architecture}

\begin{wraptable}{r}{8.25cm}
    \vspace{-0.4in}
    \caption{\textbf{Detailed ResNet-9 Architecture}}
    \vspace{-0.1in}
    \resizebox{\linewidth}{!}{
    \begin{tabular}{cccccccc}
        \toprule
        \midrule
        {\textbf{Layer}} & Input & Output & Filter Size & Stride & Dimension of $\textbf{W}^l$ \\ 
        \midrule 
        Conv 1 & 3 & 64 & 3 & 1 & $64\times3\times3\times3$ \\
        Conv 2 & 64 & 128 & 5 & 2 & $128\times64\times5\times5$ \\
        Conv 3 & 128 & 128 & 3 & 1 & $128\times128\times3\times3$ \\
        Conv 4 & 128 & 128 & 3 & 1 & $128\times128\times3\times3$ \\
        Conv 5 & 128 & 256 & 3 & 1 & $256\times128\times3\times3$ \\
        Conv 6 & 256 & 256 & 3 & 1 & $256\times256\times3\times3$ \\
        Conv 7 & 256 & 256 & 3 & 1 & $256\times256\times3\times3$ \\
        Conv 8 & 256 & 256 & 3 & 1 & $256\times256\times3\times3$ \\
        FC 1 & 256 & $C$ & - & - & $256 \times C$ \\
        \midrule \bottomrule 
    \end{tabular}}
    \label{tbl:resnet9}
\end{wraptable}

We use ResNet-9 architecture consisting of eight convolutional layers and one fully connected layer as a classifier, as described in Table~\ref{tbl:resnet9}. We use max pooling with size $2$ after Conv $5$ and an adaptive max pooling after Conv $8$ to make output width 1 for the following FC layer. The total number of parameters of the model is $2.57 M$. As we use \texttt{PyTorch} framework for implementation and the default data type of tensor of the framework is 32-bits floating point, the model size can be calculated as $2.57 \times 4 = 10.28$ Mbytes.

\begin{table*}[t!]
    \caption{\textbf{Class division and label permutation information for domain-heterogeneous scenario} We provide class division information and label permutation details for each domain. These permutations are randomly generated based on the same method used in label-heterogeneous scenario using different seeds, i.e. $\textit{fixed global seed} + \textit{client id}$.}
    \resizebox{\linewidth}{!}{
    \begin{tabular}{c c c c cccccccccccccccccccc}
        \toprule 
        \midrule
        \multirow{2}{*}{Domain} & \multirow{2}{*}{Class} & Original & \multicolumn{20}{c}{ Client No. } \\
        & & Labels  & 1 & 2 & 3 & 4 & 5 & 6 & 7 & 8 & 9 & 10 & 11 & 12 & 13 & 14 & 15 & 16 & 17 & 18 & 19 & 20 \\
        \midrule
        \midrule
        \multirow{10}{*}{\shortstack{Household \\ Objects } } 
        & Bed & 5 & 2 & 5 & 3 & 0 & - & - & - & - & - & - & - & - & - & - & - & - & - & - & - & - \\
        & Chair & 20 & 8 & 4 & 1 & 5 & - & - & - & - & - & - & - & - & - & - & - & - & - & - & - & - \\
        & Couch & 22 & 3 & 0 & 5 & 3 & - & - & - & - & - & - & - & - & - & - & - & - & - & - & - & - \\
        & Table & 25 & 5 & 9 & 0 & 7 & - & - & - & - & - & - & - & - & - & - & - & - & - & - & - & - \\
        & Wardrobe & 39 & 6 & 2 & 6 & 9 & - & - & - & - & - & - & - & - & - & - & - & - & - & - & - & - \\
        & Clock & 40 & 4 & 1 & 4 & 8 & - & - & - & - & - & - & - & - & - & - & - & - & - & - & - & - \\
        & Keyboard & 84 & 9 & 3 & 2 & 1 & - & - & - & - & - & - & - & - & - & - & - & - & - & - & - & - \\
        & Lamp & 86 & 0 & 7 & 9 & 4 & - & - & - & - & - & - & - & - & - & - & - & - & - & - & - & - \\
        & Telephone & 87 & 1 & 8 & 7 & 6 & - & - & - & - & - & - & - & - & - & - & - & - & - & - & - & - \\
        & Television & 94 & 7 & 6 & 8 & 2 & - & - & - & - & - & - & - & - & - & - & - & - & - & - & - & - \\
        \midrule
        \multirow{10}{*}{\shortstack{Fruits \\ \& Foods } }
        & Apple & 0 & - & - & - & - & 8 & 2 & 4 & 4 & - & - & - & - & - & - & - & - & - & - & - \\
        & Mushroom & 9 & - & - & - & - & 1 & 8 & 1 & 0 & - & - & - & - & - & - & - & - & - & - & - \\
        & Orange & 10 & - & - & - & - & 4 & 9 & 9 & 5 & - & - & - & - & - & - & - & - & - & - & - & - \\
        & Pear & 16 &  - & - & - & - & 9 & 3 & 8 & 8 & - & - & - & - & - & - & - & - & - & - & - & - \\
        & Sweet Pepper & 28 & - & - & - & - & 6 & 5 & 2 & 1 & - & - & - & - & - & - & - & - & - & - & - & - \\
        & Bottle & 51 & - & - & - & - & 5 & 6 & 6 & 6 & - & - & - & - & - & - & - & - & - & - & - & - \\
        & Bowl & 53 & - & - & - & - & 2 & 0 & 3 & 2 & - & - & - & - & - & - & - & - & - & - & - & - \\
        & Can & 57 & - & - & - & - & 3 & 7 & 0 & 3 & - & - & - & - & - & - & - & - & - & - & - & - \\
        & Cup & 61 & - & - & - & - & 7 & 4 & 5 & 7 & - & - & - & - & - & - & - & - & - & - & - & - \\
        & Plate & 83 & - & - & - & - & 0 & 1 & 7 & 9 & - & - & - & - & - & - & - & - & - & - & - & - \\
        \midrule
        \multirow{10}{*}{\shortstack{Trees \&  \\ Flowers } }
        & Orchid & 47  & - & - & - & - & - & - & - & - & 2 & 2 & 0 & 6 & - & - & - & - & - & - & - & - \\
        & Poppy & 52  & - & - & - & - & - & - & - & - & 9 & 7 & 7 & 3 & - & - & - & - & - & - & - \\
        & Rose & 54  & - & - & - & - & - & - & - & - & 4 & 1 & 5 & 5 & - & - & - & - & - & - & - & - \\
        & Sunflower & 56  &  - & - & - & - & - & - & - & - & 7 & 6 & 2 & 7 & - & - & - & - & - & - & - & - \\
        & Tulip & 59  & - & - & - & - & - & - & - & - & 0 & 9 & 6 & 1 & - & - & - & - & - & - & - & - \\
        & Maple Tree & 62  & - & - & - & - & - & - & - & - & 3 & 4 & 9 & 4 & - & - & - & - & - & - & - & - \\
        & Oak Tree & 70  & - & - & - & - & - & - & - & - & 6 & 3 & 3 & 0 & - & - & - & - & - & - & - & - \\
        & Palm Tree & 82  & - & - & - & - & - & - & - & - & 5 & 0 & 1 & 2 & - & - & - & - & - & - & - & - \\
        & Pine Tree & 92  & - & - & - & - & - & - & - & - & 8 & 5 & 8 & 9 & - & - & - & - & - & - & - & - \\
        & Willow Tree & 96 & - & - & - & - & - & - & - & - & 1 & 8 & 4 & 8 & - & - & - & - & - & - & - & - \\
        \midrule
        \multirow{10}{*}{\shortstack{Transport } }
        & Lawn Mower & 8 & - & - & - & - & - & - & - & - & - & - & - & - & 1 & 8 & 4 & 0 & - & - & - & - \\
        & Rocket & 13 & - & - & - & - & - & - & - & - & - & - & - & - & 6 & 2 & 0 & 4 & - & - & - & - \\
        & Streetcar & 41 & - & - & - & - & - & - & - & - & - & - & - & - & 3 & 0 & 1 & 6 & - & - & - & - \\
        & Tank & 48 &  - & - & - & - & - & - & - & - & - & - & - & - & 8 & 7 & 6 & 7 & - & - & - & - \\
        & Tractor & 58 & - & - & - & - & - & - & - & - & - & - & - & - & 7 & 3 & 7 & 5 & - & - & - & - \\
        & Bicycle & 69 & - & - & - & - & - & - & - & - & - & - & - & - & 4 & 4 & 5 & 9 & - & - & - & - \\
        & Bus & 81 & - & - & - & - & - & - & - & - & - & - & - & - & 5 & 6 & 2 & 1 & - & - & - & - \\
        & Motorcycle & 85 & - & - & - & - & - & - & - & - & - & - & - & - & 9 & 5 & 3 & 2 & - & - & - & - \\
        & Pickup Truck & 89 & - & - & - & - & - & - & - & - & - & - & - & - & 0 & 9 & 8 & 3 & - & - & - & - \\
        & Train & 90 & - & - & - & - & - & - & - & - & - & - & - & - & 2 & 1 & 9 & 8 & - & - & - & - \\
        \midrule
        \multirow{10}{*}{\shortstack{Animals} }
        & Fox & 3 & - & - & - & - & - & - & - & - & - & - & - & - & - & - & - & - & 0 & 6 & 9 & 7 \\
        & Porcupine & 34 & - & - & - & - & - & - & - & - & - & - & - & - & - & - & - & - & 1 & 8 & 1 & 4 \\
        & Possum & 42 & - & - & - & - & - & - & - & - & - & - & - & - & - & - & - & - & 3 & 7 & 6 & 5 \\
        & Raccoon & 43 &  - & - & - & - & - & - & - & - & - & - & - & - & - & - & - & - & 4 & 4 & 3 & 8 \\
        & Skunk & 63 & - & - & - & - & - & - & - & - & - & - & - & - & - & - & - & - & 5 & 3 & 8 & 9 \\
        & Bear & 64 & - & - & - & - & - & - & - & - & - & - & - & - & - & - & - & - & 8 & 1 & 2 & 0 \\
        & Leopard & 66 & - & - & - & - & - & - & - & - & - & - & - & - & - & - & - & - & 7 & 2 & 5 & 1 \\
        & Lion & 75 & - & - & - & - & - & - & - & - & - & - & - & - & - & - & - & - & 9 & 5 & 0 & 6 \\
        & Tiger & 88 & - & - & - & - & - & - & - & - & - & - & - & - & - & - & - & - & 6 & 0 & 7 & 2 \\
        & Wolf & 97 & - & - & - & - & - & - & - & - & - & - & - & - & - & - & - & - & 2 & 9 & 4 & 3 \\
        \midrule
        \bottomrule 
    \label{tbl:dataset_2}
    \end{tabular}}
\end{table*}

\subsection{Calculation of Communication Cost}
We measure the communication cost by $\{(P_{S2C}+P_{C2S}) \times 4 \}_{byte} \times K \times R$, where $P_{S2C}$ is number of server-to-client transmitted parameters and $P_{C2S}$ is number of client-to-server transmitted parameters. Depending on the FL algorithms, $P_{S2C}$ and  $P_{C2S}$ are differently calculated. For example, \texttt{FedFOMO} downloads few random models from server ($10$ as default, reported in the paper) but sends only single local model to server. Our \texttt{Factorized-FL} only sends the  small portion of model parameters, $\mathcal{U}$ and $\textbf{v}^{L-1}$, to server, while receiving a single set of $\mathcal{U}$ from server. 

\subsection{Training Details}

As default, all training configurations are equally set across all models, unless otherwise stated to ensure stricter fairness. We use ResNet-9 architecture as local backbone networks and train them on $32\times32$ sized images with $256$ for batch size. We apply data augmentations, i.e. cropping, flipping, jittering, etc, during training. Optimizer that we used is Stochastic Gradient Descent (SGD). We set $1$e-$3$ for learning rate, $1$e-$6$ for weight decay, and  $0.9$ for momentum. For baseline models, we use the reported hyper-parameters as default, or we adjust hyper-parameters so that they show the best performance for fairness. For ours and \texttt{pFedPara}, the model capacity is adjusted to around $90\%$ - $99\%$ of the original size, as we fairly compare with other methods that use full capacity ($2.57M$ number of parameters). For ours, we use [$5$e-$4$, $1$e-$3$] for $\lambda_{\text{sparsity}}$, [$0$-$0.75$] for $\tau$, [$1$, $20$] for $\epsilon$. 

\section{Additional Experimental Results}
\label{appdx:experiment}





\subsection{Sparsity Analysis on FL Scenarios}
\label{appdx:sparisty}

\begin{figure*}[h!]
\begin{tabular}{cc}
     \includegraphics[width=0.5\linewidth]{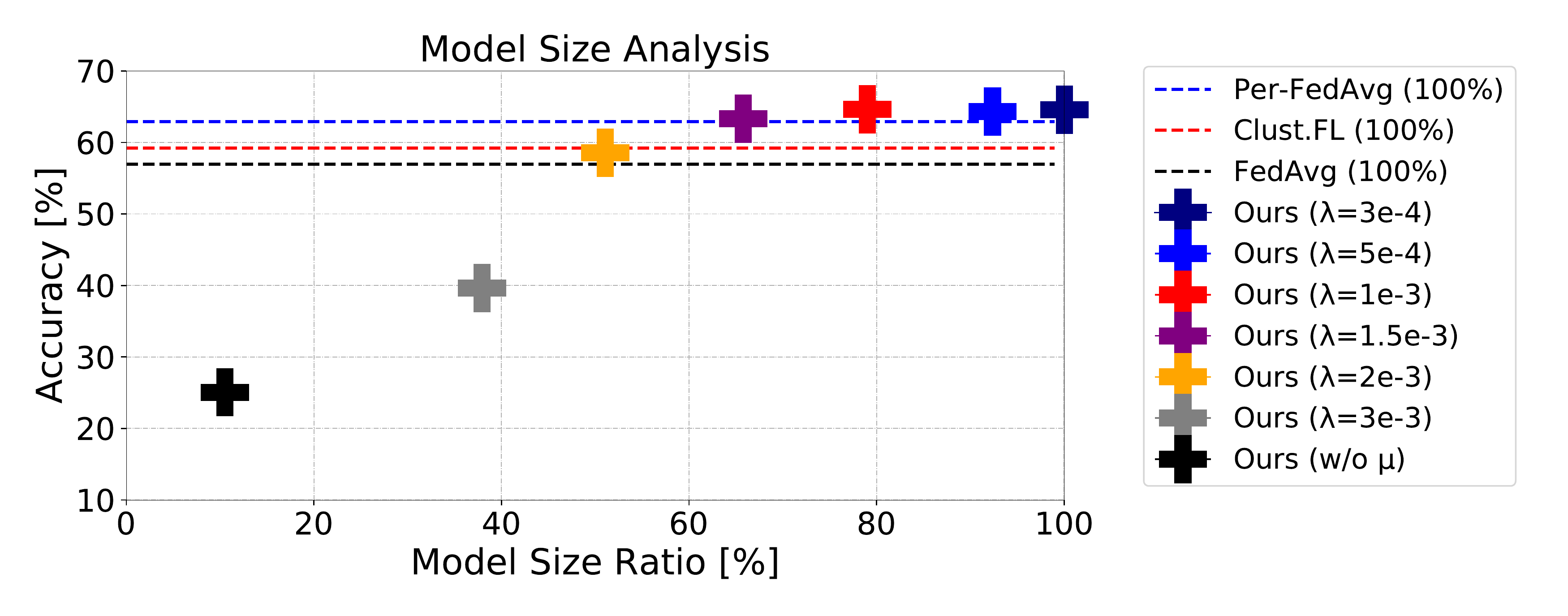} & 
     \hspace{-0.1in} \includegraphics[width=0.5\linewidth]{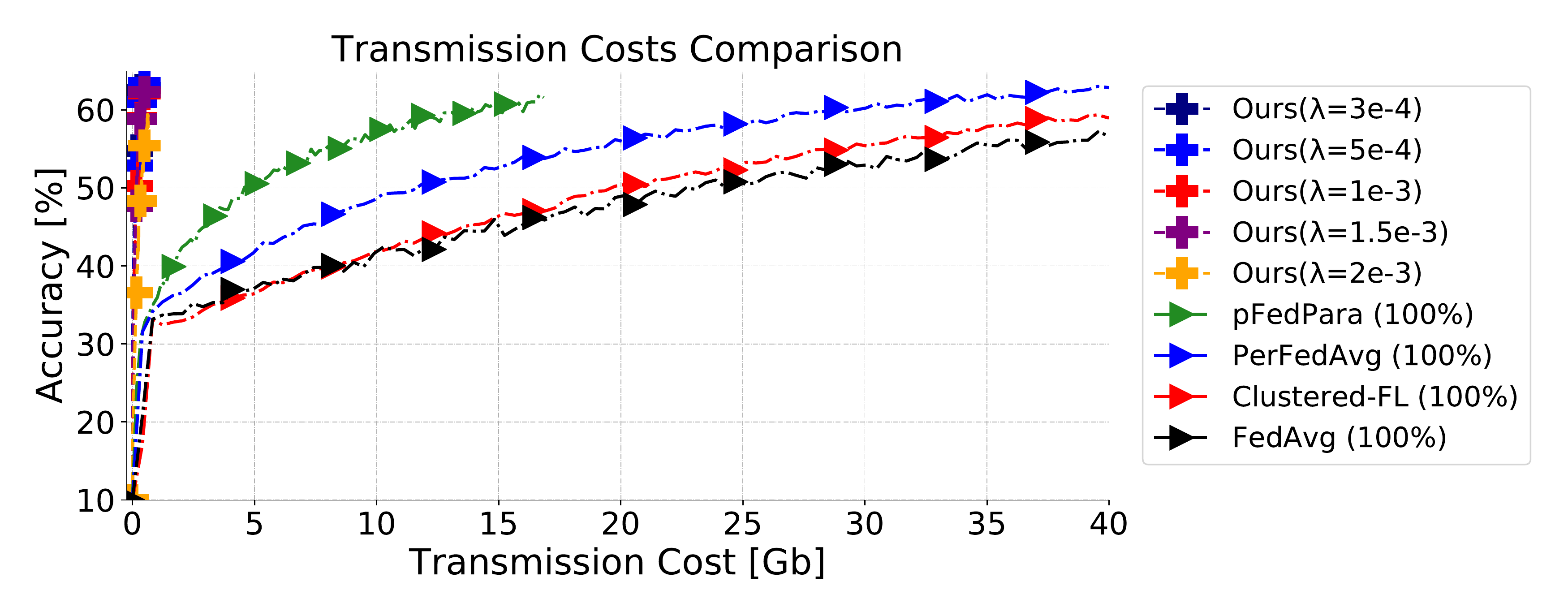}\\
     (a) Model size ratio controlled by $\lambda_{\text{sparsity}}$ &
     (b) Communication Costs 
\end{tabular}
\caption{\textbf{Model size and communication costs comparison} (a) we plot accuracy over model size on domain heterogeneous scenario. (b) we plot accuracy over transmission costs on domain heterogeneous scenario.}
\label{fig:model_size}
\end{figure*}

In the main document, we show the effect of model size and sparsity controlled by  $\lambda_{\text{sparsity}}$ for a single model. In this section, we analyze it under federated learning scenario. In Figure~\ref{fig:model_size} (a), we show the performance over model size in domain heterogeneous scenario. As shown, our method show superior performance even with around $65\%$ of the model size over the baseline model that achieves the best performance (\texttt{Per-FedAvg}) amongst other baseline models. With $50\%$ sparsity, ours still shows competitive performance compared to~\texttt{Clustered-FL} and \texttt{FedAvg}, while it starts being significantly degenerated when sparsity becomes over $50\%$. 

In Figure~\ref{fig:model_size} (b), we show accuracy over communication costs. Note that, in our method, the model size is not really related to the communication costs since we send very small portion of model parameters. For example, even though we use almost full model size ($\lambda_{\text{sparsity}}$=$3e$-$4$), our communication cost is significantly lesser than the other baseline models, as shown in the figure.

\subsection{Additional Results}

\begin{figure}
    \centering
    \includegraphics[angle=270, origin=c, width=0.25\textwidth]{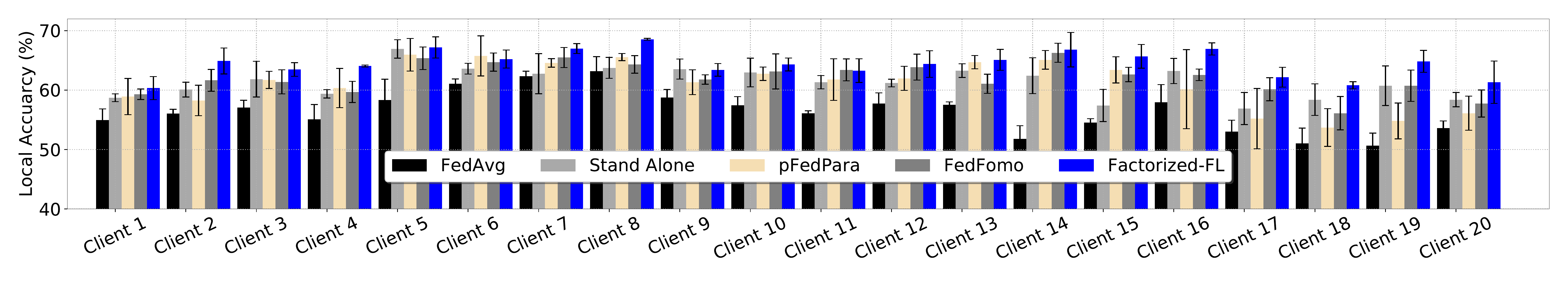}
    \caption{\textbf{Performance of all $20$ clients in domain heterogeneous scenario}: We plot performance of $20$ clients in domain-heterogeneous scenario, of which results are corresponding to Table~\ref{tbl:permuted} (Bottom). }
    \label{fig:bar}
\end{figure}

For label-heterogeneous FL scenario (Table~\ref{tbl:permuted} (Top), we provide test accuracy curves over communication rounds and transmission costs for results of CIFAR-10 and SVHN with stardard iid/non-iid and permuted iid/non-iid partitions in Figure~\ref{fig:all}. For domain-heterogeneous FL scenario (Table~\ref{tbl:permuted} (Bottom)), we provide performance of $20$ clients In Figure~~\ref{fig:bar}.

\begin{figure*}
    \centering
    
    \begin{tabular}{c c }
         
         \includegraphics[width=0.38\textwidth]{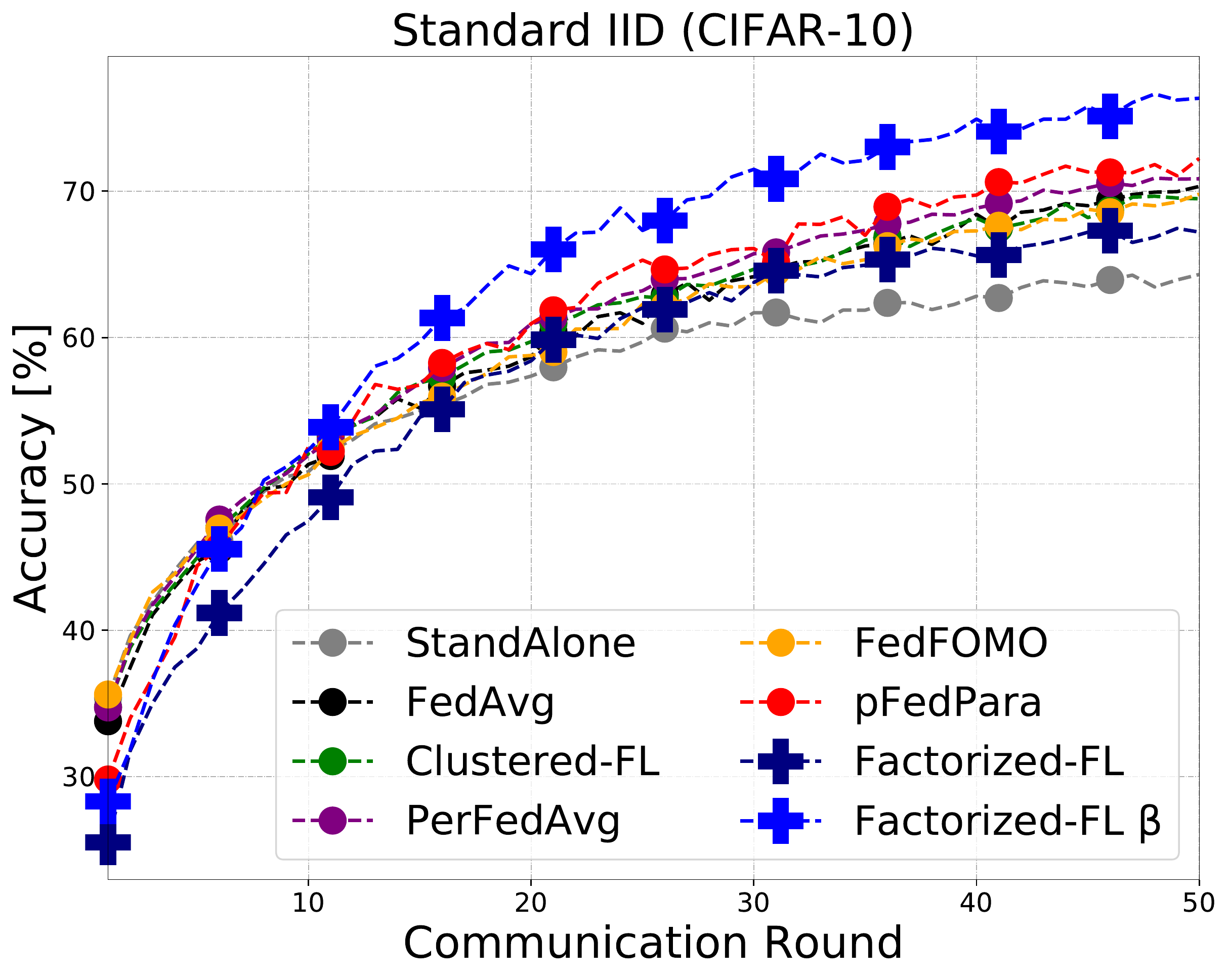} & \includegraphics[width=0.38\textwidth]{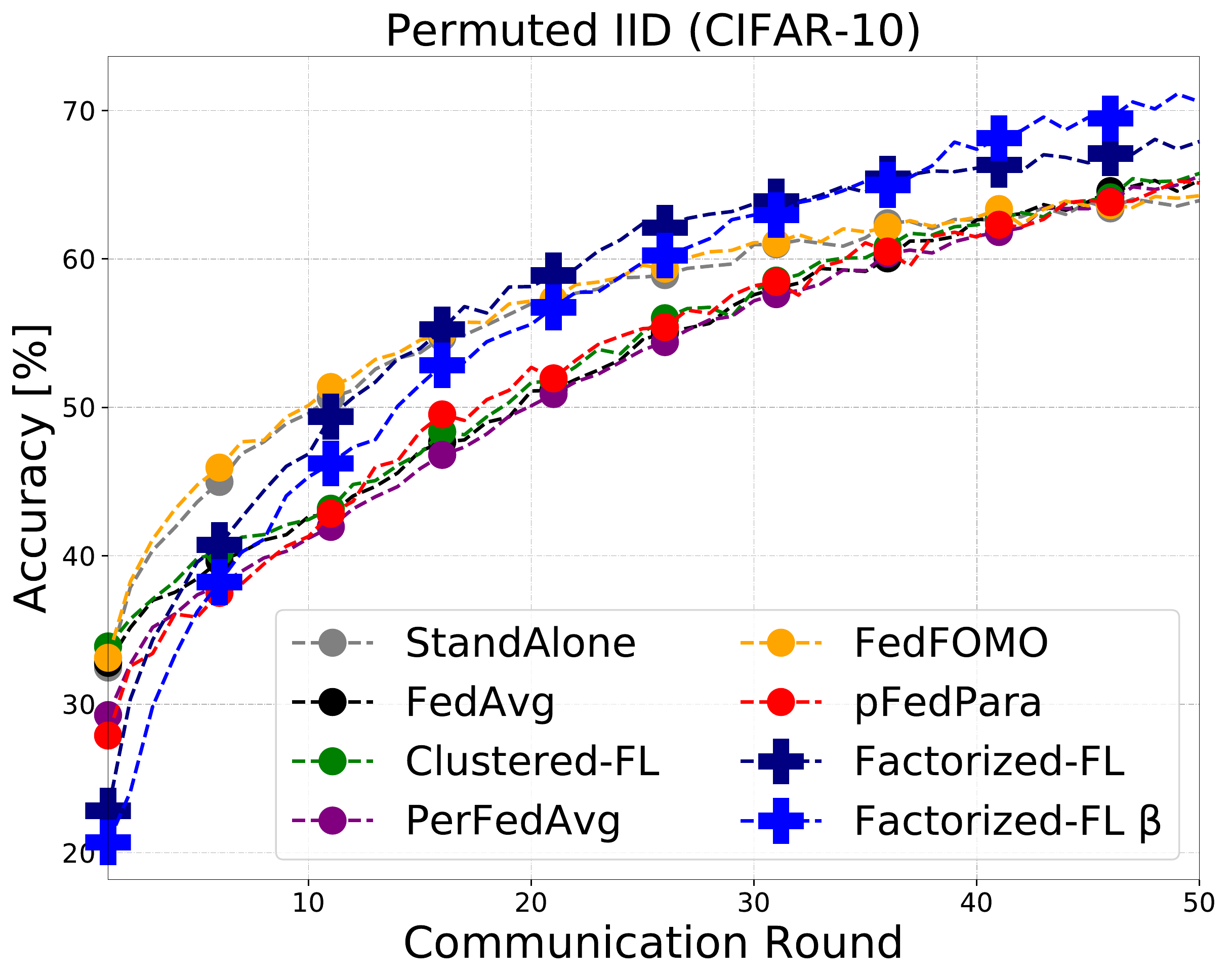} \\
         (a) Standard IID (CIFAR-10) & (b) Permuted IID (CIFAR-10) \\
         
         \includegraphics[width=0.38\textwidth]{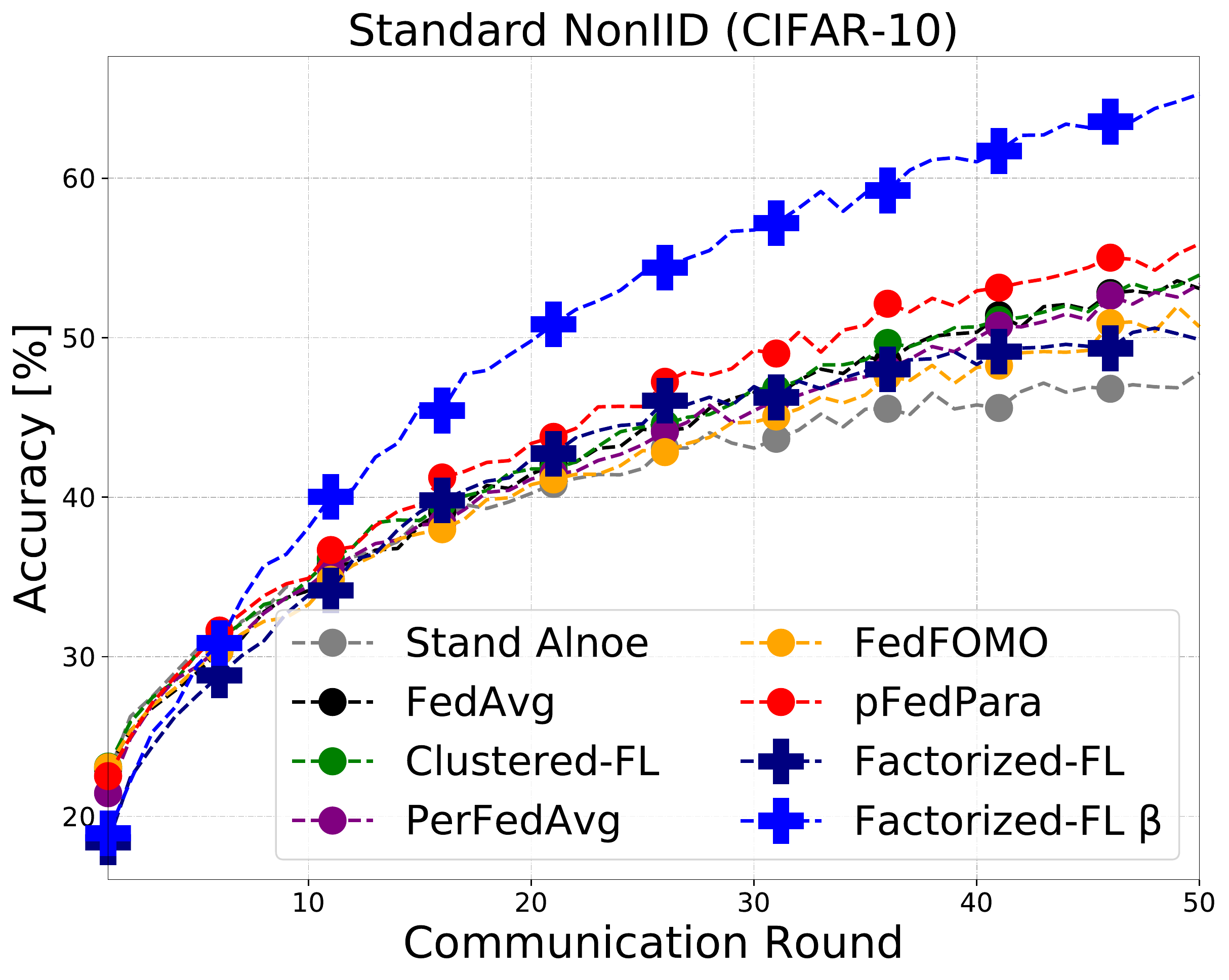} & \includegraphics[width=0.38\textwidth]{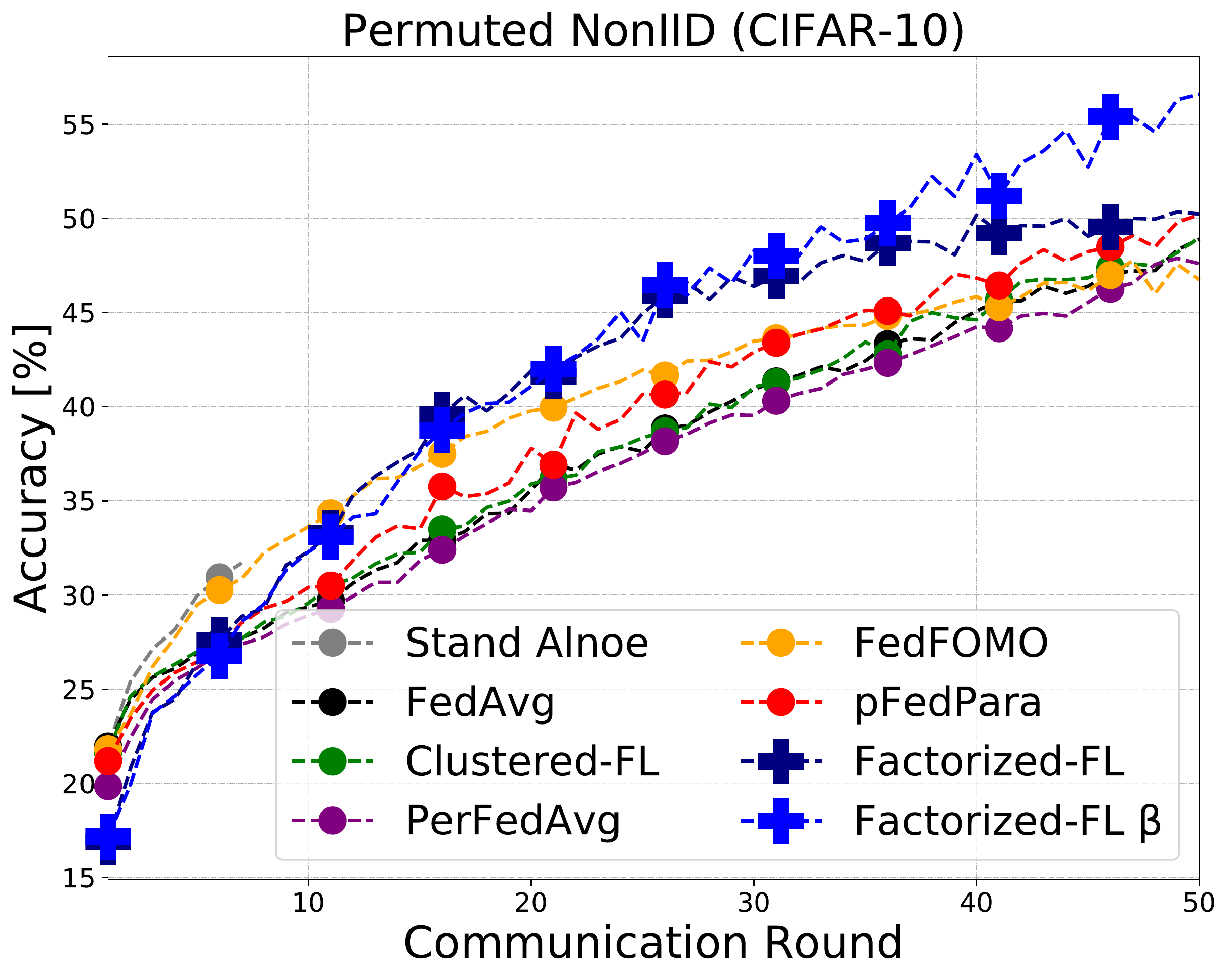} \\
         (c) Standard NonIID (CIFAR-10) & (d) Permuted NonIID (CIFAR-10) \\
         
         \includegraphics[width=0.38\textwidth]{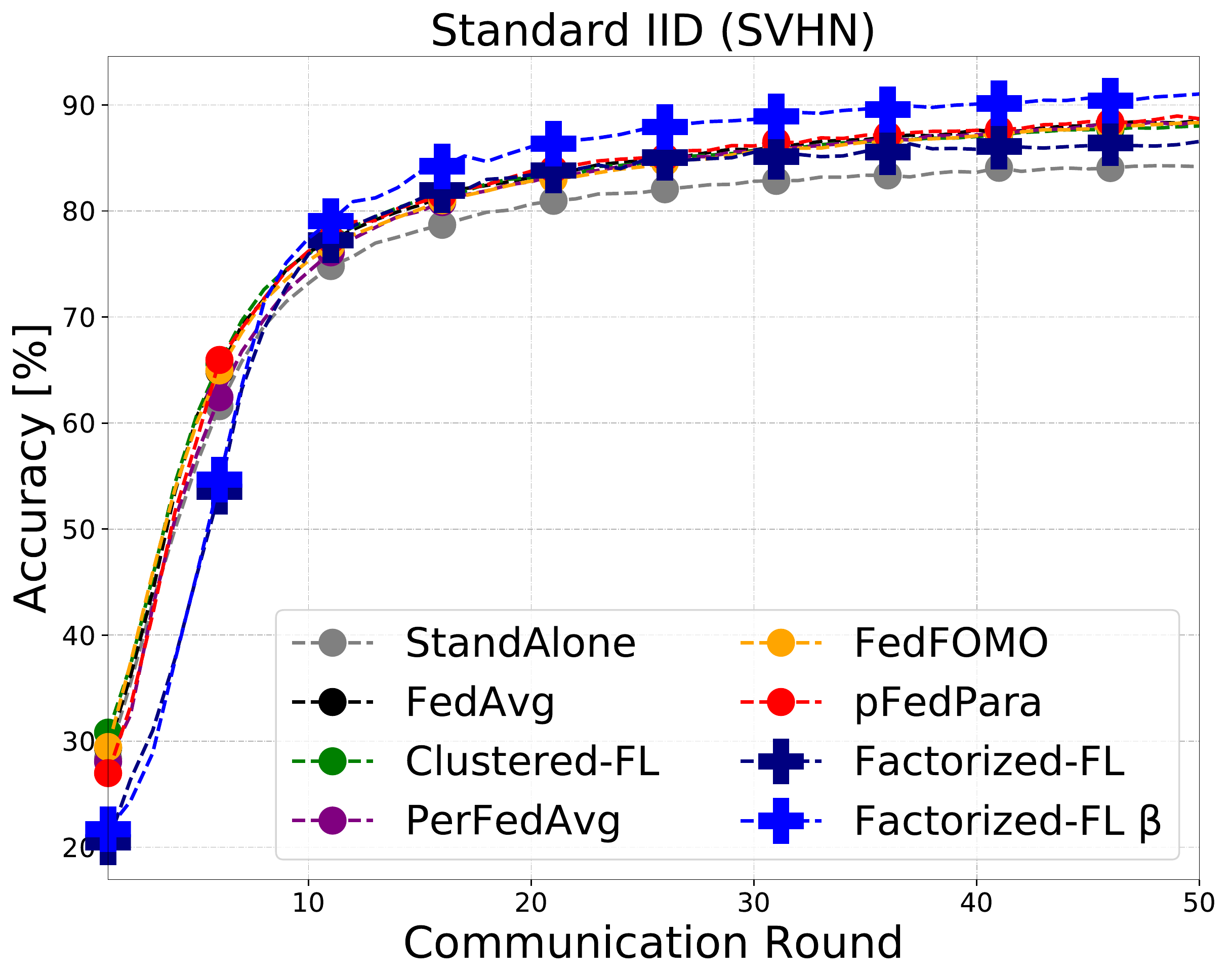} & \includegraphics[width=0.38\textwidth]{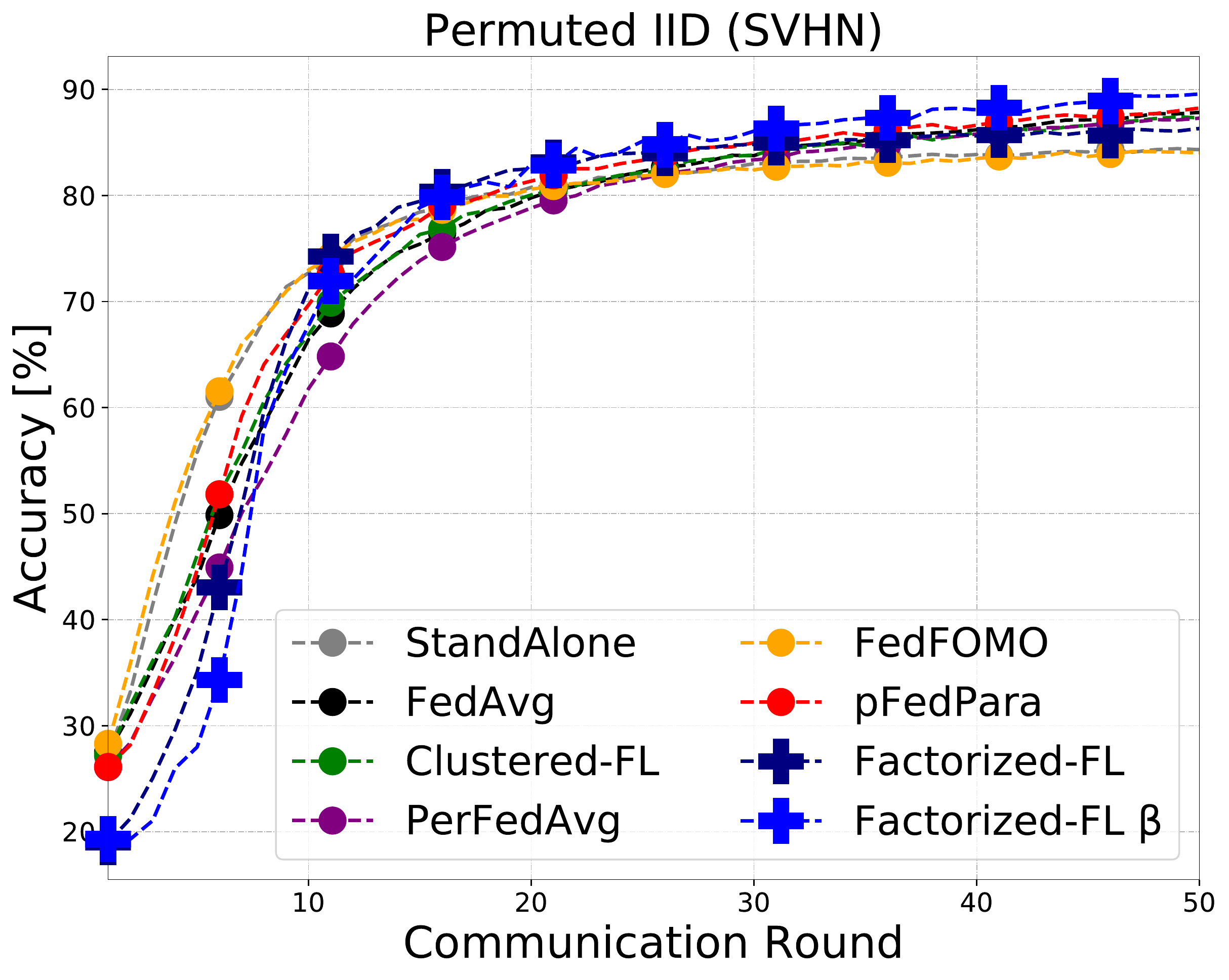} \\
         (e) Standard IID (SVHN) & (f) Permuted IID (SVHN) \\
         
         \includegraphics[width=0.38\textwidth]{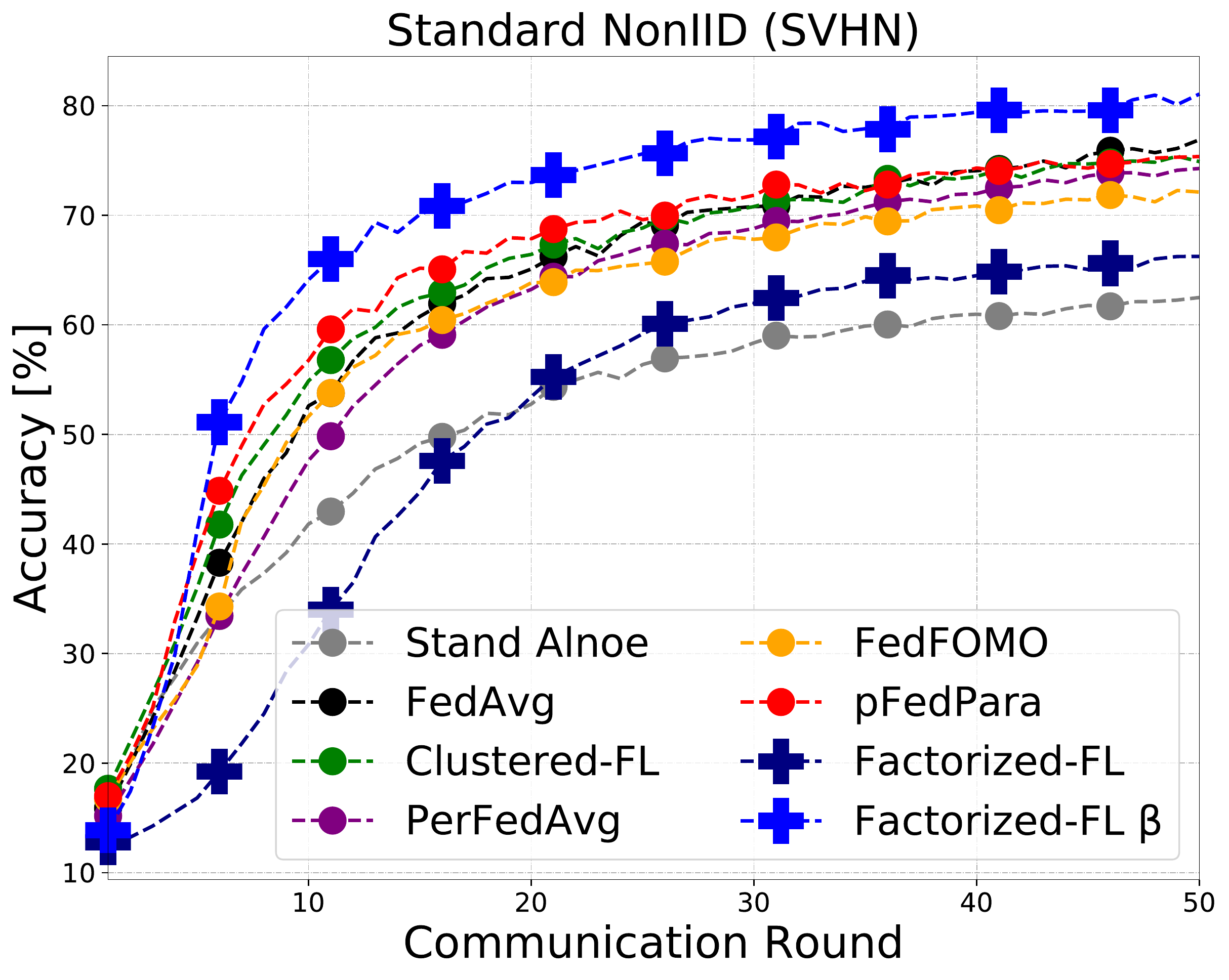} &  \includegraphics[width=0.38\textwidth]{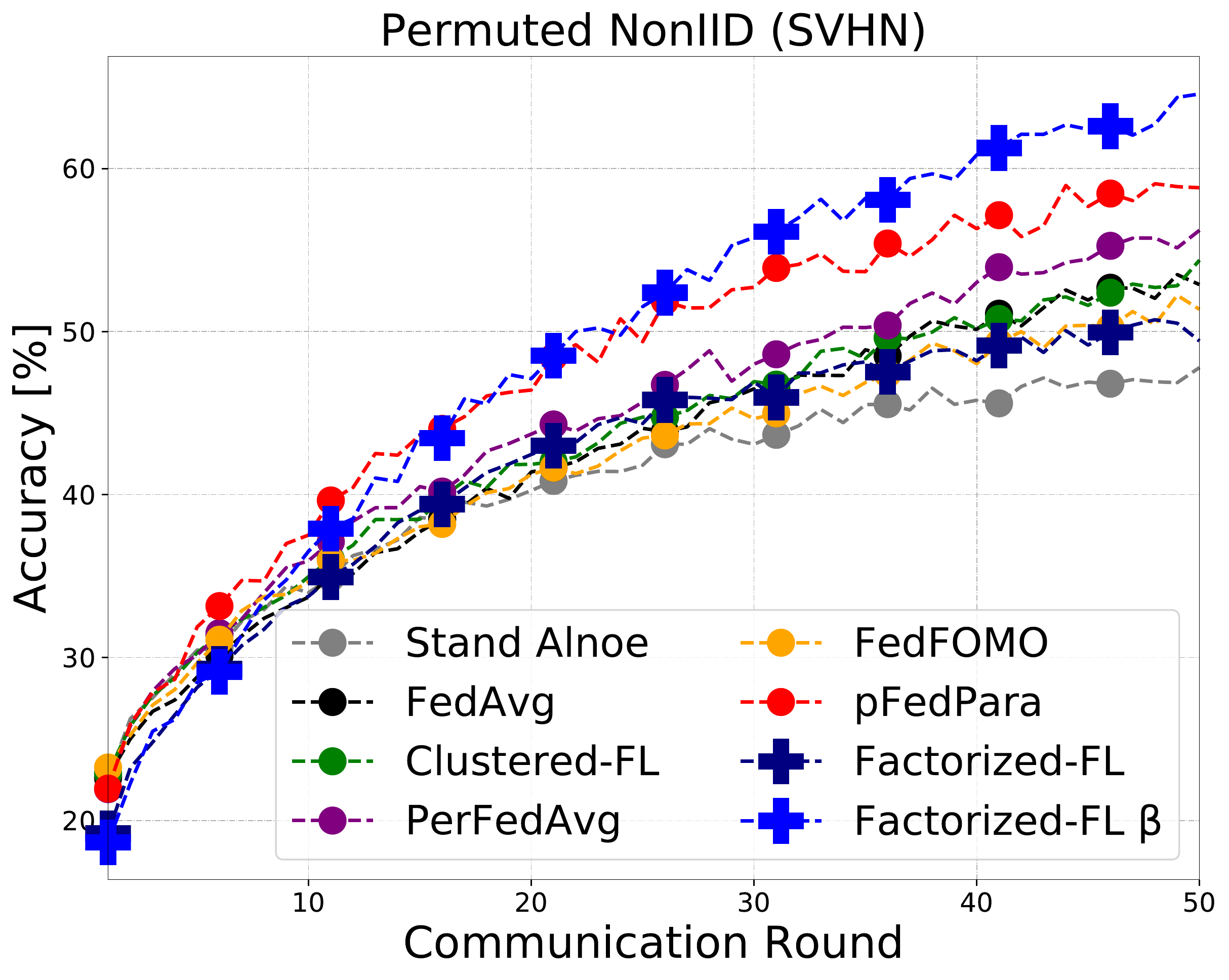} \\
         (g) Standard NonIID (SVHN) & (h) Permuted NonIID (SVHN) \\
         
    \end{tabular}
    \caption{\textbf{Test accuracy curves over communication round for standard federated learning and label-heterogeneous FL scenario}: We provide test accuracy curves on CIFAR-10 and SVHN in standard iid/non-iid and permuted iid/non-iid partitions ($E$=$5$,$R$=$50$). }
    \label{fig:all}
\end{figure*}
\begin{figure*}
    \centering
    
    \begin{tabular}{c c }
         
         \includegraphics[width=0.38\textwidth]{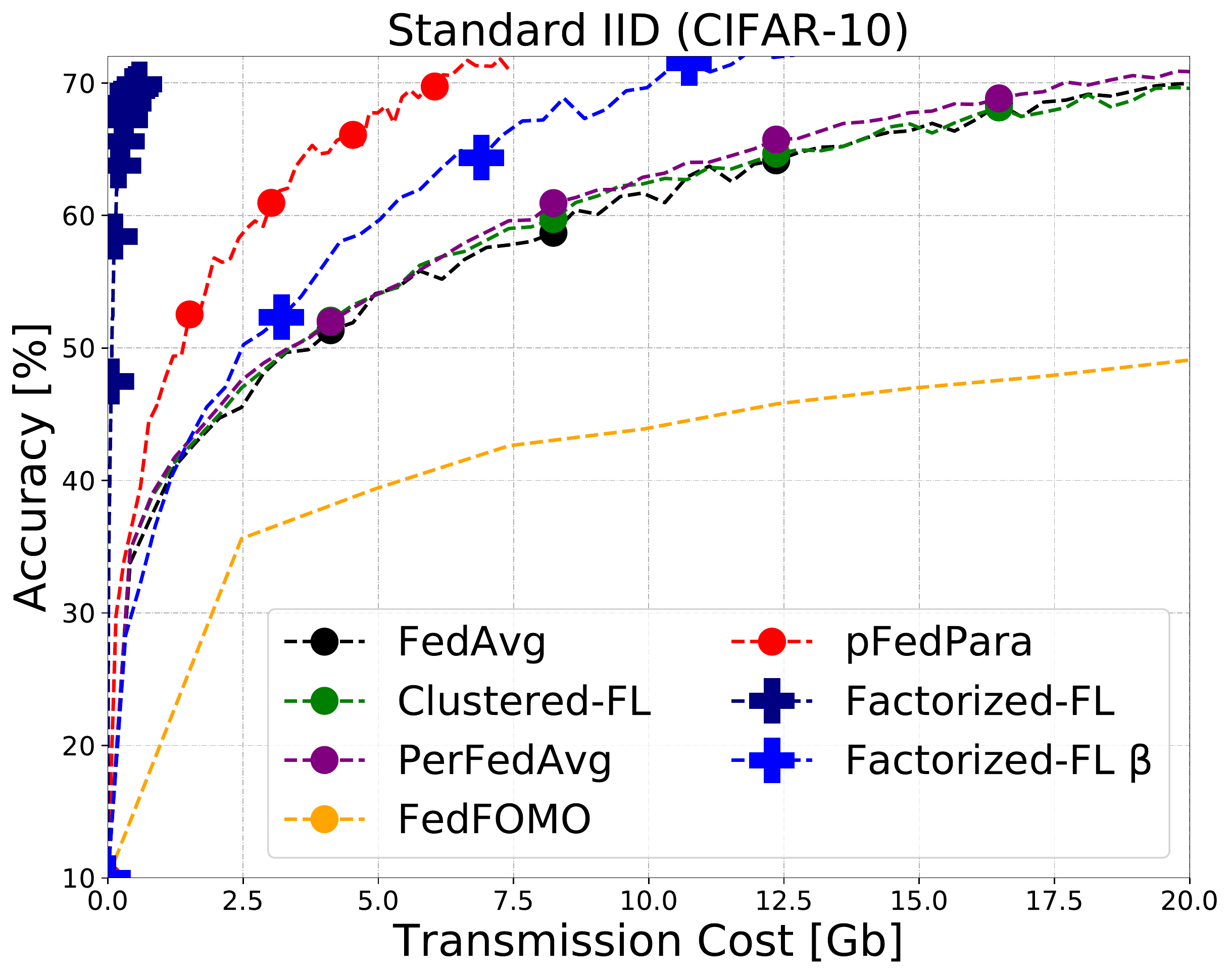} & \includegraphics[width=0.38\textwidth]{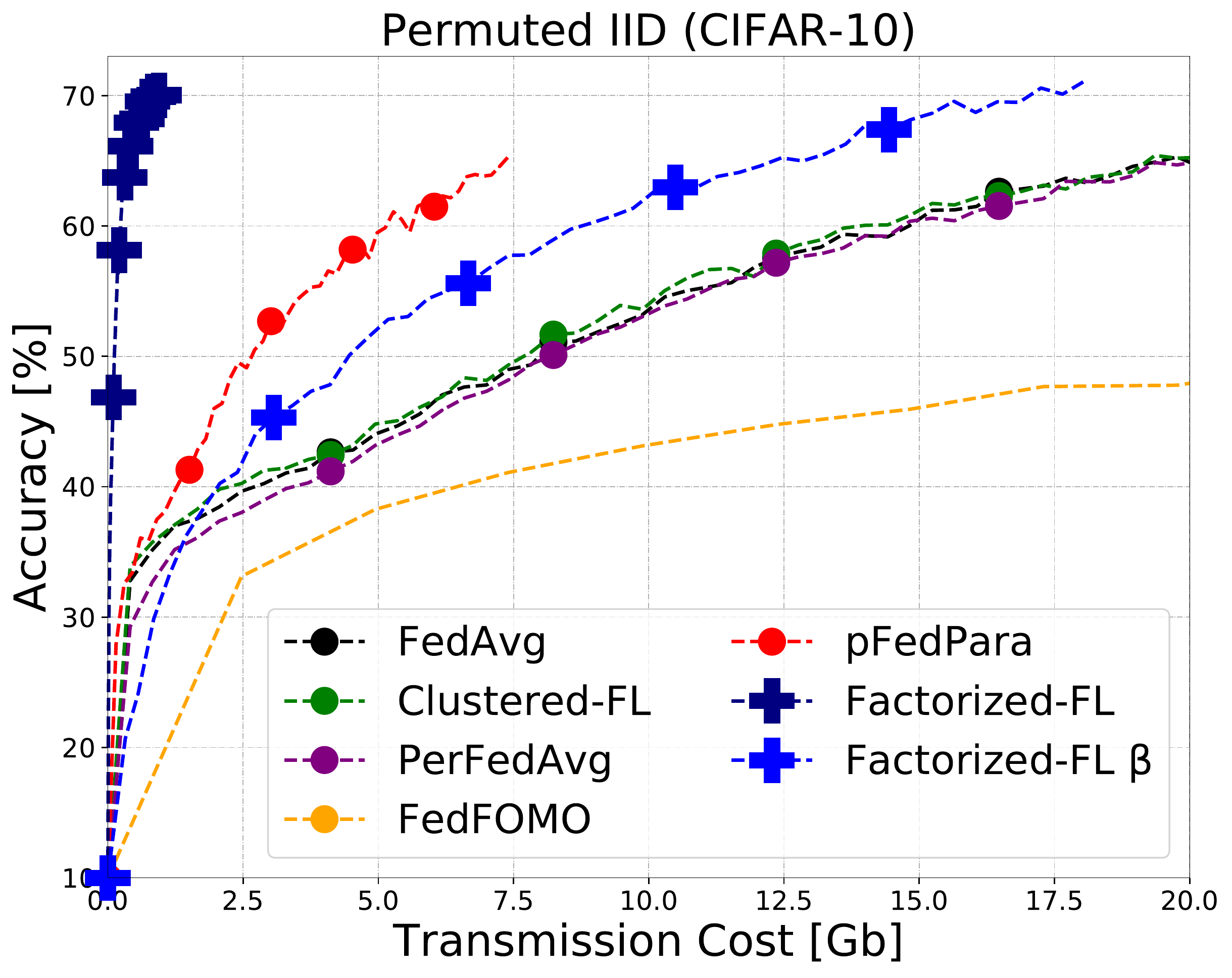} \\
         (a) Standard IID (CIFAR-10) & (b) Permuted IID (CIFAR-10) \\
         
         \includegraphics[width=0.38\textwidth]{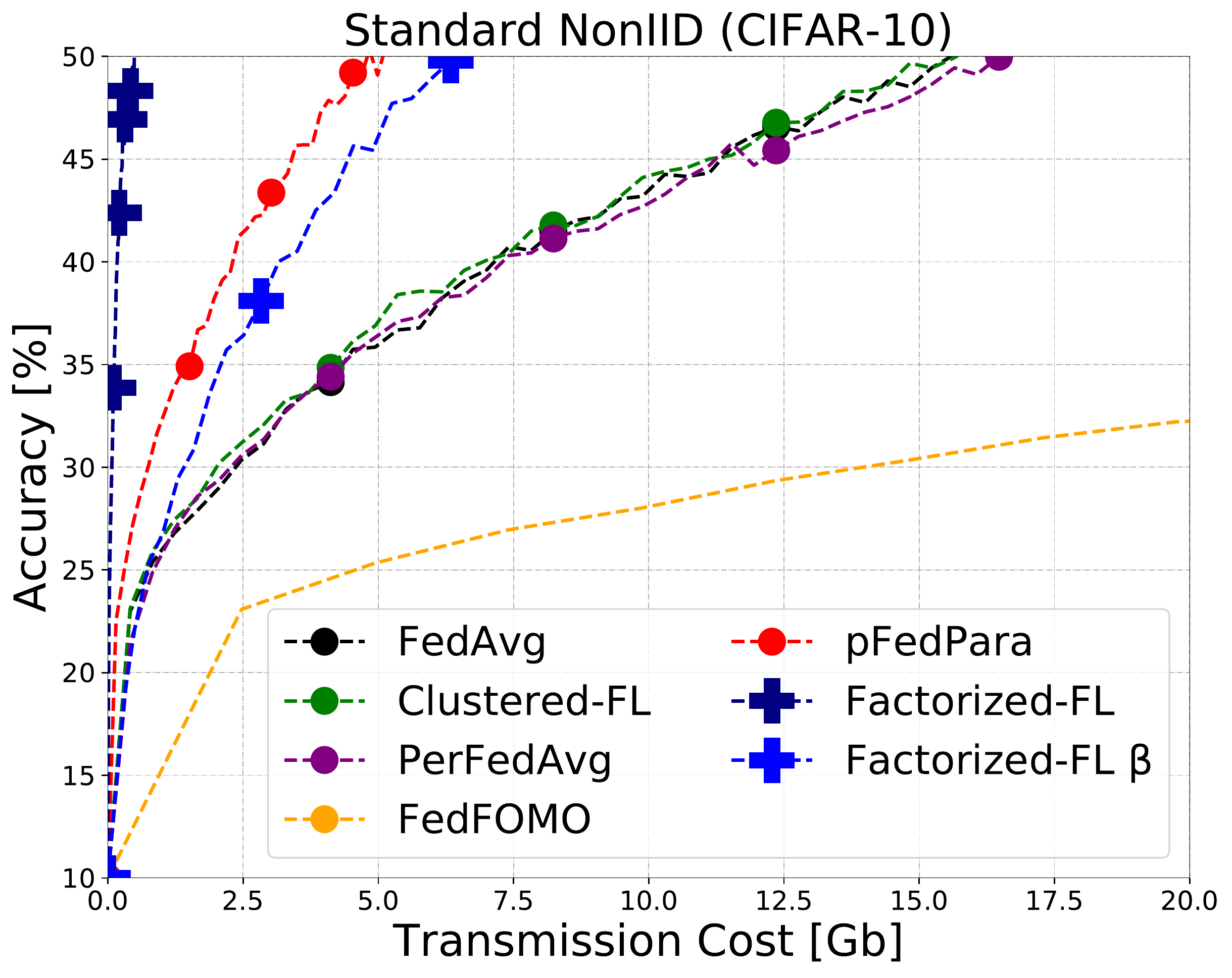} & \includegraphics[width=0.38\textwidth]{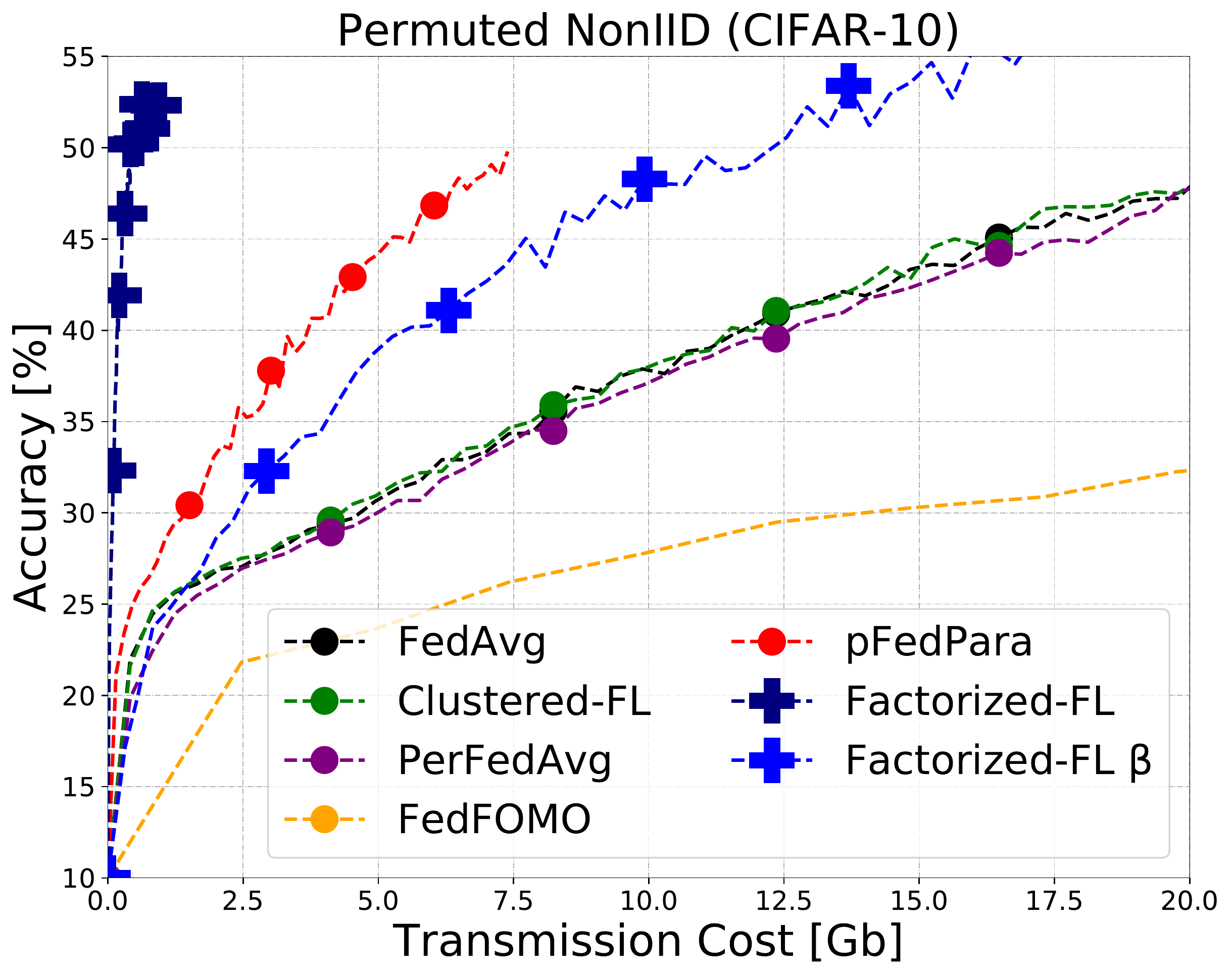} \\
         (c) Standard NonIID (CIFAR-10) & (d) Permuted NonIID (CIFAR-10) \\
         
         \includegraphics[width=0.38\textwidth]{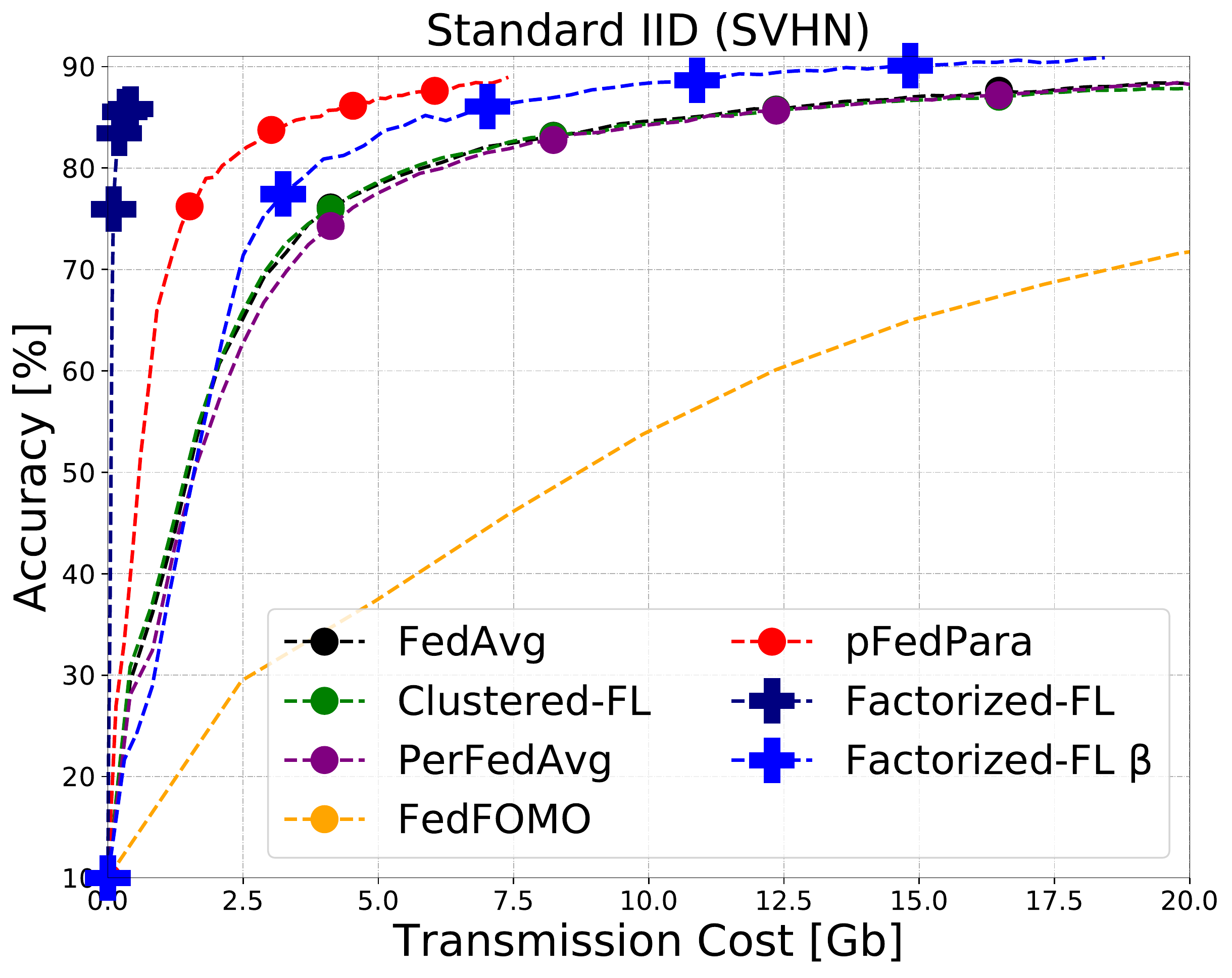} & \includegraphics[width=0.38\textwidth]{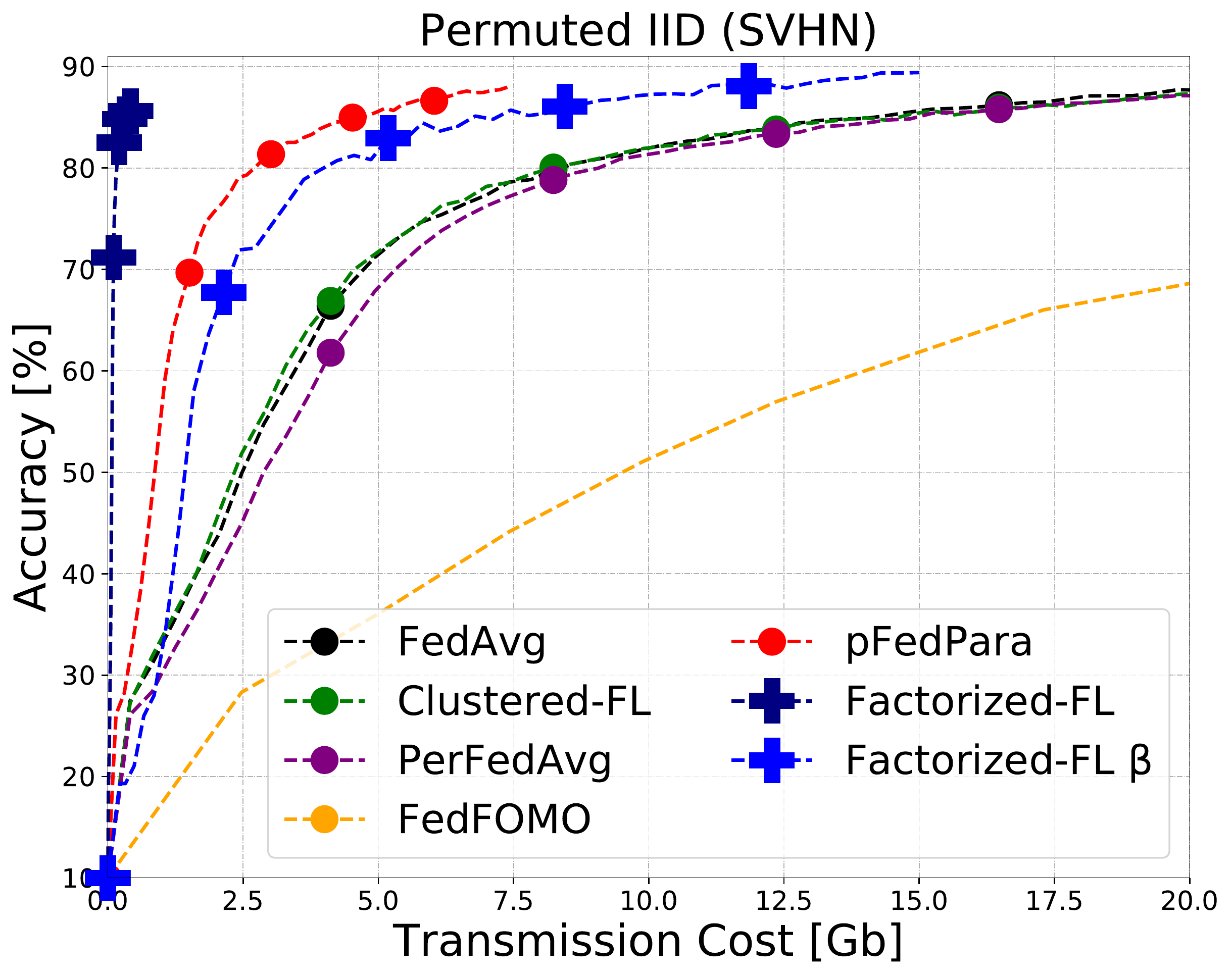} \\
         (e) Standard IID (SVHN) & (f) Permuted IID (SVHN) \\
         
         \includegraphics[width=0.38\textwidth]{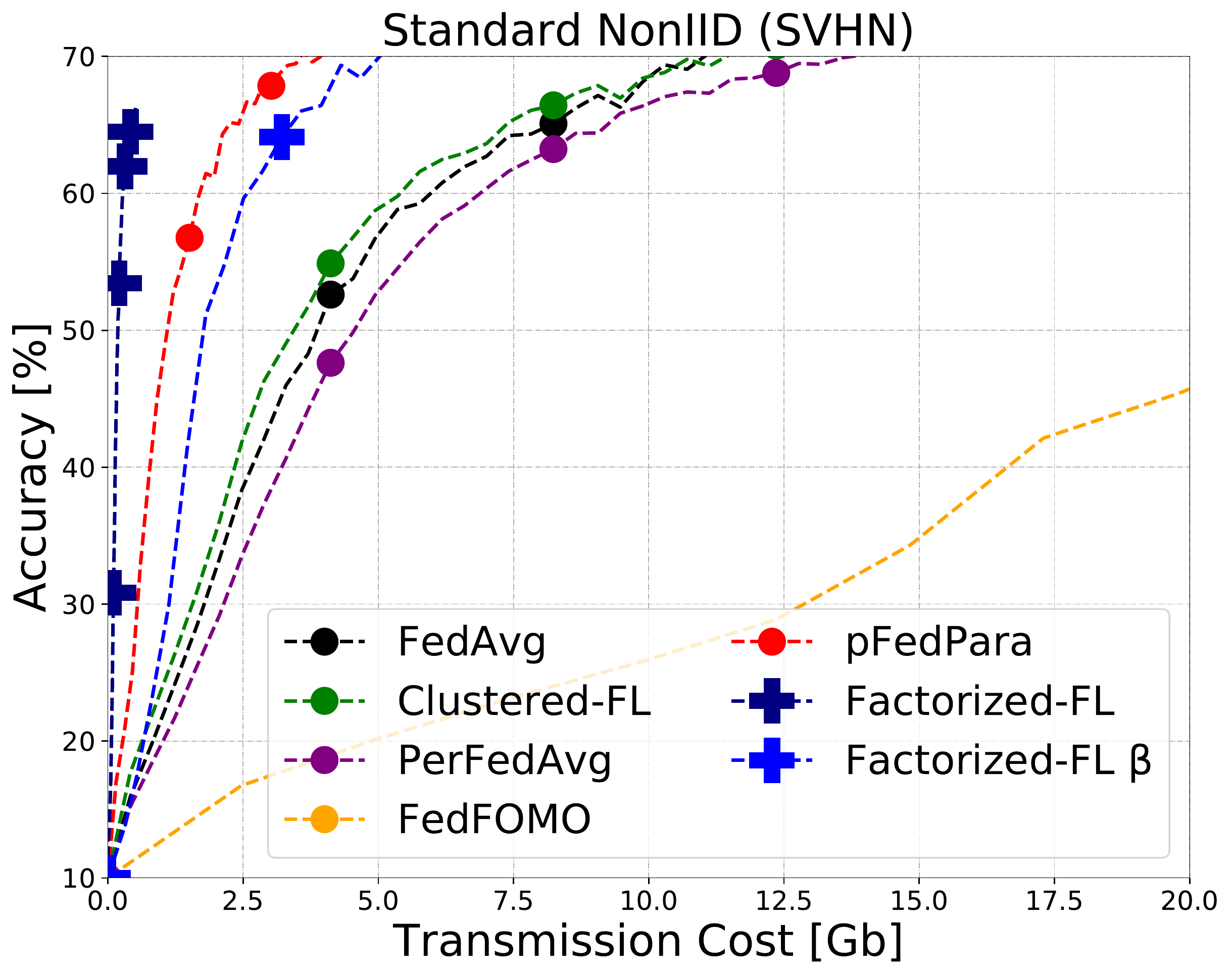} &  \includegraphics[width=0.38\textwidth]{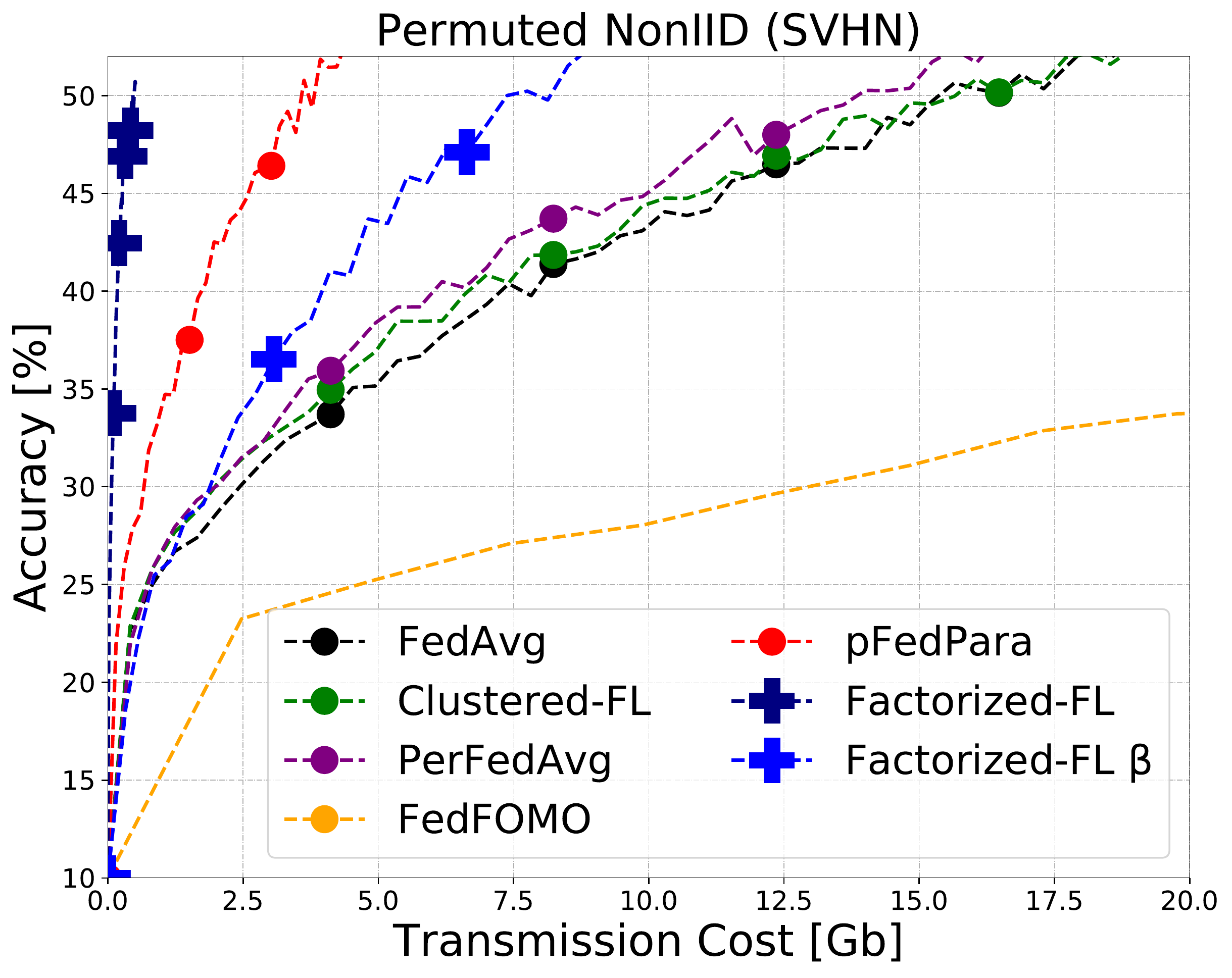} \\
         (g) Standard NonIID (SVHN) & (h) Permuted NonIID (SVHN) \\
         
    \end{tabular}
    \caption{\textbf{Test accuracy over communication costs for standard federated learning and label-heterogeneous FL scenario}: We provide test accuracy curves on CIFAR-10 and SVHN in standard iid/non-iid and permuted iid/non-iid partitions ($E$=$5$,$R$=$50$). }
    \label{fig:all}
\end{figure*}


\end{document}